\pdfoutput=1

\documentclass[11pt]{article}

\usepackage[review]{acl}

\usepackage{times}
\usepackage{latexsym}

\usepackage[T1]{fontenc}

\usepackage[utf8]{inputenc}

\usepackage{microtype}
\usepackage{tcolorbox}
\usepackage{multirow}
\usepackage{tabularx} 
\usepackage{inconsolata}

\usepackage{graphicx}
\usepackage{tcolorbox}
\usepackage{multirow}
\usepackage{tabularx} 
\usepackage{longtable}
\usepackage{ragged2e}
\usepackage{booktabs}
\usepackage{xcolor}
\definecolor{RubinRed}{HTML}{9B111E}
\title{GC-KBVQA: A New Four-Stage Framework for Enhancing Knowledge Based Visual Question Answering Performance}

\author{Mohammad Mahdi Moradi \\
  Affiliation / Address line 1 \\
  Affiliation / Address line 2 \\
  Affiliation / Address line 3 \\
  \texttt{email@domain} \\\And
  Second Author \\
  Affiliation / Address line 1 \\
  Affiliation / Address line 2 \\
  Affiliation / Address line 3 \\
  \texttt{email@domain} \\}

\author{
    \textbf{Mohammad Mahdi Moradi\textsuperscript{1}},
    \textbf{Sudhir Mudur\textsuperscript{1}},
\\
\textsuperscript{1}Concordia University,
\small{
    \textbf{Correspondence:} \href{mailto:email@domain}{mudur@cs.concordia.ca}
}
}

\begin{document}
\maketitle
\begin{abstract}
Knowledge-Based Visual Question Answering (KB-VQA) methods focus on tasks that demand reasoning with information extending beyond the explicit content depicted in the image. Early methods relied on explicit knowledge bases to provide this auxiliary information. Recent approaches leverage Large Language Models (LLMs) as implicit knowledge sources. While KB-VQA methods have demonstrated promising results, their potential remains constrained as the auxiliary text provided may not be relevant to the question context, and may also include irrelevant information that could misguide the answer predictor. We introduce a novel four-stage framework called Grounding Caption-Guided Knowledge-Based Visual Question Answering (GC-KBVQA), which enables LLMs to effectively perform zero-shot VQA tasks without the need for end-to-end multimodal training. Innovations include grounding question-aware caption generation to move beyond generic descriptions and have compact, yet detailed and context-rich information. This is combined with knowledge from external sources to create highly informative prompts for the LLM. GC-KBVQA can address a variety of VQA tasks, and does not require task-specific fine-tuning, thus reducing both costs and deployment complexity by leveraging general-purpose, pre-trained LLMs. Comparison with competing KB-VQA methods shows significantly improved performance. Our code will be made public. 
\end{abstract}

\section{Introduction}
Visual Question Answering (VQA) \cite{de2023visual} is a multimodal AI research challenge that has attracted significant interest from experts in computer vision, natural language processing, knowledge representation, and other machine learning domains. VQA is defined as the problem of answering a question in the context of a given image. The reasoning capabilities of VQA models are constrained when the correct answer demands information that extends beyond that which is explicitly present in the image and question. One example of this can be seen in Figure~\ref{fig:onecol} and more in the Supplementary Material.
Knowledge-Based VQA (KB-VQA) attempts to address this challenge by leveraging external knowledge sources and employing question-dependent captioners to generate textual descriptions of the given image. This approach aims to extract question-relevant information from both the image and external sources, enhancing the overall comprehension of the given input. Recent trends are to use a Large Language Model (LLM) as a general pretrained knowledge source instead of more specific knowledge repositories \cite{barezi2024find}.

Retrieval-based KB-VQA is the term used to refer to methods that attempt to retrieve relevant auxiliary information from external knowledge sources, such as Wikipedia, ConceptNet \cite{liu2004conceptnet}, or domain-specific knowledge bases, to enhance the accuracy and robustness of answers to posed questions. These methods are particularly useful for questions that require explicit knowledge. However, they face scalability challenges as they may need new knowledge sources for each domain-specific task and/or regular updates to external knowledge bases. Additionally, in this approach, the auxiliary information may tend to include noise or, in some cases, may fail to contain the information required to answer the question.

LLMs, through extensive pre-training on diverse text datasets, inherently acquire a substantial amount of factual knowledge \cite{wang2023survey}, enabling them to answer a wide variety of general knowledge questions without the need for task-specific retraining. 
In the context of VQA, LLMs can process textual information extracted from both questions and images, leveraging their pre-learned knowledge to complement the information explicit in the image for generating answers. This capability makes them particularly valuable in situations where implicit external knowledge is required to resolve ambiguities or fill gaps in the visual input \cite{barezi2024find}.

In \cite{guo2022images, lan2023improving, liu2024zvqaf} LLMs are used for cost-effective zero-shot VQA, removing the need for expensive and time-consuming domain-specific training data. As stated earlier, LLM-based approaches generalize well across various question types and domains. 
For instance, models like GPT-3 \cite{brown2020language} exhibit significant world knowledge, enabling them to answer questions on different domains without specific training. LLM based VQA models also do not require frequent updates or retraining, as they leverage the pre-existing knowledge in the LLM, making them scalable and flexible for new tasks.

One of the main challenges faced by current LLM based KB-VQA methods is in generating question-relevant prompts for the LLM to retrieve accurate information, often resulting in a mismatch between visual content, question context, and auxiliary information. This limitation also becomes critical in tasks like higher-order counting, which require understanding specific relationships between objects. Furthermore, since LLMs are unable to process raw visual data directly, they are heavily dependent on auxiliary inputs such as relevant captions and exemplar question-answers, where the quality and relevance of the provided information significantly impact reasoning and performance. To improve accuracy, it is essential to focus on strategies that deliver enriched, high-quality, question-relevant auxiliary content while filtering out irrelevant information to effectively bridge the gap between visual input and textual reasoning.

In this work, we propose GC-KBVQA, an alternative zero-shot LLM based KB-VQA architecture built on a four-stage framework (cf. Figure~\ref{fig:onecol}). The different stages are designed to address the principal challenges in providing auxiliary information, namely, (i) be relevant to the question context, (ii) contain key information required for predicting the correct answer, (iii) generate informative prompts that guide the answer predictor, and (iv) have minimum of irrelevant information. The GC-KBVQA framework is described in detail in Section \ref{method}. 

Our main contributions are as follows:
\begin{itemize}
    \item We present GC-KBVQA, a new zero-shot VQA method whose innovations include question-aware visual segmentation and caption generation, content-oriented caption filtering, and caption-driven question context exemplar QA pairs to guide the answer predictor, thus providing a more comprehensive understanding of the input image, while at the same time eliminating the need for end-to-end training and/or in-context examples.
    \item By incorporating a grounding block for precise visual object detection, our method tackles counting challenges better than all previous works.

\end{itemize}

\section{Related Works}
A wide range of VQA methods have been introduced, which can be broadly categorized into three groups\cite{kafle2017visual}: (i) joint embedding, (ii) attention mechanisms, and (iii) external knowledge integration, also known as KB-VQA. Joint embedding methods \cite{antol2015vqa, fukui2016multimodal, ma2016learning} map visual and textual inputs into a common latent space for answer prediction but typically produce only coarse‑grained multi‑modal features, introducing noise that degrades prediction accuracy. Attention-based methods \cite{yang2016stacked, anderson2018bottom, cadene2019murel} dynamically weight image regions and question words to improve fusion but remain limited to information explicitly in the image and question, motivating KB‑VQA approaches that integrate external knowledge.

External knowledge-based approaches can be broadly categorized into two types: explicit and implicit knowledge integration \cite{shao2023prompting}. Explicit methods \cite{wu2022multi, ravi2023vlc} retrieve knowledge from large-scale databases, such as Wikipedia, ConceptNet, or Google, to support reasoning \cite{wu2022multi}. However, these approaches are limited by their lack of task-specific information and often introduce irrelevant information, complicating the reasoning process \cite{barezi2024find}. Implicit methods \cite{lan2023improving, yang2022empirical, shao2023prompting, hu2022promptcap, tiong2022plug, du2023zero, cao2024knowledge, guo2023images}, on the other hand, leverage off-the-shelf captioning models to generate image descriptions with off‑the‑shelf captioners, fuse them with the question and relevant extracted cues, and query LLMs (e.g., GPT‑3 \cite{brown2020language}) for the answer. 

Recent examples of KB-VQA can be seen in \cite{hu2022promptcap, shao2023prompting, yang2022empirical}.  A straightforward pipeline is adopted for extracting external knowledge to facilitate answer generation as described next. Initially, a caption generator creates textual descriptions of the input image, which are then combined with the question and a set of in-context examples to form a prompt for an LLM to generate the answer. Although these methods effectively utilize LLMs in a few-shot setting, they have several limitations. Inaccurate selection of in-context examples can lead to performance degradation. Additionally, Generic image captions often provide broad overviews of visual content but may omit details crucial for answering specific questions, limiting their effectiveness in VQA tasks. 

Several zero-shot VQA frameworks have also been proposed in KB-.VQA. For instance, \citet{du2023zero, liu2024zvqaf} use LLMs to evaluate and improve captions during training, but their reliance on LLM assessments risks bias toward style over content, reducing performance. In the systems described in \cite{guo2023images, tiong2022plug}, they proposed systems that first identify image patches relevant to the question using an image-question matching module. Subsequently, diverse captions are generated based on these patches and used to create question-answer pairs, which are then provided to an LLM to produce the final answer. However, these methods face limitations such as failures in matching relevant image patches, resulting in captions that miss important regions. Additionally, the patch-based division may oversimplify complex relationships by focusing on local areas and ignoring global context, which is crucial for questions involving object interactions and overall scene understanding.

Providing LLMs with relevant, context-rich prompts from images and external knowledge is crucial, while minimizing irrelevant information to avoid answer confusion. The GC-KBVQA method present in this paper is inspired by earlier patch-based methods \cite{guo2023images, tiong2022plug}. However, as discussed in the following section, 
We introduce innovations in the zero-shot pipeline to extract comprehensive, noise-free, and question-relevant information. Additionally, we address counting task challenges and use keyword-guided visual grounding to capture complex relationships and global context, overcoming prior limitations.

Comprehensive comparison results show that GC-KBVQA significantly outperforms state-of-the-art systems. More details are presented in subsequent sections.

\section{Method}
\label{method}
GC-KBVQA is a zero-shot cost-efficient VQA framework that operates in four sequential stages. In the following sections, we describe these four stages in more detail.
\begin{figure*}[t]
  \centering
   \includegraphics[width=\textwidth]{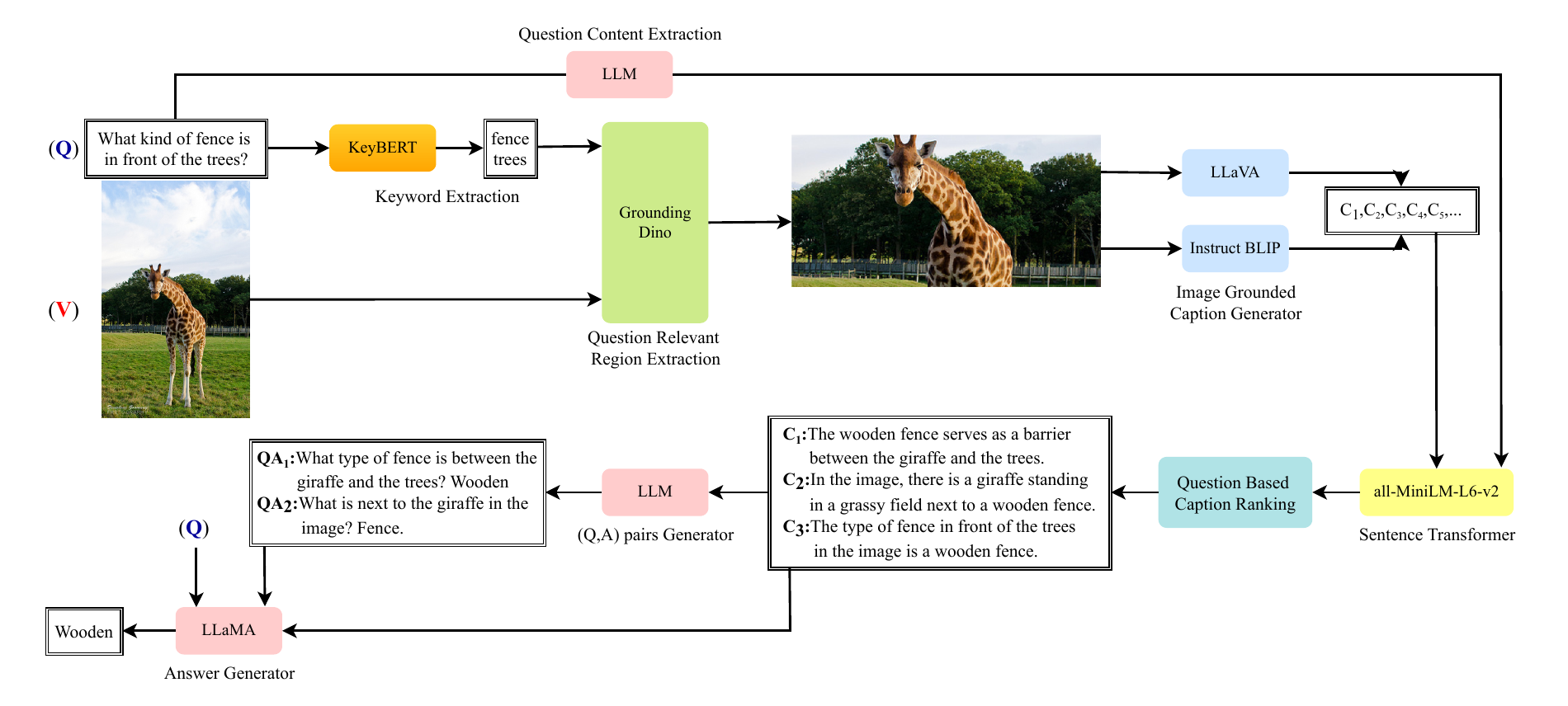}
   \caption{The architecture of GC-KBVQA leverages four pre-trained modules: (1) Question-relevant region extraction identifies and grounds the image regions pertinent to the question, (2) Pre-trained caption generator(s) generates captions for these grounded regions and applies filtering and ranking mechanisms to prioritize captions based on semantic relevance to the question, (3) Q\&A pair generator constructs task-specific question-answer pairs derived from the selected captions, (4) Answer generator integrates captions, and QA pairs to deliver contextually coherent answers.}
   \label{fig:onecol}
\end{figure*}

\subsection{Keyword-Guided Visual Grounding}
The grounding stage serves as a fundamental component and a key distinguishing aspect of our framework, responsible for identifying the most relevant regions in the input image based on the contextual cues of the question. To achieve this, we leverage Grounding DINO \cite{liu2023grounding}, a state-of-the-art object detection and grounding model that precisely locates and extracts visual elements that align with the intent of the question. 

To effectively guide Grounding DINO in the grounding process, an input prompt has to be provided. In order to construct this prompt, the input question is analyzed using KeyBERT \cite{sharma2019self}, a lightweight and efficient keyword extraction technique. KeyBERT generates question embeddings with BERT \cite{kenton2019bert} to capture the overall semantic representation. Then we compute cosine similarity between N-gram word embeddings of candidate keywords and key phrases with the given question and identify the most relevant terms. Words and phrases with a relevance score exceeding 0.4 are selected as key descriptors of the question. These extracted keywords are then combined into a structured and contextually appropriate prompt for Grounding DINO. Leveraging only the highly relevant keywords enables more precise grounding, improving region extraction and enhancing performance in downstream tasks like the caption generator.

Grounding DINO generates multiple candidate objects. To refine these, a confidence threshold of 0.25 is applied, retaining only bounding boxes with confidence scores exceeding this threshold to filter out weak or irrelevant detections. To further optimize the results, a post-processing step evaluates the overlap between bounding boxes by computing their intersection areas. If the intersection exceeds 0.9, only the larger bounding box is retained which contains more comprehensive information. For questions that require understanding spatial relationships or object locations, the detected object alone may not provide sufficient contextual information. So, we introduce a region expansion mechanism that proportionately enlarges bounding boxes based on their original dimensions to ensure that relevant surrounding areas are included, enhancing the ability of the model to capture the necessary contextual details. The refined regions are forwarded to the caption generation stage.

To address counting issues in prior work, we classify questions into two categories: counting and non-counting. We leverage LLaMA-3-8B-Instruct for question classification, distinguishing between counting and non-counting types. Counting questions are directed to Grounding DINO, which uses question-derived keywords to detect relevant objects, with the object count used as the final answer. Non-counting questions proceed through the standard reasoning pipeline (Figure \ref{fig:onecol}). This design ensures efficient resource allocation and enhances accuracy, while emphasizing the importance of reliable question classification on overall performance.
\subsection{Caption Generation and Filtering}
Captions are the main source of auxiliary information to be presented to the answer generator. Clearly, captions must be comprehensive but not excessive, have minimum noise, and must be relevant to the question. We present a 3 step strategy to achieve this. 

1) Two Vision Language Models (VLMs) are utilized for caption generation, LLaVA \cite{liu2024improved} and InstructBlip \cite{instructblip}, bringing complementary strengths in interpreting and describing visual content. We do this to help enhance the breadth of descriptive details, reduce the risk of overlooking key information, and mitigate potential biases or limitations inherent in individual captioning models. As we show in our ablation studies, Table \ref{table:2}, these two yielded the best results.  

2) Instead of limiting the VLMs to producing a single caption per region, we instruct them to generate multiple captions for each identified region of interest to increase the comprehensiveness and descriptiveness of the visual content and to help collect key information. We have noticed that this strategy captures a wider range of details, including prominent objects, textures, relationships, and actions within the image. 

3) To ensure that the answer predictor receives only the most relevant information, a filtering step is employed to evaluate and retain only the captions that are closely aligned with the question.  Questions often include verbose or extraneous content that obscures their core intent, potentially misleading the filtering process. This can result in retaining less relevant captions or discarding highly pertinent ones, ultimately compromising the accuracy of the answer prediction. To address this issue, we introduce a content extraction step. The Llama-3-8B-Instruct model is prompted with the instruction: "\textit{Determine the main idea of this question in short:}". This prompt enables the generation of a concise summary of the question, isolating its essential content and intent. The extracted main idea is subsequently used as the reference for computing the semantic relevance of each caption. By eliminating unnecessary details in the text of the question, this approach enhances the precision of the filtering process, ensuring a greater focus on selecting the most relevant captions. As our results show this filtering step significantly improves the effectiveness and accuracy of the answer prediction pipeline.

Here, relevance is determined using all-MiniLM-L6-v2 \cite{galli2024performance}, a lightweight model designed for efficient sentence and corpus embeddings. With the capability to encode up to 14,200 sentences per second, it is well suited for real-time and large-scale applications. The model maps both the distilled question and captions into a 384-dimensional dense vector. Then, cosine similarity is calculated between the embeddings of the distilled question and captions. The captions are ranked according to their semantic relevance to the distilled question. From this ranked list, the top 3 most contextually relevant captions are selected to serve as auxiliary information for the answer predictor. Our ablation studies, Table \ref{table:3}, show that the top 3 work best. 


\subsection{Caption-driven QA pairs}
In the next stage, a Llama-3-8B-Instruct model is utilized to generate caption-driven QA pairs based on the filtered captions. Two QA pairs are derived from the top three captions to further enrich the auxiliary information provided to the final answer predictor. If C1, C2 and C3 denote the top 3 captions, then LLaMA model is prompted with a clear and structured instruction as follows:
"\textit{Generate two (Question, Answer) pairs for the following captions in the format (Question, Answer):}\textbackslash n\textit{Caption 1:} [C1],\textbackslash n\textit{Caption 2:} [C2],\textbackslash n\textit{Caption 3:} [C3]." The two QA pairs, say (Q1,A1) and (Q2,A2), are then included in the answer predictor prompt to guide the model towards producing accurate and contextually relevant answers. These QA pairs illustrate the desired structure, tone, and level of detail, leveraging few-shot learning principles. 
Our experiments indicate that two QA pairs are sufficient to maintain efficiency and effectiveness. Limiting the number of QA pairs ensures a balanced representation, allowing the model to focus on high-quality, informative content while avoiding unnecessary complexity and redundancy. In contrast, providing the answer predictor with excessive QA pairs can lead to information saturation, making it difficult for the model to prioritize the most relevant details.

\subsection{Integrating Information for Answer Generation}
In the final stage, the test question, the top three question-aware captions, and the two caption-driven QA pairs are consolidated into a unified, comprehensive prompt along with instructions. This prompt is then provided to the Llama-3-8B-Instruct model. The prompt follows a structured format, carefully designed to ensure clarity and provide the model with all necessary context and relevant auxiliary information:
"\textit{Infer an answer for the following question based on the provided pieces of information and formatted Question-Answer pairs:}\textbackslash n\textit{Caption 1:} [C1]\textbackslash n\textit{Caption 2:} [C2]\textbackslash n\textit{Caption 3:} [C3]\textbackslash n[Q1]: [A1]\textbackslash n[Q2]: [A2]"
This structured format reduces ambiguity, providing the model with clear and precise guidance to improve the quality of generated answers. Additionally, presenting QA pairs in a concise manner encourages the model to infer a succinct response, ensuring brevity and clarity in its output. This step effectively integrates information from previous processing stages, enabling the model to generate a well-informed and contextually relevant answer to the given question.

\section{Experimental Evaluations}
\label{sec:ResultsDiscussion}

\subsection{Setup}
All experiments were performed on a machine equipped with an NVIDIA RTX A6000 with 45GB memory, and on three standard VQA benchmarks: OK‑VQA (5,046 test samples) \cite{marino2019ok} and A‑OKVQA (1,145 test samples) \cite{schwenk2022okvqa}, both of which require external knowledge for question answering, and VQAv2 (27,000 test samples) \cite{goyal2017making}, an enhanced version of VQAv1 designed to evaluate general VQA performance without additional knowledge retrieval.

\subsection{Main Results}
Table \ref{table:5} compares our method, categorized as zero-shot evaluation without end-to-end training, with competing approaches across three VQA paradigms. The first is zero-shot evaluation without task-specific fine-tuning, leveraging the inherent generalization of pre-trained models. The second involves zero-shot evaluation with end-to-end fine-tuning on task-specific data to improve performance. The third is few-shot evaluation, which uses a small amount of labeled data to refine predictions by combining pre-trained knowledge with limited supervision.

In a zero-shot setting without task-specific training, our model outperforms prior methods on OK-VQA, A-OKVQA, and VQAv2, demonstrating strong generalization and robustness. It improves region selection, captioning, and filtering in region-based approaches like \citet{guo2023images}, achieving gains of +8.97\%, +10.97\%, and +6.06\% while using 47B vs. 175B parameters, highlighting its design efficiency. We believe that this is due to the following: Unlike Img2LLM \cite{guo2023images} and PNP-VQA \cite{tiong2022plug}, that rely on arbitrary patch division and often disrupt spatial coherence, our framework uses Grounding DINO and KeyBERT to identify semantically meaningful regions, enabling more relevant caption generation. Simultaneous use of two different caption generators yields complementary descriptions, and a question content-driven caption filtering mechanism removes redundant and non-informative captions, and makes our method more context-aware. 

\begin{table*}[t]
  \centering
  \setlength{\tabcolsep}{5pt}
  \begin{tabular}{lccccc}  
    \toprule
    Method  & Size & OK-VQA & A-OKVQA & VQAv2 \\
    \midrule
        \multicolumn{5}{c}{\textit{Zero-shot evaluation without extra end-to-end training}} \\
    PICa \cite{yang2022empirical} & 175B & 17.7 & - & -\\
    PNP-VQA \cite{tiong2022plug} & 11B & 35.9 & - & \underline{64.8}\\
    Img2LLM \cite{guo2023images} & 30B & 41.8 & 38.7 & 60.3\\
    Img2LLM \cite{guo2023images} & 175B & 45.6 & 42.9 & 61.9\\
    LAMOC \cite{du2023zero} & 11B & 40.3 & 37.9 & - \\
    KGenVQA \cite{cao2024knowledge} & 11B & 45.4 & 39.1 & - \\
    ZVQAF \cite{liu2024zvqaf} & 11B  & 40.5 & 37.1 & 64.3 \\
    LLM\_Guided\_VQA\cite{qiang2025enhancing} & - & 43.7 & 42.1 & 61.2\\
    RQPrompt \cite{lan2023improving} & 175B & 46.4 & 43.9 & - \\
    DecomVQA \cite{khan2024exploring} & 13B & 39.79 & \underline{53.36} & - \\
    DIETCOKE \cite{li2024diversify} & 7B & \underline{49.2} & 48.6 & - \\
    \textbf{ours} & 47B & \textbf{54.57} & \textbf{53.87} & \textbf{67.96}\\
    \midrule
    \multicolumn{5}{c}{\textit{Zero-shot evaluation with extra end-to-end training}} \\
    VLKD \cite{dai2022enabling} & 408B & 13.3 & - & 44.5\\
    Flamingo \cite{alayrac2022flamingo} & 80B & 50.6 & - & 56.3 \\
    BLIP-2 \cite{li2023blip} & 12B & 45.9 & - & 65.2\\
    \midrule
    \multicolumn{5}{c}{\textit{Few-shot evaluation}} \\
    PICa \cite{yang2022empirical} & 175B & 46.5 & - & 54.3\\
    PromptCAP \cite{hu2022promptcap} & 175B & 60.4 & 56.3 & -\\
    Prophet \cite{shao2023prompting} & - & 61.1 & 58.2 & -\\
    \bottomrule
  \end{tabular}
  \caption{Performance comparison of our method with state-of-the-art approaches under zero-shot and few-shot evaluation settings across OK-VQA, A-OKVQA, and VQAv2 benchmarks.}
  \label{table:5}
\end{table*}

As can be seen in Table \ref{table:5}, our 47B GC-KBVQA model outperforms larger models like Flamingo (80B) and BLIP-2, despite their reliance on extensive pre-training or strong baselines. GC-KBVQA surpasses Flamingo with greater efficiency and achieves gains of +8.67\% on OK-VQA and +2.76\% on VQAv2 over BLIP-2, attributed to our question content-aware prompt generation strategy. The third part of the table shows that our zero-shot GC-KBVQA model, despite using fewer parameters, outperforms notable few-shot baselines such as PICa and achieves performance comparable to state-of-the-art methods like PromptCap. This demonstrates the efficiency and strong generalization capability of our modular framework, which leverages pre-trained components for knowledge-intensive VQA tasks while avoiding the overhead of supervised fine-tuning.
\begin{table}[t]
  \centering
  \setlength{\tabcolsep}{1pt}
  \begin{tabular}{lcccc} 
    \toprule
    Model & Size & OK-VQA & A-OKVQA & VQAv2  \\
    \midrule
    TinyLlaMA-1B & 19B & 49.45 & 48.04 & 66.26\\
    Qwen1.5-1.8B & 22B & 47.74 & 45.86 & 62.79\\
    Mistral-3B   & 27B & 50.30 & 50.23 & 67.11\\
    Mistral-7B   & 43B & \underline{52.17} & \underline{50.95} & \textbf{68.81}\\
    Llama3-8B  & 47B & \textbf{54.57} & \textbf{53.87} & \underline{67.96} \\
    \bottomrule
  \end{tabular}
  \caption{Comparing the performance of our proposed framework for different LLM sizes and types}
  \label{table:11}
\end{table}

\subsection{Ablation Study on LLM type and size}
Table \ref{table:11} presents the impact of different LLMs and overall framework sizes, on performance. Notably, replacing Llama‑3‑8B with Mistral‑7B yields only a marginal drop in accuracy, yet still outperforms all baseline systems except DecomVQA on the A‑OKVQA benchmark when compared to the results in Table \ref{table:5}. This demonstrates that our GC‑KBVQA framework is truly “plug‑and‑play”: one can simply substitute Llama‑3‑8B for another model of similar scale (e.g., Mistral‑7B) using identical configurations and prompts—without any additional prompt tuning—and still achieve competitive results.

Similarly, downsizing from Llama‑3‑8B to TinyLLaMA‑1B (reducing the framework size from 47B parameters to just 19B) incurs only a 1–5\% drop in accuracy, yet the TinyLLaMA‑1B variant still surpasses all other models except DecomVQA on A‑OKVQA. This highlights that GC‑KBVQA not only tolerates reductions in LLM capacity with minimal performance degradation but also maintains state‑of‑the‑art accuracy even with significantly fewer parameters. These results underscore the versatility and efficiency of our approach across a range of LLM sizes.

\subsection{Ablation Study on Grounding Dino Prompt Design}
Table \ref{table:1} analyzes prompt design in the grounding stage of GC-KBVQA by comparing using the full question versus KeyBERT-extracted key phrases as prompts for Grounding DINO. Results show that using KeyBERT-extracted key phrases significantly improves performance by +10.36, +5.28, and +9.97 points in OK-VQA, A-OKVQA, and VQAv2. The Results reveal that full-question prompts often include irrelevant details that hinder focus on key visual regions. In contrast, KeyBERT-extracted key phrases guide Grounding DINO to more accurately localize relevant content, improve performance on complex queries and detailed images, and enhance generalization across benchmarks and domains.
\begin{table}[h]
  \centering
  \setlength{\tabcolsep}{3pt}
  \begin{tabular}{lccc} 
    \toprule
    Dino Promp & OK-VQA & A-OKVQA & VQAv2  \\
    \midrule
    Question   & 44.21 & 48.69 & 57.99 \\
    KeyBERT    & \textbf{54.57} & \textbf{53.87} & \textbf{67.96} \\
    \bottomrule
  \end{tabular}
  \caption{Comparing the performance of Grounding Dino prompt types (Question vs. KeyBERT) across three VQA datasets: OK-VQA, A-OKVQA, and VQAv2}
  \label{table:1}
\end{table}

\subsection{Ablation Study on Use of Two Caption Generators}
Table \ref{table:2} evaluates the performance of caption generation strategies, comparing the use of LLaVA and InstructBLIP individually against their combined use. LLaVA, when used as the sole caption generator, outperforms InstructBLIP in all benchmarks, demonstrating its superior alignment with downstream VQA tasks. However, combining captions from LLaVA and InstructBLIP yields significant performance improvements by producing a diverse and contextually rich set of captions that can help enhance reasoning capabilities. The use of two caption generators mitigates the limitations of incomplete or narrow captions, enhancing generalization across datasets and question formats. Although this approach incurs slightly higher computational costs, it yields substantial accuracy gains, justifying the trade-off. Notably, increasing the number of caption generators beyond two did not lead to further gains. 
\begin{table}[h]
  \centering
  \setlength{\tabcolsep}{3pt}
  \begin{tabular}{cccc} 
    \toprule
    \multirow{2}{*}{\shortstack[c]{Caption \\ Generators}} & \multirow{2}{*}{OK-VQA} & \multirow{2}{*}{A-OKVQA} & \multirow{2}{*}{VQAv2}  \\
     & & & \\
    \midrule
    LLaVA                  & \underline{43.84} & \underline{50.31} & \underline{60.25} \\
    InstrcutBLIP           & 38.43 & 46.89 & 51.98 \\
    \multirow{2}{*}{\shortstack[c]{LLaVA+ \\ InstructBLIP}}   & \multirow{2}{*}{\textbf{54.57}} & \multirow{2}{*}{\textbf{53.87}} & \multirow{2}{*}{\textbf{67.96}} \\
    & & &\\
    \bottomrule
  \end{tabular}
  \caption{Impact of single vs. combined caption generators on GC-KBVQA performance.}
  \label{table:2}
\end{table}

\subsection{Ablation Study on the number of Captions}
Table \ref{table:3} highlights the effect of varying the number of captions on GC-KBVQA performance. The results show that the absence of captions leads to the lowest performance, highlighting the necessity of contextual knowledge for effective question answering. Performance shows significant improvement with the addition of captions, starting from a single caption and increasing progressively. The use of the top three captions yielded the best results in all datasets. This configuration effectively balances reducing redundancy and maximizing contextual utility for the LLM. However, adding a fourth or fifth caption seems to increase the likelihood of irrelevant or less question-focused content, which can dilute overall information quality and impede the LLM’s reasoning process. 

\begin{table}[h]
  \centering
  \setlength{\tabcolsep}{3pt}
  \begin{tabular}{lccc} 
    \toprule
    \multirow{2}{*}{\shortstack[c]{Number of \\ Captions}} & \multirow{2}{*}{OK-VQA} & \multirow{2}{*}{A-OKVQA} & \multirow{2}{*}{VQAv2}  \\
     & & & \\
    \midrule
    No Caption   & 18.81 & 11.01 & 27.85 \\
    1 Caption    & 40.38 & 47.51 & 53.15 \\
    2 Captions   & 42.62 & 51.97 & 58.83 \\
    3 Captions   & \textbf{54.57} & \textbf{53.87} & \textbf{67.96} \\
    4 Captions   & 55.40 & \underline{52.40} & 66.85 \\
    5 Captions   & \underline{56.47} & 48.13 & \underline{67.39} \\
    \bottomrule
  \end{tabular}
  \caption{Impact of varying the number of captions generated by dual generators}
  \label{table:3}
\end{table}

\subsection{Ablation Study on Answer Generator Prompt Design}
The performance of the Answer Predictor in GC-KBVQA is critically influenced by prompt design. Table \ref{table:4} evaluates the impact of different prompt designs. Prompts that contain only instructions result in significantly lower performance across datasets due to the absence of contextual information required to reason about visual content and questions. Incorporating QA pairs improves performance by simulating reasoning patterns; however, they could miss out on sufficient image-related content. Captions help the model associate specific regions and objects in the image with textual descriptions, making it easier to locate relevant entities when answering. However, incorporating structured reasoning examples can further enhance accuracy. The integration of instructions, captions, and QA pairs achieved the best results in all datasets. One point to note in this table is that the addition of QA pairs guides the answer predictor to make more accurate inferences. In summary, captions lead to more accurate knowledge retrieval, QA pairs enhance reasoning patterns, and instructions align the prompt with the task objective. This cohesive design enables the model to tackle diverse and complex questions effectively, ensuring comprehensive contextual understanding and strong overall performance.

\begin{table}[h]
  \centering
  \setlength{\tabcolsep}{0.75pt}
  \begin{tabular}{cccc} 
    \toprule
    \multirow{2}{*}{\shortstack[c]{Instructional \\ Combinations}} & \multirow{2}{*}{OK-VQA} & \multirow{2}{*}{A-OKVQA} & \multirow{2}{*}{VQAv2}  \\
     & & & \\
    \midrule
    Instruction Only  & 18.81 & 11.01 & 27.85 \\
    \multirow{2}{*}{\shortstack[c]{Instruction+ \\ Captions}} & \multirow{2}{*}{\underline{39.50}} & \multirow{2}{*}{\underline{50.56}} & \multirow{2}{*}{\underline{59.35}}  \\
     & & & \\
    \multirow{2}{*}{\shortstack[c]{Instruction+ \\ QA Pairs}} & \multirow{2}{*}{33.27} & \multirow{2}{*}{34.67} & \multirow{2}{*}{40.61}  \\
     & & & \\
    \multirow{2}{*}{\shortstack[c]{Instruction+ \\ Captions+QA Pairs}} & \multirow{2}{*}{\textbf{54.57}} & \multirow{2}{*}{\textbf{53.87}} & \multirow{2}{*}{\textbf{67.96}}  \\
     & & & \\
    \bottomrule
  \end{tabular}
  \caption{The effect of incorporating captions and QA pairs in Answer Generator prompt design.}
  \label{table:4}
\end{table}

\section{Conclusion}
\label{sec:Conclusion}
GC-KBVQA is a new and efficient variant of the zero-shot VQA framework that eliminates the reliance on end-to-end training or in-context examples. An innovative four-stage architecture integrates question-based precise grounding for visual content extraction, question-aware caption generation, and a robust content-oriented caption filtering mechanism to effectively generate and prune auxiliary information provided to the answer predictor. The incorporation of a grounding block addresses counting challenges, enhancing the framework's capability. Several illustrative examples are included in the supplementary material. Our approach demonstrates strong zero-shot KB-VQA performance with significantly fewer parameters and no additional training, showcasing its scalability and robustness. This highlights the importance of modular architectures and effective prompting. 

\section{Limitations}
As illustrated in the supplementary material, our framework has difficulty inferring temporal context from visual cues like clothing or architecture when explicit time indicators are absent. Additionally, the Grounding DINO module struggles with questions requiring information from distant, unlinked image regions, revealing challenges similar to those in patch-based methods. Our method has difficulty with tasks involving approximation or estimation of numerical values or ranges, such as age, capacity, cost, weight, or occupancy. These challenging cases require reasoning beyond our current framework. Future work will focus on incorporating temporal reasoning, enhancing global image understanding, and improving numerical reasoning capabilities.

\bibliography{main}

\newpage

\appendix

\onecolumn
\section{Example Appendix}
\label{sec:appendix}

This section presents a detailed qualitative analysis of our proposed framework, structured into two parts. In the first part, we showcase a diverse range of sample cases where our framework successfully generates accurate responses, followed by an analysis of various question types where our model fails to produce correct answers. The second part provides a comparative evaluation between our GC-KBVQA and Img2LLM, one of the most prominent works in this domain. Specifically, we highlight cases where Img2LLM, as reported in their paper, encounters failures, and contrast them with the performance of our approach. This analysis underscores the strengths and limitations of our framework in relation to existing state-of-the-art methods.

\noindent\rule{\textwidth}{0.5pt}

\noindent
\textbf{Question:} Which number birthday is probably being celebrated?

\noindent
\textbf{GT:} \textcolor{magenta}{thirty}/\textcolor{magenta}{30th}

\begin{minipage}{0.3\textwidth} 
\noindent
\includegraphics[width=\textwidth]{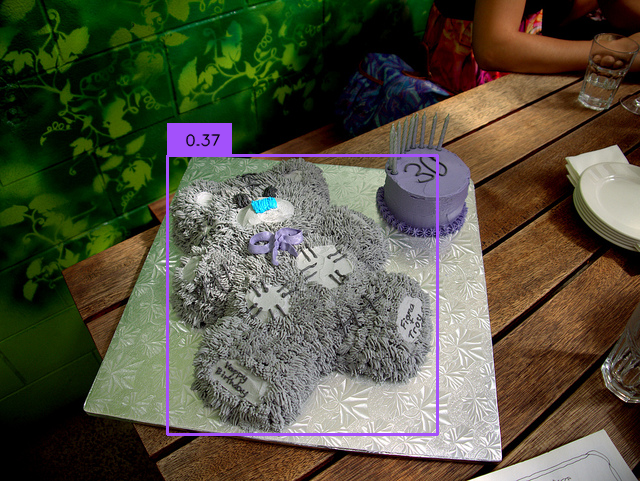} 
\end{minipage}%
\hspace{0.02\textwidth}
\begin{minipage}{0.6\textwidth} 
    \textbf{KeyWords:} number birthday
    \vspace{0.3cm}
    
    \textbf{Question Concept:} most common birthday
    \vspace{0.3cm}
    
    \textbf{Caption1:} The image suggests a \textcolor{cyan}{30th birthday} celebration, as there is a cake shaped like a gray teddy bear on top of a silver tray.

    \textbf{Caption2:}  Celebrating the most common birthday with a teddy bear cake that's as sweet as the day itself.

    \textbf{Caption3:} The image seems to represent a \textcolor{cyan}{30th birthday}.
    
    
\end{minipage}

\noindent
\textbf{QA1:} What birthday is most likely depicted in the image? \textcolor{orange}{Thirty}

\noindent
\textbf{QA2:} What is on top of the silver tray? Cake

\noindent
\vspace{0.3cm}
\textbf{Predicted Answer:} \textcolor{green}{Thirty}

\noindent\rule{\textwidth}{0.5pt}

\noindent
\textbf{Question:} What is the most popular flavor of this kind of pastry?

\noindent
\textbf{GT:} \textcolor{magenta}{icing}/\textcolor{magenta}{frosting}

\begin{minipage}{0.3\textwidth} 
\noindent
\includegraphics[width=0.9\textwidth]{Images/2.png} 
\end{minipage}%
\hspace{0.02\textwidth}
\begin{minipage}{0.6\textwidth} 
    \textbf{KeyWords:} white substance, cupcakes
    \vspace{0.3cm}
    
    \textbf{Question Concept:} white substance on top of the cupcakes
    \vspace{0.3cm}
    
    \textbf{Caption1:} The white substance on top of the cupcakes in this image is \textcolor{cyan}{frosting}.
    
    \textbf{Caption2:} A dusting of powdered sugar on top of the cupcakes.

    \textbf{Caption3:} The white substance on top of the cupcakes in the image appears to be \textcolor{cyan}{frosting} or \textcolor{cyan}{icing}, which is a common decoration used to enhance the appearance and flavor of cupcakes.
    
    
\end{minipage}

\noindent
\textbf{QA1:} What is the white substance on top of the cupcakes? \textcolor{orange}{Frosting}

\noindent
\textbf{QA2:} What is added to the cupcakes to enhance their appearance and flavor? \textcolor{orange}{Icing}

\noindent
\vspace{0.3cm}
\textbf{Predicted Answer:} \textcolor{green}{Frosting}

\noindent\rule{\textwidth}{0.5pt}

\noindent
\textbf{Question:} The thing on the animal to the left's head is similar to what is on the head of what else?

\noindent
\textbf{GT:} \textcolor{magenta}{bull}/\textcolor{magenta}{sheep}/\textcolor{magenta}{horns}/\textcolor{magenta}{ram}

\begin{minipage}{0.3\textwidth} 
\noindent
\includegraphics[width=\textwidth]{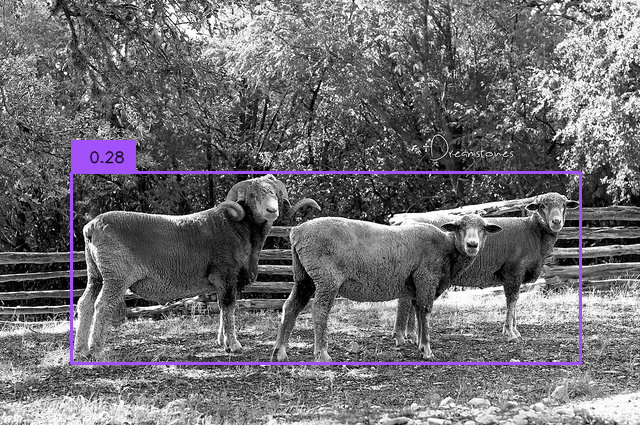} 
\end{minipage}%
\hspace{0.02\textwidth}
\begin{minipage}{0.6\textwidth} 
    \textbf{KeyWords:} head, animal
    \vspace{0.3cm}
    
    \textbf{Question Concept:} similarity between two animals' head features
    \vspace{0.3cm}
    
    \textbf{Caption1:} They both have \textcolor{cyan}{horns}.
    
    \textbf{Caption2:}  Two heads, one \textcolor{cyan}{horn}, a striking similarity in this black and white snapshot of rural life.

    \textbf{Caption3:} The similarity between the two animals' head features in the image is that they both have \textcolor{cyan}{horns}.
    
    
\end{minipage}

\noindent
\textbf{QA1:} What do the two animals in the image have in common? \textcolor{orange}{Horns}

\noindent
\textbf{QA2:} What feature of the two animals' heads is similar in the image? \textcolor{orange}{Horns}

\noindent
\vspace{0.3cm}
\textbf{Predicted Answer:} \textcolor{green}{Horns}

\noindent\rule{\textwidth}{0.5pt}

\noindent
\textbf{Question:} What is the person on the left doing with their body?

\noindent
\textbf{GT:} \textcolor{magenta}{surfing}/\textcolor{magenta}{crouching}/\textcolor{magenta}{bending}/

\begin{minipage}{0.3\textwidth} 
\noindent
\includegraphics[width=\textwidth]{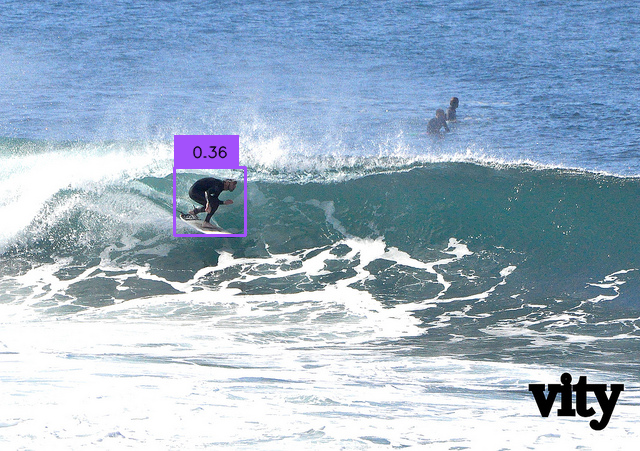} 
\end{minipage}%
\hspace{0.02\textwidth}
\begin{minipage}{0.6\textwidth} 
    \textbf{KeyWords:} body, person, left
    \vspace{0.3cm}
    
    \textbf{Question Concept:} action or posture of the person on the left
    \vspace{0.3cm}
    
    \textbf{Caption1:} The person on the left in the image is \textcolor{cyan}{crouching} down while riding a wave on a surfboard.
    
    \textbf{Caption2:} A man \textcolor{cyan}{surfing} on a wave in the ocean.

    \textbf{Caption3:}  They are positioned in a \textcolor{cyan}{crouching} or squatting stance, with their knees bent and their body leaning forward as they ride the wave.
    
    
\end{minipage}

\noindent
\textbf{QA1:} What is the person on the left in the image doing? Riding

\noindent
\textbf{QA2:} What is the person's stance while riding the wave? \textcolor{orange}{Crouching}

\noindent
\vspace{0.3cm}
\textbf{Predicted Answer:} \textcolor{green}{Crouching}

\noindent\rule{\textwidth}{0.5pt}

\noindent
\textbf{Question:} What country do these planes belong to?

\noindent
\textbf{GT:} \textcolor{magenta}{united states}/\textcolor{magenta}{usa}/\textcolor{magenta}{america}

\begin{minipage}{0.3\textwidth} 
\noindent
\includegraphics[width=\textwidth]{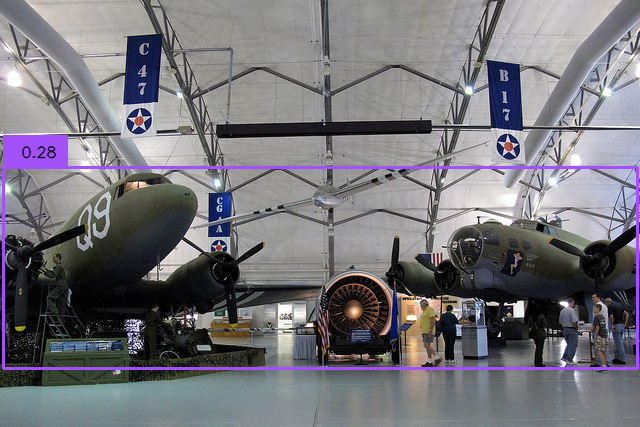} 
\end{minipage}%
\hspace{0.02\textwidth}
\begin{minipage}{0.6\textwidth} 
    \textbf{KeyWords:} planes, country
    \vspace{0.3cm}
    
    \textbf{Question Concept:} nationality of the planes
    \vspace{0.3cm}
    
    \textbf{Caption1:} The two airplanes in the hangar are of different nationalities, with one being a German plane and the other being an \textcolor{cyan}{American} plane.
    
    \textbf{Caption2:} Amidst the hustle and bustle of aviation history, these vintage aircraft proudly display their \textcolor{cyan}{American} heritage.

    \textbf{Caption3:} The markings and insignia on the aircraft suggest they are from the \textcolor{cyan}{United States}, as the camouflage pattern and the style of the aircraft are characteristic of \textcolor{cyan}{American} military aircraft from that period.
    
    
\end{minipage}

\noindent
\textbf{QA1:} What are the nationalities of the two airplanes? \textcolor{orange}{American}

\noindent
\textbf{QA2:} What is characteristic of American military aircraft from that period? Camouflage pattern and style

\noindent
\vspace{0.3cm}
\textbf{Predicted Answer:} \textcolor{green}{America}

\noindent\rule{\textwidth}{0.5pt}

\noindent
\textbf{Question:} What are the orange vehicles for?

\noindent
\textbf{GT:} \textcolor{magenta}{service airplanes}/\textcolor{magenta}{transport}/\textcolor{magenta}{guide planes}/\textcolor{magenta}{maintenance}

\begin{minipage}{0.3\textwidth} 
\noindent
\includegraphics[width=\textwidth]{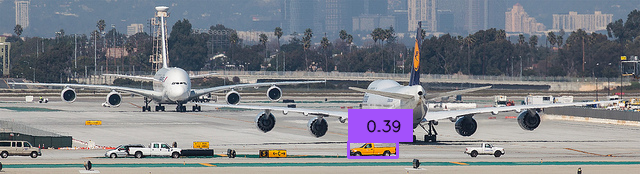} 
\end{minipage}%
\hspace{0.02\textwidth}
\begin{minipage}{0.6\textwidth} 
    \textbf{KeyWords:} orange vehicles
    \vspace{0.3cm}
    
    \textbf{Question Concept:} purpose or function of orange vehicles
    \vspace{0.3cm}
    
    \textbf{Caption1:} The purpose or function of the orange vehicles in this image is to provide \textcolor{cyan}{transportation} services, such as taxi or ride-sharing services.
    
    \textbf{Caption2:} The presence of an orange truck and a yellow pickup truck on the road suggests that these vehicles are being used to \textcolor{cyan}{transport} people from one location to another.

    \textbf{Caption3:} The orange vehicles in the image are used for \textcolor{cyan}{transporting} passengers
    
    
\end{minipage}

\noindent
\textbf{QA1:} What are the orange vehicles in the image used for? \textcolor{cyan}{Transportation} services

\noindent
\textbf{QA2:} What type of service do the orange vehicles offer? ride-sharing

\noindent
\vspace{0.3cm}
\textbf{Predicted Answer:} \textcolor{green}{Transport}

\noindent\rule{\textwidth}{0.5pt}

\noindent
\textbf{Question:} Why is his arm up in the air?

\noindent
\textbf{GT:} \textcolor{magenta}{hitting ball}/\textcolor{magenta}{hitting}/\textcolor{magenta}{swinging}/\textcolor{magenta}{serving}/\textcolor{magenta}{returning serve}/

\begin{minipage}{0.3\textwidth} 
\noindent
\includegraphics[height=130pt,width=\textwidth]{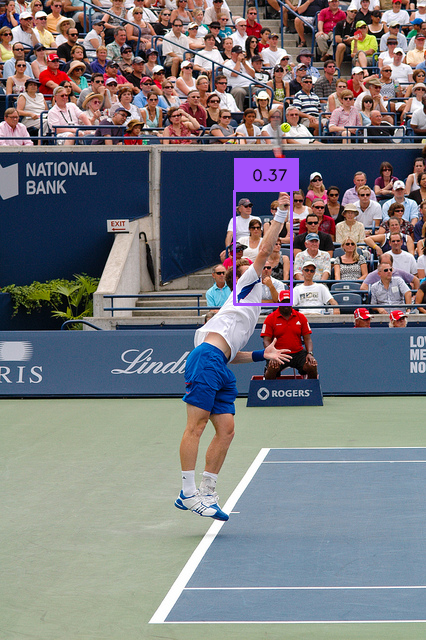} 
\end{minipage}%
\hspace{0.02\textwidth}
\begin{minipage}{0.6\textwidth} 
    \textbf{KeyWords:} arm, air
    \vspace{0.3cm}
    
    \textbf{Question Concept:} reason behind someone's arm being raised in the air
    \vspace{0.3cm}
    
    \textbf{Caption1:} The man's arm is raised in the air as he \textcolor{cyan}{swings} his tennis racket to hit the ball during a game of tennis.
    
    \textbf{Caption2:} He is likely trying to \textcolor{cyan}{hit} the ball with precision and power, aiming to score a point or gain an advantage over his opponent.

    \textbf{Caption3:} In the image, a person is seen raising their arm in the air, which is a common gesture in sports, particularly in tennis.
    
    
\end{minipage}

\noindent
\textbf{QA1:} What is the man's arm raised in the air as he swings? Tennis racket

\noindent
\textbf{QA2:} What is the person in the image likely trying to do? Score

\noindent
\vspace{0.3cm}
\textbf{Predicted Answer:} \textcolor{green}{Hitting}
  \\ \noindent\rule{\textwidth}{0.5pt}

\noindent
\textbf{Question:} What type of structure is being used?

\noindent
\textbf{GT:} \textcolor{magenta}{cement}/\textcolor{magenta}{concrete}/\textcolor{magenta}{barrier}/\textcolor{magenta}{concrete barrier}

\begin{minipage}{0.3\textwidth} 
\noindent
\includegraphics[height=130pt,width=\textwidth]{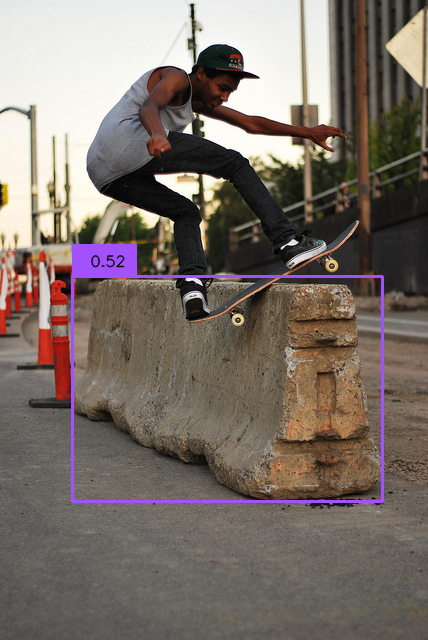} 
\end{minipage}%
\hspace{0.02\textwidth}
\begin{minipage}{0.6\textwidth} 
    \textbf{KeyWords:} structure
    \vspace{0.3cm}
    
    \textbf{Question Concept:} type of structure being used
    \vspace{0.3cm}
    
    \textbf{Caption1:} Skateboarding on a makeshift ramp made of urban elements.
    
    \textbf{Caption2:} In the image, a person is riding a skateboard on top of a \textcolor{cyan}{concrete barrier}.

    \textbf{Caption3:} This particular barrier is designed to be a vertical structure, with a series of horizontal slabs that are stacked on top of each other to create a solid, stable wall.
    
    
\end{minipage}

\noindent
\textbf{QA1:} What is the person riding on top of in the image? \textcolor{orange}{Barrier}

\noindent
\textbf{QA2:} What is the barrier designed to be? Vertical structure.

\noindent
\vspace{0.3cm}
\textbf{Predicted Answer:} \textcolor{green}{Concrete}
  \\ \noindent\rule{\textwidth}{0.5pt}

\noindent
\textbf{Question:} Which shape outer walls do the persons sit in for the best view of the airport?

\noindent
\textbf{GT:} \textcolor{magenta}{round}/\textcolor{magenta}{circular}

\begin{minipage}{0.3\textwidth} 
\noindent
\includegraphics[width=\textwidth]{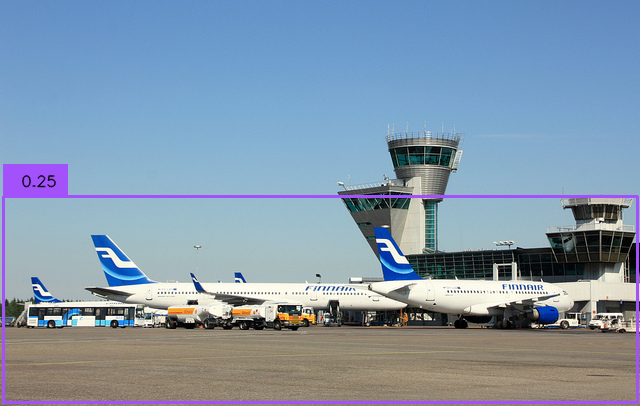} 
\end{minipage}%
\hspace{0.02\textwidth}
\begin{minipage}{0.6\textwidth} 
    \textbf{KeyWords:} outer walls, airport
    \vspace{0.3cm}
    
    \textbf{Question Concept:} shape of the outer walls
    \vspace{0.3cm}
    
    \textbf{Caption1:} The shape of the outer walls where people can sit to get the best view of the airport in this image is \textcolor{cyan}{circular}.
    
    \textbf{Caption2:} Airport Observation Deck: Where the sky meets the runway, and the journey begins.

    \textbf{Caption3:} The outer walls where people can sit to get the best view of the airport in the image are shaped like triangles.
    
    
\end{minipage}

\noindent
\textbf{QA1:} What is the shape of the outer walls in the image? \textcolor{orange}{Circular}

\noindent
\textbf{QA2:} Where can people sit to watch the airport? Outer wall

\noindent
\vspace{0.3cm}
\textbf{Predicted Answer:} \textcolor{green}{Circular}
  \\ \noindent\rule{\textwidth}{0.5pt}

\noindent
\textbf{Question:} What trick is the man performing?

\noindent
\textbf{GT:} \textcolor{magenta}{rail grind}/\textcolor{magenta}{railing}/\textcolor{magenta}{grind}

\begin{minipage}{0.3\textwidth} 
\noindent
\includegraphics[height=130pt, width=\textwidth]{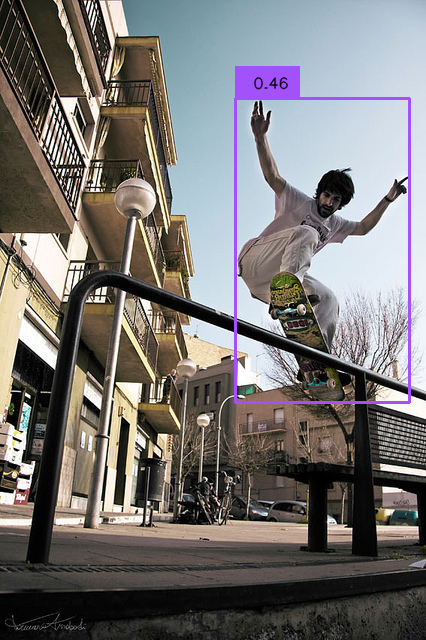} 
\end{minipage}%
\hspace{0.02\textwidth}
\begin{minipage}{0.6\textwidth} 
    \textbf{KeyWords:} trick, man
    \vspace{0.3cm}
    
    \textbf{Question Concept:} type of trick
    \vspace{0.3cm}
    
    \textbf{Caption1:} The man in the image is performing a skateboarding technique called "\textcolor{cyan}{rail grinding}," where he is riding his skateboard along the edge of a metal railing.
    
    \textbf{Caption2:} Performing this skateboarding technique requires precise balance, control, and skill to execute, as the rider must maintain contact with the railing while maneuvering the skateboard.

    \textbf{Caption3:} Some popular skateboarding tricks include the kickflip, heelflip, and 360 flip.
    
    
\end{minipage}

\noindent
\textbf{QA1:} What is the man in the image doing? \textcolor{orange}{Rail grinding}

\noindent
\textbf{QA2:} What type of skateboarding technique requires precise balance, control, and skill to execute? \textcolor{orange}{Rail grinding}

\noindent
\vspace{0.3cm}
\textbf{Predicted Answer:} \textcolor{green}{Rail grind}
  \\ \noindent\rule{\textwidth}{0.5pt}

\noindent
\textbf{Question:} What type of building appears in the background?

\noindent
\textbf{GT:} \textcolor{magenta}{government}/\textcolor{magenta}{royal building}/\textcolor{magenta}{courthouse}/\textcolor{magenta}{palace}

\begin{minipage}{0.3\textwidth} 
\noindent
\includegraphics[height=130pt, width=\textwidth]{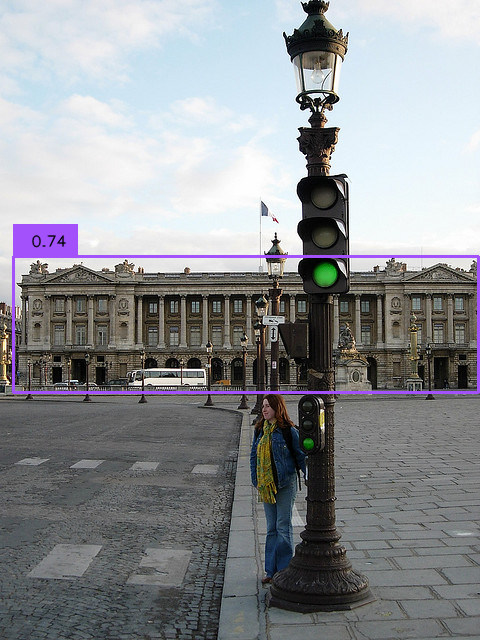} 
\end{minipage}%
\hspace{0.02\textwidth}
\begin{minipage}{0.6\textwidth} 
    \textbf{KeyWords:} building, background
    \vspace{0.3cm}
    
    \textbf{Question Concept:} type of building in the background
    \vspace{0.3cm}
    
    \textbf{Caption1:} In the background of the image, there is a large, ornate building that appears to be a \textcolor{cyan}{palace} or a \textcolor{cyan}{government} building.
    
    \textbf{Caption2:} A grand, historic building with a commanding presence, standing tall amidst the urban landscape.

    \textbf{Caption3:} The style of the building in the background suggests a \textcolor{cyan}{government} building or similar official structure.
    
    
\end{minipage}

\noindent
\textbf{QA1:} What type of building is shown in the background of the image? \textcolor{orange}{Palace}

\noindent
\textbf{QA2:} What is the architectural style of the building described?
 Ornate

\noindent
\vspace{0.3cm}
\textbf{Predicted Answer:} \textcolor{green}{Palace}
  \\ \noindent\rule{\textwidth}{0.5pt}

\noindent
\textbf{Question:} Why are the trees without leaves?

\noindent
\textbf{GT:} \textcolor{magenta}{fall season}/\textcolor{magenta}{autumn}/\textcolor{magenta}{it's fall}

\begin{minipage}{0.3\textwidth} 
\noindent
\includegraphics[width=\textwidth]{Images/21.png} 
\end{minipage}%
\hspace{0.02\textwidth}
\begin{minipage}{0.6\textwidth} 
    \textbf{KeyWords:} trees, leaves
    \vspace{0.3cm}
    
    \textbf{Question Concept:} lack of leaves on the trees
    \vspace{0.3cm}
    
    \textbf{Caption1:} There is a bare, leafless judging by its lack of leaves, and is situated in front of a multi-story building.
    
    \textbf{Caption2:} \textcolor{cyan}{Autumn's} touch: The bare trees stand tall against the cityscape, their leaves fallen to the ground, signaling the season's change.

    \textbf{Caption3:} There are no leaves on the trees.
    
    
\end{minipage}

\noindent
\textbf{QA1:} What is the state of the trees in Autumn's touch? Bare

\noindent
\textbf{QA2:} What can be attributed to the lack of leaves on the trees? \textcolor{orange}{Autumn}

\noindent
\vspace{0.3cm}
\textbf{Predicted Answer:} \textcolor{green}{Autumn}
  \\ \noindent\rule{\textwidth}{0.5pt}

\noindent
\textbf{Question:} What is the man in green's job?

\noindent
\textbf{GT:} \textcolor{magenta}{construction worker}/\textcolor{magenta}{road construction}/\textcolor{magenta}{street maintenance}

\begin{minipage}{0.3\textwidth} 
\noindent
\includegraphics[height=130pt, width=\textwidth]{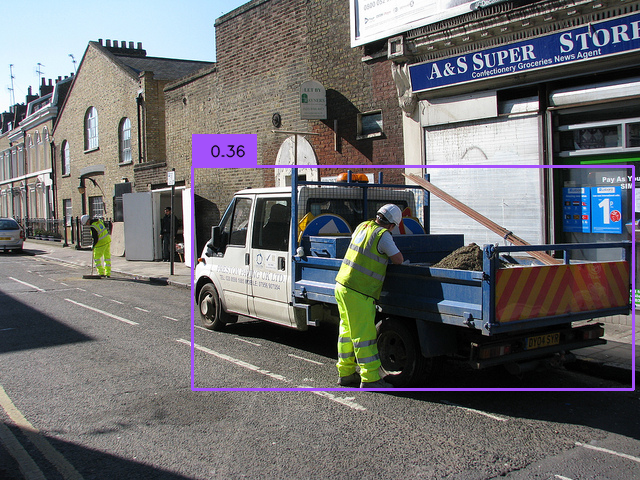} 
\end{minipage}%
\hspace{0.02\textwidth}
\begin{minipage}{0.6\textwidth} 
    \textbf{KeyWords:} green
    \vspace{0.3cm}
    
    \textbf{Question Concept:} occupation of a person in green
    \vspace{0.3cm}
    
    \textbf{Caption1:} The man in green is likely a \textcolor{cyan}{construction worker}, as he is standing next to a dump truck and appears to be loading materials into the back of the truck.
    
    \textbf{Caption2:} A \textcolor{cyan}{construction worker} in action, ensuring the safety of the site with his high-visibility attire.

    \textbf{Caption3:} The person described as the man in green in the image is likely a \textcolor{cyan}{construction worker}.
    
    
\end{minipage}

\noindent
\textbf{QA1:} Who is the man in green likely to be? \textcolor{orange}{Construction worker}

\noindent
\textbf{QA2:} What is the man in green doing in the image? Loading

\noindent
\vspace{0.3cm}
\textbf{Predicted Answer:} \textcolor{green}{Construction worker}
  \\ \noindent\rule{\textwidth}{0.5pt}

\noindent
\textbf{Question:} What is the possible hazard faced by the animal?

\noindent
\textbf{GT:} \textcolor{magenta}{being stuck}/\textcolor{magenta}{getting stuck}/\textcolor{magenta}{face stuck}/\textcolor{magenta}{glass}

\begin{minipage}{0.3\textwidth} 
\noindent
\includegraphics[height=130pt, width=\textwidth]{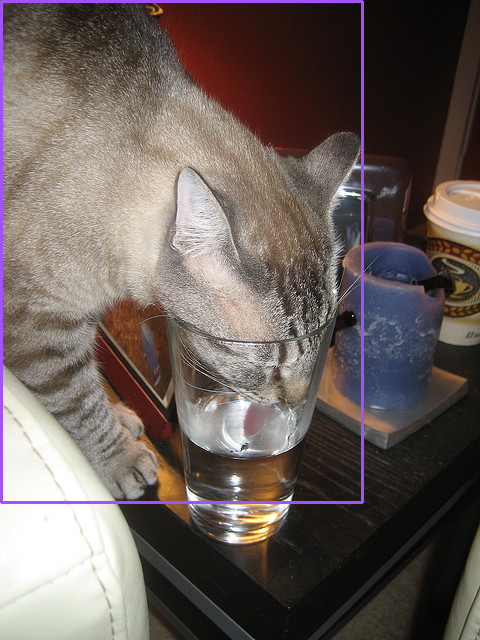} 
\end{minipage}%
\hspace{0.02\textwidth}
\begin{minipage}{0.6\textwidth} 
    \textbf{KeyWords:} possible hazard, animal
    \vspace{0.3cm}
    
    \textbf{Question Concept:} animal hazard risk
    \vspace{0.3cm}
    
    \textbf{Caption1:} The image depicts a cat drinking water from a \textcolor{cyan}{glass}, which poses a potential hazard risk to the cat.
    
    \textbf{Caption2:} When curiosity meets hydration: A cautionary tale of a feline's thirsty adventure

    \textbf{Caption3:} There are no obvious hazards in the image that would be of concern for the cat.
    
    
\end{minipage}

\noindent
\textbf{QA1:} What is the cat doing in the image? Drinking

\noindent
\textbf{QA2:} What is the potential hazard risk to the cat? \textcolor{orange}{Glass}

\noindent
\vspace{0.3cm}
\textbf{Predicted Answer:} \textcolor{green}{Glass}
  \\ \noindent\rule{\textwidth}{0.5pt}

\noindent
\textbf{Question:} What breed is the brown cow?

\noindent
\textbf{GT:} \textcolor{magenta}{hereford}

\begin{minipage}{0.3\textwidth} 
\noindent
\includegraphics[width=\textwidth]{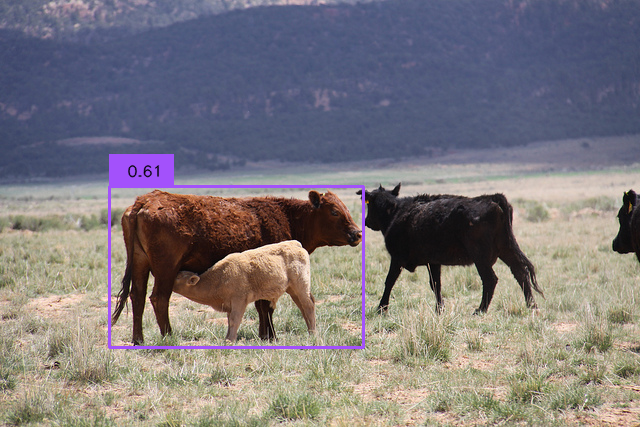} 
\end{minipage}%
\hspace{0.02\textwidth}
\begin{minipage}{0.6\textwidth} 
    \textbf{KeyWords:} brown cow, breed
    \vspace{0.3cm}
    
    \textbf{Question Concept:} breed of a brown cow
    \vspace{0.3cm}
    
    \textbf{Caption1:} The breed of the brown cow in this image is a Jersey cow. 
    
    \textbf{Caption2:} A brown cow, possibly a \textcolor{cyan}{Hereford} or Angus, enjoys a moment of nourishment in the open field.

    \textbf{Caption3:} The breed of the brown cow in the image is a \textcolor{cyan}{Hereford} cow.
    
    
\end{minipage}

\noindent
\textbf{QA1:} What is the breed of the brown cow in the image? \textcolor{orange}{Hereford}

\noindent
\textbf{QA2:} What breed of cow is possibly enjoying the moment of nourishment in the open field? \textcolor{orange}{Hereford}

\noindent
\vspace{0.3cm}
\textbf{Predicted Answer:} \textcolor{green}{Hereford}
  \\ \noindent\rule{\textwidth}{0.5pt}

\noindent
\textbf{Question:} What is the dog doing with the bovines here?

\noindent
\textbf{GT:} \textcolor{magenta}{herding}

\begin{minipage}{0.3\textwidth} 
\noindent
\includegraphics[width=\textwidth]{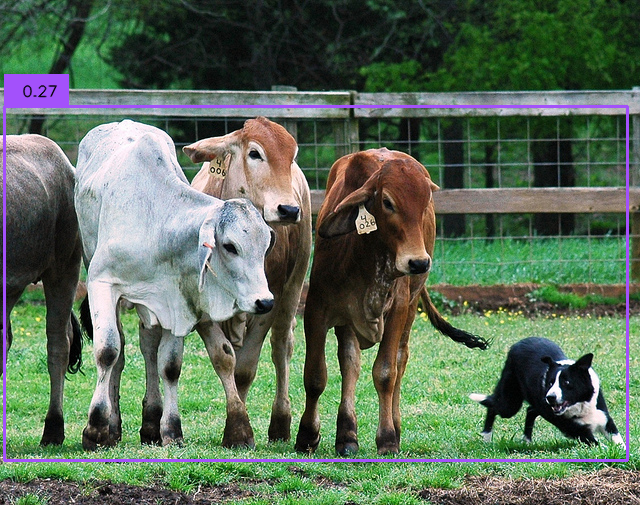} 
\end{minipage}%
\hspace{0.02\textwidth}
\begin{minipage}{0.6\textwidth} 
    \textbf{KeyWords:} bovines, dog
    \vspace{0.3cm}
    
    \textbf{Question Concept:} dog is interacting with cows
    \vspace{0.3cm}
    
    \textbf{Caption1:} The dog is approaching the cows and seems to be curious about them.
    
    \textbf{Caption2:} The cows appear to be calm and accepting of the dog's presence, not perceiving it as a threat.

    \textbf{Caption3:} The dog's behavior suggests that it may be \textcolor{cyan}{herding} the cows, as dogs are often used for this purpose in farming and ranching operations.
    
    
\end{minipage}

\noindent
\textbf{QA1:} What is the dog's behavior suggesting? \textcolor{orange}{Herding}

\noindent
\textbf{QA2:} What is the dog's presence perceived as by the cows? Not a threat

\noindent
\vspace{0.3cm}
\textbf{Predicted Answer:} \textcolor{green}{Herding}
  \\ \noindent\rule{\textwidth}{0.5pt}

\noindent
\textbf{Question:} What is the name of the lip piercing that the girl in the foreground has?

\noindent
\textbf{GT:} \textcolor{magenta}{vertical labret}/\textcolor{magenta}{lip ring}

\begin{minipage}{0.3\textwidth} 
\noindent
\includegraphics[width=\textwidth]{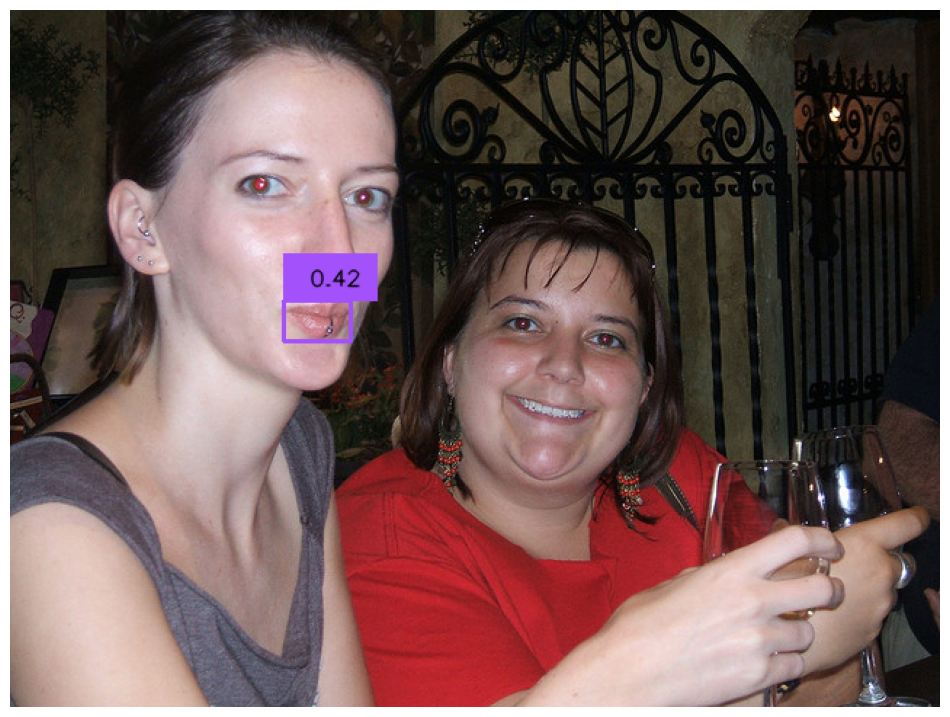} 
\end{minipage}%
\hspace{0.02\textwidth}
\begin{minipage}{0.6\textwidth} 
    \textbf{KeyWords:} lip piercing
    \vspace{0.3cm}
    
    \textbf{Question Concept:} type of lip piercing
    \vspace{0.3cm}
    
    \textbf{Caption1:} The girl in the foreground of the image is wearing a lip piercing that appears to be a \textcolor{cyan}{vertical labret} piercing.
    
    \textbf{Caption2:} She's got a playful and edgy style with her lip piercing, adding a touch of rebellion to her smile.

    \textbf{Caption3:} The girl is wearing a \textcolor{cyan}{lip ring} that adds to her unique style.
    
    
\end{minipage}

\noindent
\textbf{QA1:} What type of body modification is featured in the image? \textcolor{orange}{Lip ring}

\noindent
\textbf{QA2:} What does the lip piercing add to the girl's smile? Rebellion

\noindent
\vspace{0.3cm}
\textbf{Predicted Answer:} \textcolor{green}{Vertical labret}
  \\ \noindent\rule{\textwidth}{0.5pt}

\noindent
\textbf{Question:} What is the name of the player's hair style?

\noindent
\textbf{GT:} \textcolor{magenta}{ponytail}/\textcolor{magenta}{braid}

\begin{minipage}{0.3\textwidth} 
\noindent
\includegraphics[height=130pt, width=\textwidth]{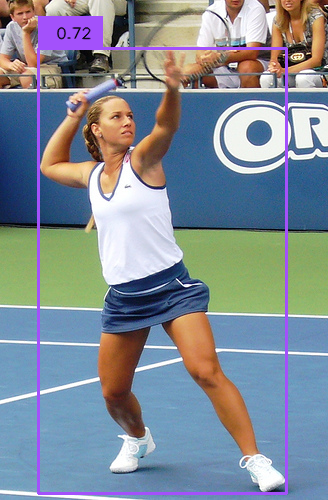} 
\end{minipage}%
\hspace{0.02\textwidth}
\begin{minipage}{0.6\textwidth} 
    \textbf{KeyWords:} hair style, player
    \vspace{0.3cm}
    
    \textbf{Question Concept:} name of a player's hairstyle
    \vspace{0.3cm}
    
    \textbf{Caption1:} The player in the image has a hairstyle that appears to be a \textcolor{cyan}{braid}.
    
    \textbf{Caption2:} The player is light-skinned with dark brown hair pulled back in a \textcolor{cyan}{ponytail}.

    \textbf{Caption3:} Serving up a powerful swing with a hairstyle that's all about style and strength.
    
    
\end{minipage}

\noindent
\textbf{QA1:} What is the player's hairstyle in the image? \textcolor{orange}{braid}

\noindent
\textbf{QA2:} What is the player doing in the image? Serving

\noindent
\vspace{0.3cm}
\textbf{Predicted Answer:} \textcolor{green}{Braid}
  \\ \noindent\rule{\textwidth}{0.5pt}

\noindent
\textbf{Question:} What part of the body is the orange vegetable good for?

\noindent
\textbf{GT:} \textcolor{magenta}{eyes}

\begin{minipage}{0.3\textwidth} 
\noindent
\includegraphics[width=\textwidth]{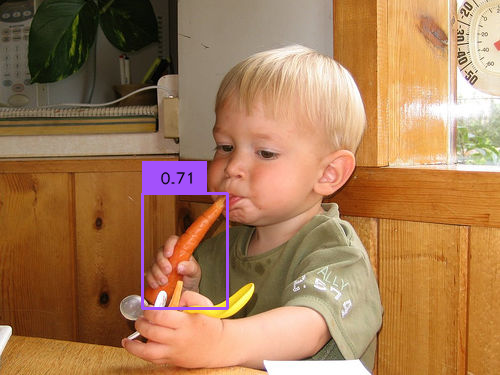} 
\end{minipage}%
\hspace{0.02\textwidth}
\begin{minipage}{0.6\textwidth} 
    \textbf{KeyWords:} orange vegetable
    \vspace{0.3cm}
    
    \textbf{Question Concept:} health benefits of a specific vegetable
    \vspace{0.3cm}
    
    \textbf{Caption1:} Carrots are rich in beta-carotene, which the body converts into vitamin A, supporting good \textcolor{cyan}{vision}, immune function, and healthy skin.
    
    \textbf{Caption2:} Carrots are a rich source of vitamins and minerals, such as vitamin A, vitamin K, and potassium.

    \textbf{Caption3:} Carrots are a good source of potassium, which helps regulate blood pressure, and fiber, which promotes digestive health.
    
    
\end{minipage}

\noindent
\textbf{QA1:} What is a good source of potassium? Carrots

\noindent
\textbf{QA2:} What is a source of fiber? Carrots

\noindent
\vspace{0.3cm}
\textbf{Predicted Answer:} \textcolor{green}{Eyes}
  \\ \noindent\rule{\textwidth}{0.5pt}

\noindent
\textbf{Question:} What type of fuel does this vehicle take?

\noindent
\textbf{GT:} \textcolor{magenta}{diesel}/\textcolor{magenta}{gasoline}

\begin{minipage}{0.3\textwidth} 
\noindent
\includegraphics[width=\textwidth]{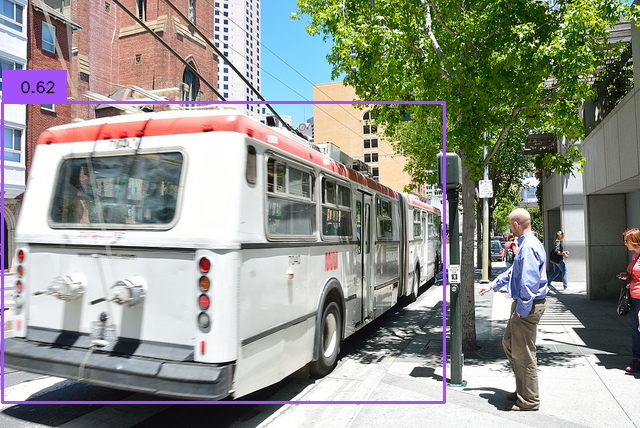} 
\end{minipage}%
\hspace{0.02\textwidth}
\begin{minipage}{0.6\textwidth} 
    \textbf{KeyWords:} fuel, vehicle
    \vspace{0.3cm}
    
    \textbf{Question Concept:} type of fuel used by a vehicle
    \vspace{0.3cm}
    
    \textbf{Caption1:} This vehicle is powered by a blend of modern technology and traditional fuel, as it runs on a combination of electricity and \textcolor{cyan}{diesel}. 
    
    \textbf{Caption2:} The image shows a bus using \textcolor{cyan}{diesel} fuel, which is common for public transportation due to its cost-efficiency.

    \textbf{Caption3:} The bus in the image is a standard urban bus, typically used for public transportation.
    
    
\end{minipage}

\noindent
\textbf{QA1:} What type of fuel does the vehicle run on? \textcolor{orange}{Diesel}

\noindent
\textbf{QA2:} What kind of bus is it? Urban bus

\noindent
\vspace{0.3cm}
\textbf{Predicted Answer:} \textcolor{green}{Diesel}
  \\ \noindent\rule{\textwidth}{0.5pt}

\noindent
\textbf{Question:} Where could i find this?

\noindent
\textbf{GT:} \textcolor{magenta}{cafeteria}/\textcolor{magenta}{restaurant}

\begin{minipage}{0.3\textwidth} 
\noindent
\includegraphics[width=\textwidth]{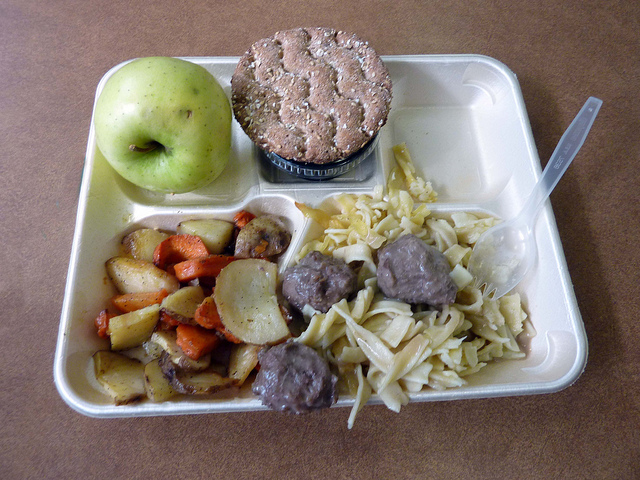} 
\end{minipage}%
\hspace{0.02\textwidth}
\begin{minipage}{0.6\textwidth} 
    \textbf{KeyWords:} -
    \vspace{0.3cm}
    
    \textbf{Question Concept:} source of this
    \vspace{0.3cm}
    
    \textbf{Caption1:} A plastic tray filled with a variety of foods, including an apple, a muffin, and pasta.
    
    \textbf{Caption2:} The image features a plastic lunch tray filled with a variety of food items.

    \textbf{Caption3:} The tray is divided into four compartments containing a green apple, a round dark-brown cookie, roasted potatoes and carrots, and pasta with meatballs.
    
    
\end{minipage}

\noindent
\textbf{QA1:} What is being displayed in the image? Tray

\noindent
\textbf{QA2:} What type of container is being used to hold the food items? Plastic

\noindent
\vspace{0.3cm}
\textbf{Predicted Answer:} \textcolor{green}{Cafeteria}
  \\ \noindent\rule{\textwidth}{0.5pt}

\noindent
\textbf{Question:} What type of food are these people eating?

\noindent
\textbf{GT:} \textcolor{magenta}{chinese}/\textcolor{magenta}{asian}/\textcolor{magenta}{soup}/\textcolor{magenta}{curry}

\begin{minipage}{0.3\textwidth} 
\noindent
\includegraphics[width=\textwidth]{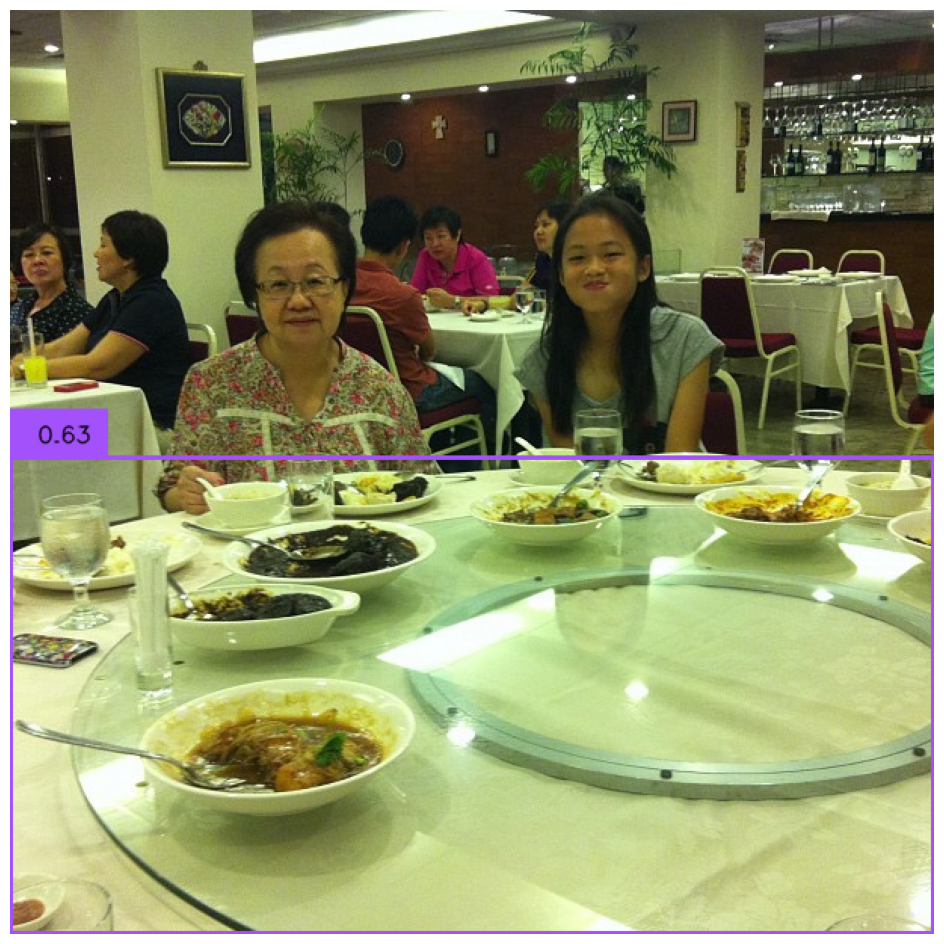} 
\end{minipage}%
\hspace{0.02\textwidth}
\begin{minipage}{0.6\textwidth} 
    \textbf{KeyWords:} food
    \vspace{0.3cm}
    
    \textbf{Question Concept:} type of food
    \vspace{0.3cm}
    
    \textbf{Caption1:} The food on the table includes various types of dishes, such as bowls of rice, noodles, and vegetables.
    
    \textbf{Caption2:} They appear to be enjoying a meal together, as they are smiling and posing for a photo.

    \textbf{Caption3:} In the image, there are several dishes that appear to be part of a meal.
    
    
\end{minipage}

\noindent
\textbf{QA1:} What are they doing in the image? Posing

\noindent
\textbf{QA2:} What is visible on the table? Dishes

\noindent
\vspace{0.3cm}
\textbf{Predicted Answer:} \textcolor{green}{Asian}
  \\ \noindent\rule{\textwidth}{0.5pt}

\noindent
\textbf{Question:} When was this airline founded?

\noindent
\textbf{GT:} \textcolor{magenta}{1996}/\textcolor{magenta}{nineties}/\textcolor{magenta}{february 29 1996}

\begin{minipage}{0.3\textwidth} 
\noindent
\includegraphics[width=\textwidth]{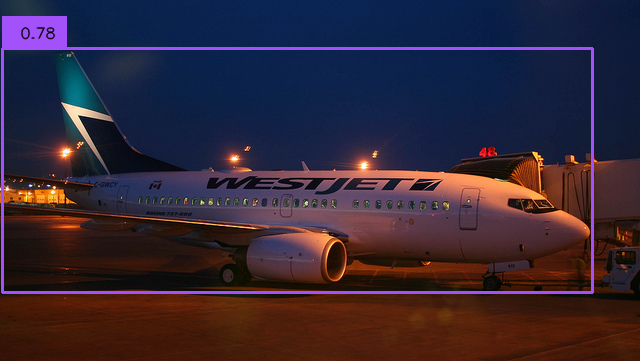} 
\end{minipage}%
\hspace{0.02\textwidth}
\begin{minipage}{0.6\textwidth} 
    \textbf{KeyWords:} airline
    \vspace{0.3cm}
    
    \textbf{Question Concept:} founding date of the airline
    \vspace{0.3cm}
    
    \textbf{Caption1:} Founded in \textcolor{cyan}{1996}, WestJet continues to soar with its fleet of modern aircraft, ready to take passengers on their next journey.
    
    \textbf{Caption2:} Westjet was founded in \textcolor{cyan}{1996}.

    \textbf{Caption3:} The airline in the image is WestJet, which was founded in \textcolor{cyan}{1996}.
    
    
\end{minipage}

\noindent
\textbf{QA1:} When was WestJet founded? \textcolor{cyan}{1996}

\noindent
\textbf{QA2:} What airline is depicted in the image? WestJet

\noindent
\vspace{0.3cm}
\textbf{Predicted Answer:} \textcolor{green}{1996}
  \\ \noindent\rule{\textwidth}{0.5pt}

\noindent
\textbf{Question:} What time period are these people from?

\noindent
\textbf{GT:} \textcolor{magenta}{medieval}/\textcolor{magenta}{mid evil}/\textcolor{magenta}{1600s}/\textcolor{magenta}{civil war}

\begin{minipage}{0.3\textwidth} 
\noindent
\includegraphics[width=\textwidth]{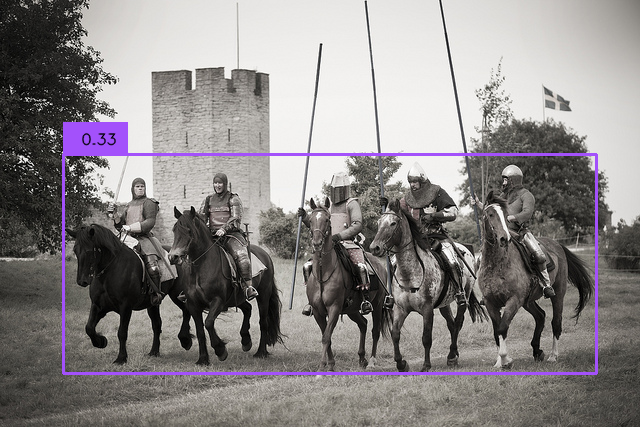} 
\end{minipage}%
\hspace{0.02\textwidth}
\begin{minipage}{0.6\textwidth} 
    \textbf{KeyWords:} time period, people
    \vspace{0.3cm}
    
    \textbf{Question Concept:} time period associated with people
    \vspace{0.3cm}
    
    \textbf{Caption1:} The time period associated with the group of people in this image is \textcolor{cyan}{medieval}, as they are riding horses and wearing \textcolor{cyan}{medieval}-style armor.
    
    \textbf{Caption2:} This suggests that they are part of a \textcolor{cyan}{medieval}-themed event or reenactment, where participants dress up and engage in activities reminiscent of the Middle Ages.

    \textbf{Caption3:} The time period associated with the group of people in the image is difficult to determine, as the image is black and white.
    
    
\end{minipage}

\noindent
\textbf{QA1:} What is the time period associated with the group of people in the image? \textcolor{orange}{Medieval}

\noindent
\textbf{QA2:} What is the image likely to be? Reenactment

\noindent
\vspace{0.3cm}
\textbf{Predicted Answer:} \textcolor{green}{medieval}
  \\ \noindent\rule{\textwidth}{0.5pt}

\noindent
\textbf{Question:} What age group does the human facing the camera fall into? 

\noindent
\textbf{GT:} \textcolor{magenta}{toddler}/\textcolor{magenta}{2 3}

\begin{minipage}{0.3\textwidth} 
\noindent
\includegraphics[width=\textwidth]{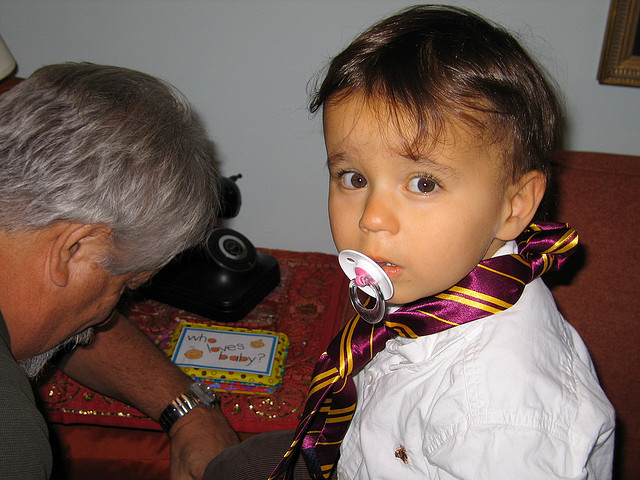} 
\end{minipage}%
\hspace{0.02\textwidth}
\begin{minipage}{0.6\textwidth} 
    \textbf{KeyWords:} age group, camera
    \vspace{0.3cm}
    
    \textbf{Question Concept:} age group of the person facing the camera
    \vspace{0.3cm}
    
    \textbf{Caption1:} The individual facing the camera in this image is likely a young child or \textcolor{cyan}{toddler} based on their apparent age group.
    
    \textbf{Caption2:} A young child, possibly a \textcolor{cyan}{toddler}, is captured in this candid moment.

    \textbf{Caption3:} The person facing the camera in the image is a young boy, likely between the ages of 2 and 5 years old.
    
    
\end{minipage}

\noindent
\textbf{QA1:} Who is facing the camera in the image? Boy

\noindent
\textbf{QA2:} What is the age group of the person in the image? \textcolor{orange}{Toddler}

\noindent
\vspace{0.3cm}
\textbf{Predicted Answer:} \textcolor{green}{Toddler}
  \\ \noindent\rule{\textwidth}{0.5pt}

\noindent
\textbf{Question:} What state is this child most likely in?

\noindent
\textbf{GT:} \textcolor{magenta}{massachusetts}/\textcolor{magenta}{boston}/\textcolor{magenta}{united states}

\begin{minipage}{0.3\textwidth} 
\noindent
\includegraphics[width=\textwidth]{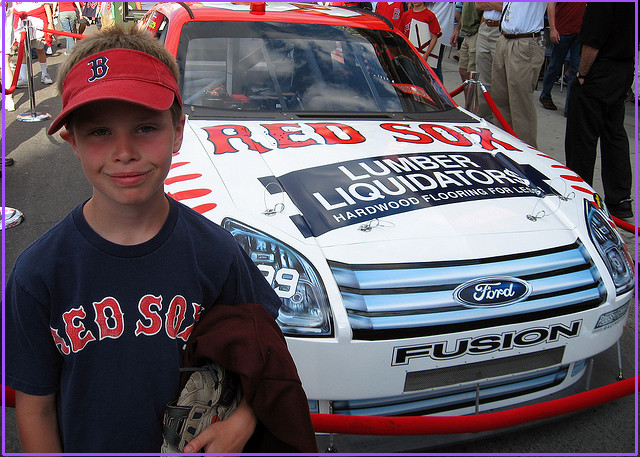} 
\end{minipage}%
\hspace{0.02\textwidth}
\begin{minipage}{0.6\textwidth} 
    \textbf{KeyWords:} state, child
    \vspace{0.3cm}
    
    \textbf{Question Concept:} likely state of a child
    \vspace{0.3cm}
    
    \textbf{Caption1:}  In the image, a young boy is standing in front of a race car with a Red Sox logo on it.
    
    \textbf{Caption2:} He is wearing a baseball cap and holding a baseball glove, which suggests that he may be a fan of the \textcolor{cyan}{Boston} Red Sox team.

    \textbf{Caption3:} The child appears to be a boy, possibly in the age range of 5 to 10 years old in \textcolor{cyan}{Massachusetts}.
    
    
\end{minipage}

\noindent
\textbf{QA1:} What is on the race car in the image? Red Sox logo

\noindent
\textbf{QA2:} What is the age range of the boy in the image? 5 to 10

\noindent
\vspace{0.3cm}
\textbf{Predicted Answer:} \textcolor{green}{Massachusetts}
  \\ \noindent\rule{\textwidth}{0.5pt}

\noindent
\textbf{Question:} What year was the background restaurant founded?

\noindent
\textbf{GT:} \textcolor{magenta}{1955}/\textcolor{magenta}{1951}/\textcolor{magenta}{1952}

\begin{minipage}{0.3\textwidth} 
\noindent
\includegraphics[width=\textwidth]{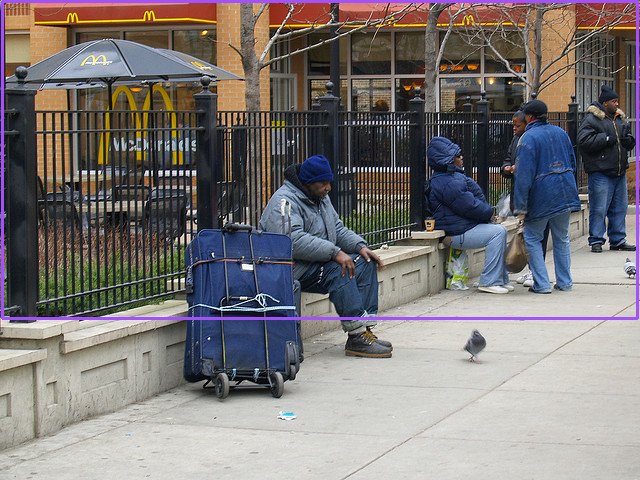} 
\end{minipage}%
\hspace{0.02\textwidth}
\begin{minipage}{0.6\textwidth} 
    \textbf{KeyWords:} background restaurant
    \vspace{0.3cm}
    
    \textbf{Question Concept:} founding year of a background restaurant
    \vspace{0.3cm}
    
    \textbf{Caption1:} The founding year of the background restaurant in this image is \textcolor{cyan}{1955}.
    
    \textbf{Caption2:} Founded in \textcolor{cyan}{1955}, this McDonald's location has been serving up delicious meals to the community for over six decades.

    \textbf{Caption3:}  The background restaurant in the image is McDonald's, which was founded in \textcolor{cyan}{1955} by Ray Kroc.
    
    
\end{minipage}

\noindent
\textbf{QA1:} When was the background restaurant founded? \textcolor{orange}{1955}

\noindent
\textbf{QA2:} What restaurant is depicted in the image? McDonald's

\noindent
\vspace{0.3cm}
\textbf{Predicted Answer:} \textcolor{green}{1955}
  \\ \noindent\rule{\textwidth}{0.5pt}

\noindent
\textbf{Question:} When was the activity shown first invented?

\noindent
\textbf{GT:} \textcolor{magenta}{1922}/\textcolor{magenta}{1920s}/\textcolor{magenta}{1950}/\textcolor{magenta}{1985}

\begin{minipage}{0.3\textwidth} 
\noindent
\includegraphics[width=\textwidth]{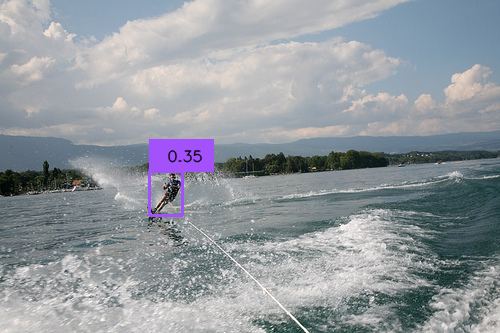} 
\end{minipage}%
\hspace{0.02\textwidth}
\begin{minipage}{0.6\textwidth} 
    \textbf{KeyWords:} activity
    \vspace{0.3cm}
    
    \textbf{Question Concept:} invention date of a specific activity
    \vspace{0.3cm}
    
    \textbf{Caption1:} Water skiing was invented in \textcolor{cyan}{1922}.
    
    \textbf{Caption2:} The exhilarating sport of water skiing traces its roots back to the \textcolor{cyan}{1920s} when pioneers like Ralph Samuelson and his wooden skis revolutionized the world of water sports.

    \textbf{Caption3:} Samuelson's invention quickly gained popularity, and by the 1930s, water skiing was a popular recreational activity.
    
    
\end{minipage}

\noindent
\textbf{QA1:} When was water skiing invented? \textcolor{orange}{1922}

\noindent
\textbf{QA2:} Who revolutionized the world of water sports with his wooden skis? Ralph Samuelson

\noindent
\vspace{0.3cm}
\textbf{Predicted Answer:} \textcolor{green}{1922}
  \\ \noindent\rule{\textwidth}{0.5pt}

\noindent
\textbf{Question:} How heavy is this object?

\noindent
\textbf{GT:} \textcolor{magenta}{100 lbs}/\textcolor{magenta}{very heavy}/\textcolor{magenta}{20 pounds}/\textcolor{magenta}{500 lbs}

\begin{minipage}{0.3\textwidth} 
\noindent
\includegraphics[width=\textwidth]{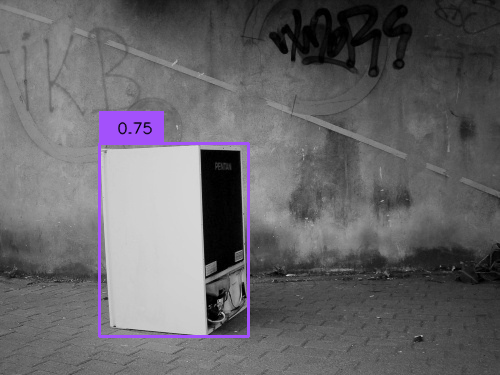} 
\end{minipage}%
\hspace{0.02\textwidth}
\begin{minipage}{0.6\textwidth} 
    \textbf{KeyWords:} object
    \vspace{0.3cm}
    
    \textbf{Question Concept:} weight of the object 
    \vspace{0.3cm}
    
    \textbf{Caption1:} Based on the refrigerator's appearance in the image, it appears to be a standard upright model with an estimated weight of \textcolor{cyan}{100 lbs}.
    
    \textbf{Caption2:} Based on the appearance of the refrigerator in the image it seems to be a standard upright model with \textcolor{cyan}{100 lbs} weight.

    \textbf{Caption3:} The refrigerator is rectangular, predominantly white, with a dark panel on the front and weight of \textcolor{cyan}{100 lbs}.
    
    
\end{minipage}

\noindent
\textbf{QA1:} How much does the refrigerator weigh? \textcolor{orange}{100 lbs}

\noindent
\textbf{QA2:} What is the shape of the refrigerator? Rectangular

\noindent
\vspace{0.3cm}
\textbf{Predicted Answer:} \textcolor{green}{100 lbs}
  \\ \noindent\rule{\textwidth}{0.5pt}

\noindent
\textbf{Question:} What is the tallest of this type of building in the world?

\noindent
\textbf{GT:} \textcolor{magenta}{big ben}/\textcolor{magenta}{1972 feet high}/\textcolor{magenta}{burj khalifa}/\textcolor{magenta}{ulm minster}

\begin{minipage}{0.3\textwidth} 
\noindent
\includegraphics[width=\textwidth]{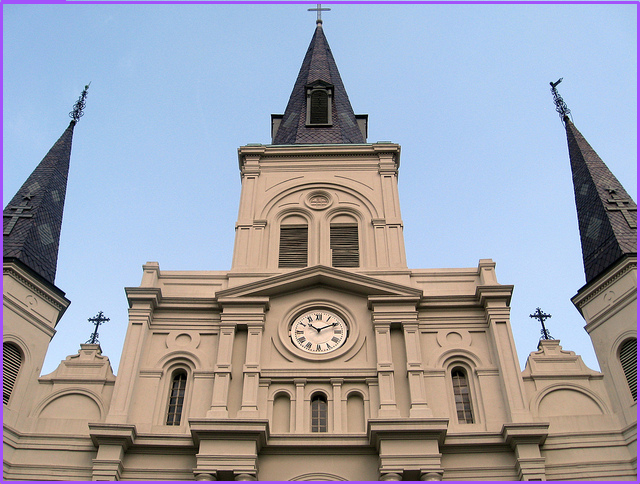} 
\end{minipage}%
\hspace{0.02\textwidth}
\begin{minipage}{0.6\textwidth} 
    \textbf{KeyWords:} building
    \vspace{0.3cm}
    
    \textbf{Question Concept:} tallest building of a specific type in the world
    \vspace{0.3cm}
    
    \textbf{Caption1:} The tallest building of its type in the world can be seen in this image, which is a church with three steeples is \textcolor{cyan}{Burj Khalifa}.
    
    \textbf{Caption2:} The \textcolor{cyan}{Burj Khalifa} is not only the tallest building in the world but also holds several other records, such as the highest occupied floor, highest outdoor observation deck, elevator with the longest travel distance, and the tallest freestanding structure in the world.

    \textbf{Caption3:} The majestic clock tower of the world's tallest church, standing tall against the sky.
    
    
\end{minipage}

\noindent
\textbf{QA1:} What is seen in the image? Church

\noindent
\textbf{QA2:} What is the Burj Khalifa? Building

\noindent
\vspace{0.3cm}
\textbf{Predicted Answer:} \textcolor{green}{Burj Khalifa}
  \\ \noindent\rule{\textwidth}{0.5pt}

\noindent
\textbf{Question:} What is the metal pipe device in old homes that heats them called?

\noindent
\textbf{GT:} \textcolor{magenta}{radiator}/\textcolor{magenta}{heater}/\textcolor{magenta}{furnace}

\begin{minipage}{0.3\textwidth} 
\noindent
\includegraphics[width=\textwidth]{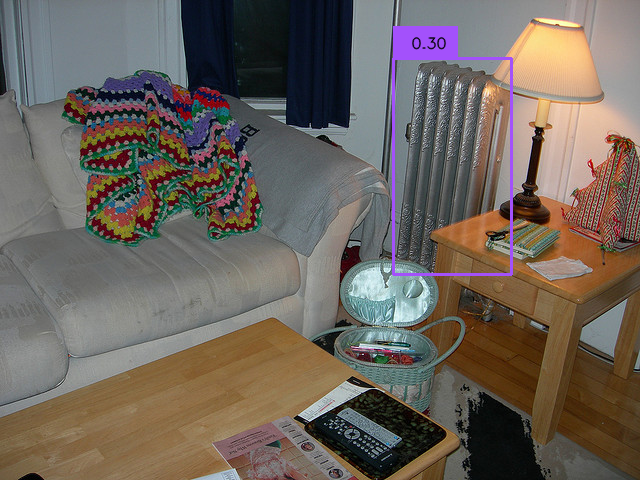} 
\end{minipage}%
\hspace{0.02\textwidth}
\begin{minipage}{0.6\textwidth} 
    \textbf{KeyWords:} metal pipe device
    \vspace{0.3cm}
    
    \textbf{Question Concept:} name of the Heating pipe
    \vspace{0.3cm}
    
    \textbf{Caption1:} On the end table, a vintage-style \textcolor{cyan}{radiator} (silver color), appearing to be covered with a light-colored insulating material.
    
    \textbf{Caption2:} A vintage \textcolor{cyan}{radiator} stands guard in the corner, a silent sentinel of warmth in this cozy corner of the home.

    \textbf{Caption3:} The metal pipe device in the image is an old-fashioned \textcolor{cyan}{radiator}, which is a type of heating system commonly found in older homes.
    
    
\end{minipage}

\noindent
\textbf{QA1:} What is the device shown in the image? \textcolor{orange}{Radiator}

\noindent
\textbf{QA2:} What type of heating system is commonly found in older homes? \textcolor{orange}{Radiator}

\noindent
\vspace{0.3cm}
\textbf{Predicted Answer:} \textcolor{green}{Radiator}
  \\ \noindent\rule{\textwidth}{0.5pt}

\noindent
\textbf{Question:} What piece of ski equipment do all of these people have?

\noindent
\textbf{GT:} \textcolor{magenta}{skis}/\textcolor{magenta}{poles}/\textcolor{magenta}{ski stick}

\begin{minipage}{0.3\textwidth} 
\noindent
\includegraphics[width=\textwidth]{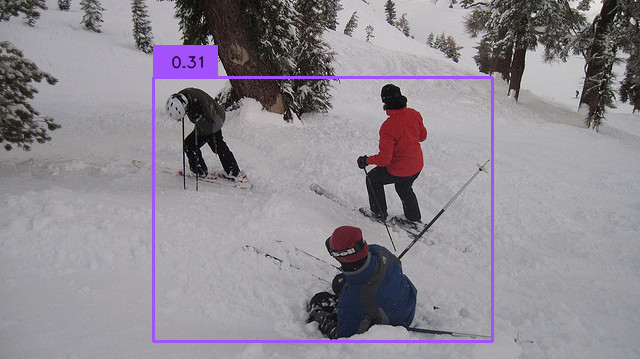} 
\end{minipage}%
\hspace{0.02\textwidth}
\begin{minipage}{0.6\textwidth} 
    \textbf{KeyWords:} ski equipment
    \vspace{0.3cm}
    
    \textbf{Question Concept:} a common ski equipment item everyone has.
    \vspace{0.3cm}
    
    \textbf{Caption1:} People in the image are equipped with \textcolor{cyan}{ski poles}.
    
    \textbf{Caption2:} Skiing together in the snow, united by the common thread of their \textcolor{cyan}{skis}, creating a vibrant winter scene.

    \textbf{Caption3:} Everyone in the image is holding a pair of \textcolor{cyan}{ski poles}, a standard accessory for skiing.
    
    
\end{minipage}

\noindent
\textbf{QA1:} What is the common piece of ski equipment that all the people have in the image? \textcolor{orange}{Ski poles}

\noindent
\textbf{QA2:} What unites the people in the winter scene? \textcolor{orange}{Skis}

\noindent
\vspace{0.3cm}
\textbf{Predicted Answer:} \textcolor{green}{Poles}
  \\ \noindent\rule{\textwidth}{0.5pt}

\noindent
\textbf{Question:} What climate does the yellow fruit grow in?

\noindent
\textbf{GT:} \textcolor{magenta}{tropical}/\textcolor{magenta}{warm}

\begin{minipage}{0.3\textwidth} 
\noindent
\includegraphics[width=\textwidth]{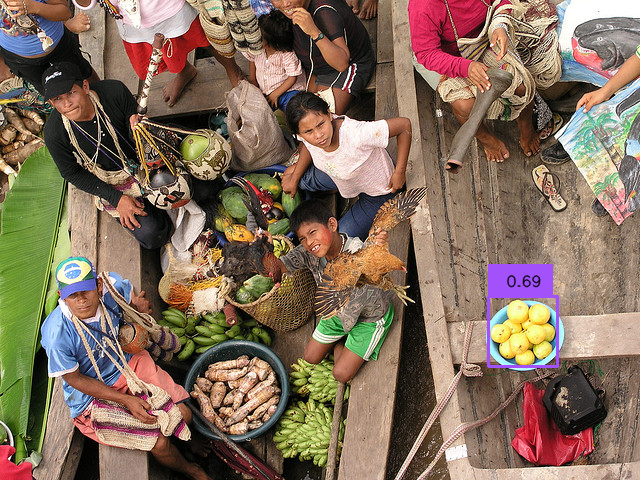} 
\end{minipage}%
\hspace{0.02\textwidth}
\begin{minipage}{0.6\textwidth} 
    \textbf{KeyWords:} yellow fruit, climate
    \vspace{0.3cm}
    
    \textbf{Question Concept:} climate where a specific yellow fruit grows
    \vspace{0.3cm}
    
    \textbf{Caption1:} The yellow fruit shown in this image grows in a  \textcolor{cyan}{tropical} environment, providing optimal conditions for its cultivation.
    
    \textbf{Caption2:} A \textcolor{cyan}{tropical} climate where the sun shines brightly and the lemons grow in abundance.

    \textbf{Caption3:} The climate in these regions is characterized by high temperatures, low humidity, and moderate rainfall, making it conducive to the growth of citrus trees and the production of lemons.
    
    
\end{minipage}

\noindent
\textbf{QA1:} What kind of climate is characterized by high temperatures, low humidity, and moderate rainfall? \textcolor{orange}{Tropical}

\noindent
\textbf{QA2:} What type of trees grow in a tropical climate where the sun shines brightly and the lemons grow in abundance? Citrus

\noindent
\vspace{0.3cm}
\textbf{Predicted Answer:} \textcolor{green}{Tropical}
  \\ \noindent\rule{\textwidth}{0.5pt}

\noindent
\textbf{Question:} What can i make with this?

\noindent
\textbf{GT:} \textcolor{magenta}{smoothie}/\textcolor{magenta}{breakfast}

\begin{minipage}{0.3\textwidth} 
\noindent
\includegraphics[width=\textwidth]{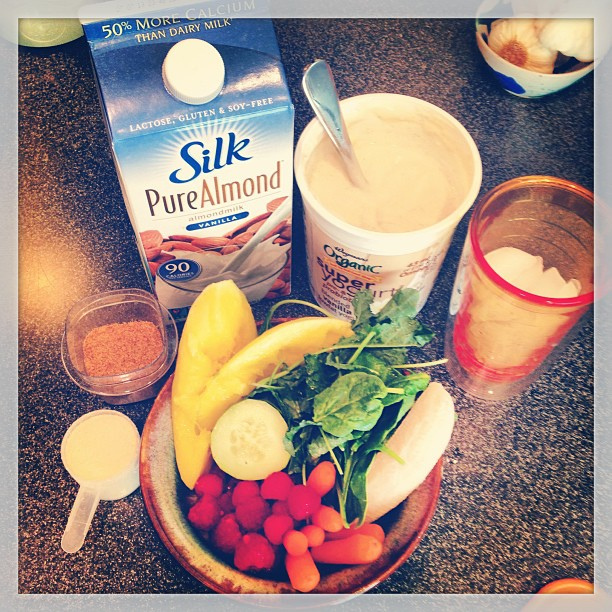} 
\end{minipage}%
\hspace{0.02\textwidth}
\begin{minipage}{0.6\textwidth} 
    \textbf{KeyWords:} -
    \vspace{0.3cm}
    
    \textbf{Question Concept:} What can I create using these
    \vspace{0.3cm}
    
    \textbf{Caption1:} A variety of ingredients laid out on a dark-gray countertop, seemingly prepared for a \textcolor{cyan}{smoothie} or similar blended drink.
    
    \textbf{Caption2:} They can be blended into \textcolor{cyan}{smoothies}, sliced and added to salads or sandwiches, or used as a topping for yogurt or oatmeal.

    \textbf{Caption3:} A bowl of vibrant vegetables and fruits, ready to be transformed into a nutritious dish.
    
    
\end{minipage}

\noindent
\textbf{QA1:} What can be blended with fresh ingredients to make a smoothie? Vegetables and fruits

\noindent
\textbf{QA2:} What drink can the ingredients be used to make? \textcolor{orange}{Smoothie}

\noindent
\vspace{0.3cm}
\textbf{Predicted Answer:} \textcolor{green}{Smoothie}
  \\ \noindent\rule{\textwidth}{0.5pt}

\noindent
\textbf{Question:} Why aren't the animals eating meat?

\noindent
\textbf{GT:} \textcolor{magenta}{they eat grass}/\textcolor{magenta}{they are grass fed cattle}/\textcolor{magenta}{they are cows}/\textcolor{magenta}{they are herbivores}

\begin{minipage}{0.3\textwidth} 
\noindent
\includegraphics[width=\textwidth]{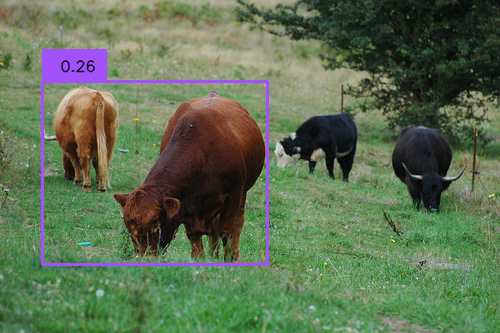} 
\end{minipage}%
\hspace{0.02\textwidth}
\begin{minipage}{0.6\textwidth} 
    \textbf{KeyWords:} meat, animals
    \vspace{0.3cm}
    
    \textbf{Question Concept:} animals' diet
    \vspace{0.3cm}
    
    \textbf{Caption1:} The cows seem to be enjoying their diet of grass, which is a common food source for cattle and other grazing animals.
    
    \textbf{Caption2:} The presence of both cows and a dog suggests that the area is a farm or a pasture where livestock is raised and cared for.

    \textbf{Caption3:} Both cows are eating the grass, which is a natural part of their diet.
    
    
\end{minipage}

\noindent
\textbf{QA1:} What do the cows eat? Grass

\noindent
\textbf{QA2:} What type of area is suggested by the presence of cows and a dog? Farm

\noindent
\vspace{0.3cm}
\textbf{Predicted Answer:} \textcolor{green}{They are herbivores}
  \\ \noindent\rule{\textwidth}{0.5pt}

\noindent
\textbf{Question:} What is the purpose of the white circle?

\noindent
\textbf{GT:} \textcolor{magenta}{time telling}/\textcolor{magenta}{clock tells time}/\textcolor{magenta}{tell time}/\textcolor{magenta}{clock}

\begin{minipage}{0.3\textwidth} 
\noindent
\includegraphics[width=\textwidth]{Images/56.png} 
\end{minipage}%
\hspace{0.02\textwidth}
\begin{minipage}{0.6\textwidth} 
    \textbf{KeyWords:} white circle
    \vspace{0.3cm}
    
    \textbf{Question Concept:} function of the white circle
    \vspace{0.3cm}
    
    \textbf{Caption1:} The white circle in the image is a \textcolor{cyan}{clock}, which serves as a \textcolor{cyan}{time-telling} device.
    
    \textbf{Caption2:} A white circle \textcolor{cyan}{clock} on a concrete wall, serving as a silent sentinel of the passing hours.

    \textbf{Caption3:} A large, round \textcolor{cyan}{clock} face is affixed to a light grayish-tan colored concrete wall of the building.
    
    
\end{minipage}

\noindent
\textbf{QA1:} What is the white circle in the image? \textcolor{orange}{Clock}

\noindent
\textbf{QA2:} What is the white circle clock serving as? \textcolor{orange}{Time-telling} device

\noindent
\vspace{0.3cm}
\textbf{Predicted Answer:} \textcolor{green}{Time telling}
  \\ \noindent\rule{\textwidth}{0.5pt}

\noindent
\textbf{Question:} What's the next thing this player is likely to do?

\noindent
\textbf{GT:} \textcolor{magenta}{swing}/\textcolor{magenta}{batting}/\textcolor{magenta}{run}

\begin{minipage}{0.3\textwidth} 
\noindent
\includegraphics[width=\textwidth]{Images/57.png} 
\end{minipage}%
\hspace{0.02\textwidth}
\begin{minipage}{0.6\textwidth} 
    \textbf{KeyWords:} player
    \vspace{0.3cm}
    
    \textbf{Question Concept:} player's next action
    \vspace{0.3cm}
    
    \textbf{Caption1:} The player is likely to \textcolor{cyan}{swinging} the bat next, attempting to hit the ball, either for a base hit or a home run.
    
    \textbf{Caption2:} Takahashi prepares to \textcolor{cyan}{swing}, ready to make contact with the incoming pitch.

    \textbf{Caption3:} The player is expected to \textcolor{cyan}{swing} the bat and attempt to hit the ball thrown by the opposing team's pitcher.
    
    
\end{minipage}

\noindent
\textbf{QA1:} What is the player's next action likely to involve? \textcolor{orange}{Swinging}

\noindent
\textbf{QA2:} What does Takahashi prepare to do? \textcolor{orange}{Swing}

\noindent
\vspace{0.3cm}
\textbf{Predicted Answer:} \textcolor{green}{Swing}
  \\ \noindent\rule{\textwidth}{0.5pt}

\noindent
\textbf{Question:} Why is he eating this?

\noindent
\textbf{GT:} \textcolor{magenta}{snack}/\textcolor{magenta}{carnival}/\textcolor{magenta}{hungry}

\begin{minipage}{0.3\textwidth} 
\noindent
\includegraphics[width=\textwidth]{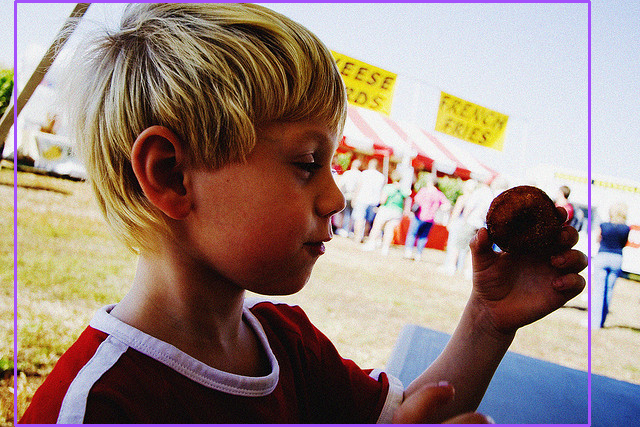} 
\end{minipage}%
\hspace{0.02\textwidth}
\begin{minipage}{0.6\textwidth} 
    \textbf{KeyWords:} eating
    \vspace{0.3cm}
    
    \textbf{Question Concept:} reason behind eating
    \vspace{0.3cm}
    
    \textbf{Caption1:} His action of eating the doughnut may stem from several factors, such as \textcolor{cyan}{hunger}, a desire for a \textcolor{cyan}{snack}, or simply enjoying the taste and texture of the doughnut.
    
    \textbf{Caption2:} The presence of other people nearby suggests that the boy might be sharing the doughnut with others or taking a break from an outdoor activity to indulge in a quick \textcolor{cyan}{snack}.

    \textbf{Caption3:} He might be enjoying a \textcolor{cyan}{snack} during a picnic or outdoor event, or it could be that he is participating in a \textcolor{cyan}{carnival} or fair where food vendors offer various types of treats like doughnuts.
    
    
\end{minipage}

\noindent
\textbf{QA1:} What might be the reason behind the boy's action of eating the doughnut? \textcolor{orange}{Hunger}

\noindent
\textbf{QA2:} Where might the boy be eating the doughnut? \textcolor{orange}{Carnival}

\noindent
\vspace{0.3cm}
\textbf{Predicted Answer:} \textcolor{green}{Snack}
  \\ \noindent\rule{\textwidth}{0.5pt}

\noindent
\textbf{Question:} Does this food taste salty sweet sour or bitter?

\noindent
\textbf{GT:} \textcolor{magenta}{sweet}

\begin{minipage}{0.3\textwidth} 
\noindent
\includegraphics[width=\textwidth]{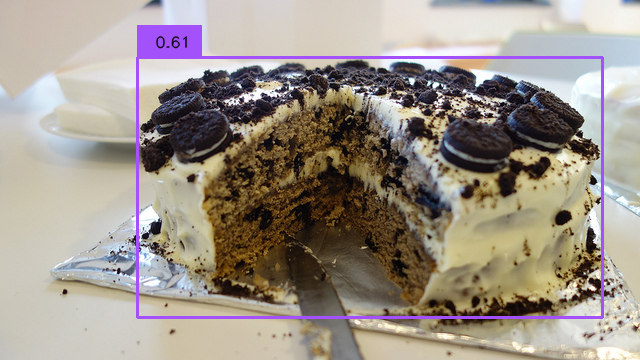} 
\end{minipage}%
\hspace{0.02\textwidth}
\begin{minipage}{0.6\textwidth} 
    \textbf{KeyWords:} salty sweet sour
    \vspace{0.3cm}
    
    \textbf{Question Concept:} taste of a food
    \vspace{0.3cm}
    
    \textbf{Caption1:} Indulge in the rich, chocolatey delight of this Oreo-topped cake, where every bite is a symphony of flavors and textures.
    
    \textbf{Caption2:} The cake appears to be a popular dessert choice, as it is well-liked by those who have tried it.

    \textbf{Caption3:} The chocolate chips would provide a crunchy texture and a slightly bitter taste, complementing the \textcolor{cyan}{sweetness} of the frosting and the cake.
    
    
\end{minipage}

\noindent
\textbf{QA1:} What is the topping of the cake mentioned in the text? Oreo

\noindent
\textbf{QA2:} What would the chocolate chips provide to the cake, according to the text? Crunchy texture

\noindent
\vspace{0.3cm}
\textbf{Predicted Answer:} \textcolor{green}{Sweet}
  \\ \noindent\rule{\textwidth}{0.5pt}

\noindent
\textbf{Question:} What utensil is needed to eat this meal?

\noindent
\textbf{GT:} \textcolor{magenta}{fork}

\begin{minipage}{0.3\textwidth} 
\noindent
\includegraphics[width=\textwidth]{Images/60.png} 
\end{minipage}%
\hspace{0.02\textwidth}
\begin{minipage}{0.6\textwidth} 
    \textbf{KeyWords:} utensil, meal
    \vspace{0.3cm}
    
    \textbf{Question Concept:} necessary utensil for eating this meal
    \vspace{0.3cm} 
    
    \textbf{Caption1:} The utensil needed to eat the meal in this image is a \textcolor{cyan}{fork}.
    
    \textbf{Caption2:} A \textcolor{cyan}{fork} is placed next to the plate with chicken and vegetables, indicating that it is the preferred utensil for eating that particular meal.

    \textbf{Caption3:} A \textcolor{cyan}{fork} and knife are essential for this meal, as the food is served on a plate and likely needs to be cut into smaller portions for convenient eating.
    
    
\end{minipage}

\noindent
\textbf{QA1:} What utensil is necessary for eating a particular meal in the image? \textcolor{orange}{Fork}

\noindent
\textbf{QA2:} What utensils would be necessary for eating a meal with food that is served on a plate and cut into smaller pieces? \textcolor{orange}{Fork} and knife

\noindent
\vspace{0.3cm}
\textbf{Predicted Answer:} \textcolor{green}{Fork}
  \\ \noindent\rule{\textwidth}{0.5pt}

\noindent
\textbf{Question:} What is the colorful object on the man's head used for?

\noindent
\textbf{GT:} \textcolor{magenta}{shade}/\textcolor{magenta}{protection}/\textcolor{magenta}{rain protection}/\textcolor{magenta}{umbrella}

\begin{minipage}{0.3\textwidth} 
\noindent
\includegraphics[width=\textwidth]{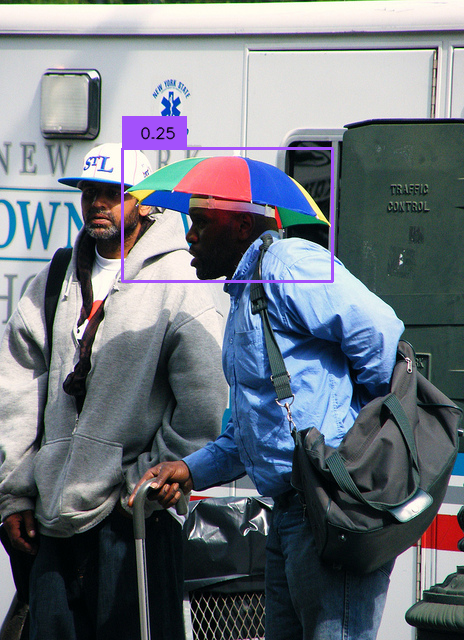} 
\end{minipage}%
\hspace{0.02\textwidth}
\begin{minipage}{0.6\textwidth} 
    \textbf{KeyWords:} colorful object, head
    \vspace{0.3cm}
    
    \textbf{Question Concept:} functionality of the colorful object
    \vspace{0.3cm}
    
    \textbf{Caption1:} The colorful object on the man's head is used to provide \textcolor{cyan}{shade} and \textcolor{cyan}{protection} from the sun.
    
    \textbf{Caption2:} The man is wearing an \textcolor{cyan}{umbrella} hat, which is a type of headwear that features an umbrella-like structure with a colorful design.

    \textbf{Caption3:} The purpose of the colorful object on the man's head in the image is to serve as an \textcolor{cyan}{umbrella}.
    
    
\end{minipage}

\noindent
\textbf{QA1:} What is the purpose of the colorful object on the man's head in this image? \textcolor{orange}{Shade}

\noindent
\textbf{QA2:} What type of headwear is the man wearing? \textcolor{orange}{Umbrella}

\noindent
\vspace{0.3cm}
\textbf{Predicted Answer:} \textcolor{green}{Shade}
  \\ \noindent\rule{\textwidth}{0.5pt}

\noindent
\textbf{Question:} What number comes after the number on the man's jersey when you count to twenty?

\noindent
\textbf{GT:} \textcolor{magenta}{11}/\textcolor{magenta}{eleven}

\begin{minipage}{0.3\textwidth} 
\noindent
\includegraphics[width=\textwidth]{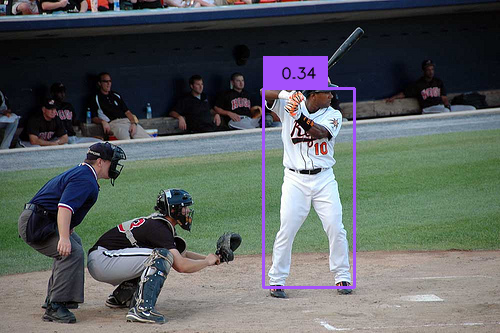} 
\end{minipage}%
\hspace{0.02\textwidth}
\begin{minipage}{0.6\textwidth} 
    \textbf{KeyWords:} number, jersey
    \vspace{0.3cm}
    
    \textbf{Question Concept:} number that comes after the number on the man's jersey 
    \vspace{0.3cm}
    
    \textbf{Caption1:} The number that comes after 10 in this image is \textcolor{cyan}{11}, as the batter is ready to swing at the next pitch.
    
    \textbf{Caption2:} In the image, the number that comes after the number on the man's jersey when counting to 20 is "\textcolor{cyan}{11}".

    \textbf{Caption3:} The batter's uniform is primarily white and orange/peach, and the number "10" is visible on the front of the shirt.
    
    
\end{minipage}

\noindent
\textbf{QA1:} What is the number that comes after 10 in the image? \textcolor{orange}{11}

\noindent
\textbf{QA2:} What is the number that comes after the number on the man's jersey when counting to 20 in the image? \textcolor{orange}{11}

\noindent
\vspace{0.3cm}
\textbf{Predicted Answer:} \textcolor{green}{11}
  \\ \noindent\rule{\textwidth}{0.5pt}

\noindent
\textbf{Question:} In which way are the adults shown here likely related to the child?

\noindent
\textbf{GT:} \textcolor{magenta}{grandparents}/\textcolor{magenta}{parents}

\begin{minipage}{0.3\textwidth} 
\noindent
\includegraphics[width=\textwidth]{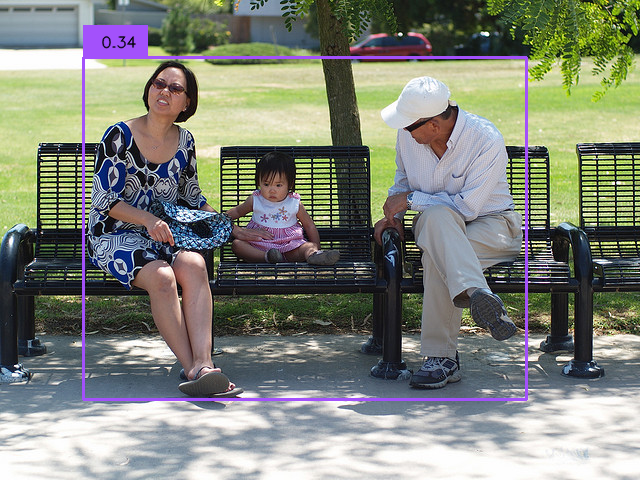} 
\end{minipage}%
\hspace{0.02\textwidth}
\begin{minipage}{0.6\textwidth} 
    \textbf{KeyWords:} adults, child
    \vspace{0.3cm}
    
    \textbf{Question Concept:} relationship between the adults and the child
    \vspace{0.3cm}
    
    \textbf{Caption1:} The relationship between the adults and the child is likely that of \textcolor{cyan}{grandparents} and their grandchild, as they are spending time together in a park setting.
    
    \textbf{Caption2:} The child is wearing a pink dress and appears to be enjoying the company of the adults.

    \textbf{Caption3:} It is likely that the adults are the child's \textcolor{cyan}{parents} or caregivers, and they are spending time with the child in a relaxed and casual setting.
    
    
\end{minipage}

\noindent
\textbf{QA1:} What is the relationship between the adults and the child likely to be? \textcolor{orange}{Grandparents}

\noindent
\textbf{QA2:} What is the child wearing? Dress

\noindent
\vspace{0.3cm}
\textbf{Predicted Answer:} \textcolor{green}{Grandparents}
  \\ \noindent\rule{\textwidth}{0.5pt}

\noindent
\textbf{Question:} The man on the left dressed in a blue uniform has what on his hand?

\noindent
\textbf{GT:} \textcolor{magenta}{glove}/\textcolor{magenta}{mitt}/\textcolor{magenta}{baseball mitt}

\begin{minipage}{0.3\textwidth} 
\noindent
\includegraphics[width=\textwidth]{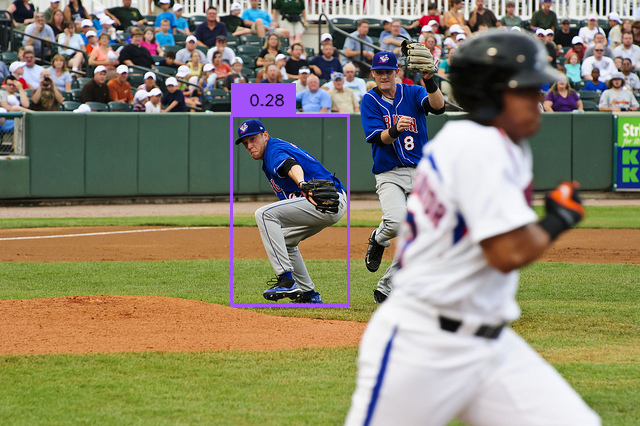} 
\end{minipage}%
\hspace{0.02\textwidth}
\begin{minipage}{0.6\textwidth} 
    \textbf{KeyWords:} left, blue uniform, hand
    \vspace{0.3cm}
    
    \textbf{Question Concept:} object on the man's hand
    \vspace{0.3cm}
    
    \textbf{Caption1:} In the image, the man is holding a baseball \textcolor{cyan}{glove}.
    
    \textbf{Caption2:} The pitcher's \textcolor{cyan}{glove}, a crucial tool in the game of baseball, ready to catch the incoming ball.

    \textbf{Caption3:} The man is crouched slightly forward in an athletic stance, suggesting focus and readiness for action.
    
    
\end{minipage}

\noindent
\textbf{QA1:} : What is the man holding in the image?
\textcolor{orange}{Glove}

\noindent
\textbf{QA2:} What is a crucial tool in the game of baseball? \textcolor{orange}{Glove}

\noindent
\vspace{0.3cm}
\textbf{Predicted Answer:} \textcolor{green}{Glove}

\noindent\rule{\textwidth}{0.5pt}

\newpage

Examples involving counting number of objects.

\vspace{12pt}

\begin{minipage}{0.3\textwidth} 
\noindent
\includegraphics[width=\textwidth]{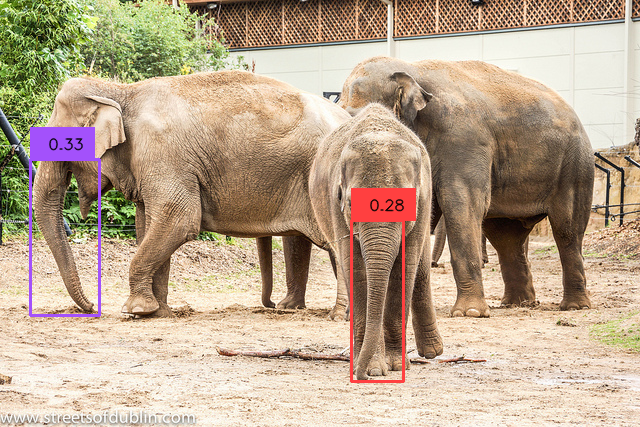} 
\end{minipage}%
\hspace{0.02\textwidth}
\begin{minipage}{0.6\textwidth} 
    \textbf{Question:} How many animal trunks are visible here?
    \vspace{0.3cm}
    
    \textbf{GT:} \textcolor{magenta}{2}
    \vspace{0.3cm}
    
    \textbf{KeyWords:} animal trunks
    \vspace{0.3cm}

    \textbf{Predicted Answer:} \textcolor{green}{2}
    
    
\end{minipage}
  \\ \noindent\rule{\textwidth}{0.5pt}

\begin{minipage}{0.3\textwidth} 
\noindent
\includegraphics[width=\textwidth]{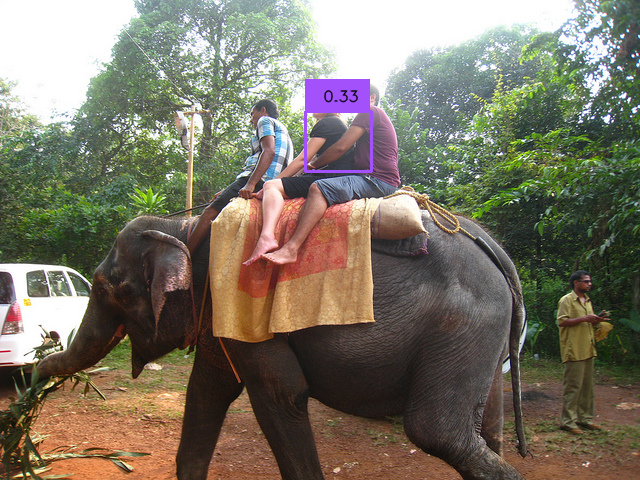} 
\end{minipage}%
\hspace{0.02\textwidth}
\begin{minipage}{0.6\textwidth} 
    \textbf{Question:} How many people have their arms around the woman?
    \vspace{0.3cm}
    
    \textbf{GT:} \textcolor{magenta}{1}
    \vspace{0.3cm}
    
    \textbf{KeyWords:} people, arms, woman
    \vspace{0.3cm}

    \textbf{Predicted Answer:} \textcolor{green}{1}
    
    
\end{minipage}
  \\ \noindent\rule{\textwidth}{0.5pt}

\begin{minipage}{0.3\textwidth} 
\noindent
\includegraphics[width=\textwidth]{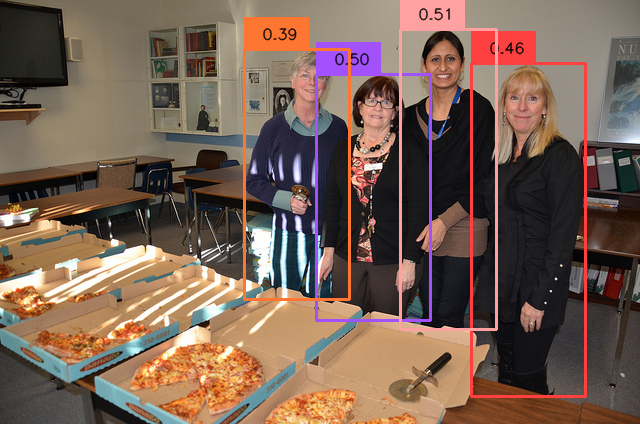} 
\end{minipage}%
\hspace{0.02\textwidth}
\begin{minipage}{0.6\textwidth} 
    \textbf{Question:} How many women are in the picture?
    \vspace{0.3cm}
    
    \textbf{GT:} \textcolor{magenta}{4}
    \vspace{0.3cm}
    
    \textbf{KeyWords:} woman
    \vspace{0.3cm}

    \textbf{Predicted Answer:} \textcolor{green}{4}
    
    
\end{minipage}
  \\ \noindent\rule{\textwidth}{0.5pt}

\begin{minipage}{0.3\textwidth} 
\noindent
\includegraphics[width=\textwidth]{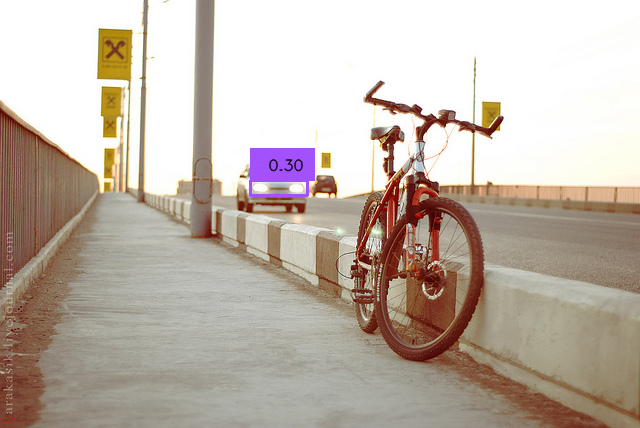} 
\end{minipage}%
\hspace{0.02\textwidth}
\begin{minipage}{0.6\textwidth} 
    \textbf{Question:} How many vehicles have their lights on?
    \vspace{0.3cm}
    
    \textbf{GT:} \textcolor{magenta}{1}
    \vspace{0.3cm}
    
    \textbf{KeyWords:} vehicles, lights
    \vspace{0.3cm}

    \textbf{Predicted Answer:} \textcolor{green}{1}
    
    
\end{minipage}

\noindent\rule{\textwidth}{0.5pt}

\noindent
\textbf{Question:} What kind of fruit is cut in half and darker than the other?

\noindent
\textbf{GT:} \textcolor{magenta}{grape}

\begin{minipage}{0.3\textwidth} 
\noindent
\includegraphics[width=\textwidth]{Images/5.png} 
\end{minipage}%
\hspace{0.02\textwidth}
\begin{minipage}{0.6\textwidth} 
    \textbf{KeyWords:} fruit
    \vspace{0.3cm}
    
    \textbf{Question Concept:} fruit with one darker half
    \vspace{0.3cm}
    
    \textbf{Caption1:} A slice of cherry, where the darker half of the fruit reveals a hidden depth of flavor
    
    \textbf{Caption2:} A plate of food with a fork and knife

    \textbf{Caption3:} The fruit in the image that is typically cut in half and has a darker color on one half compared to the other is a \textcolor{cyan}{grape}.
    
    
\end{minipage}

\noindent
\textbf{QA1:} What type of fruit is typically cut in half and has a darker color on one half compared to the other? \textcolor{orange}{Grape}

\noindent
\textbf{QA2:} What type of fruit is described as having a hidden depth of flavor when cut in half? Cherry

\noindent
\vspace{0.3cm}
\textbf{Predicted Answer:} \textcolor{red}{Cherry}
  \\ \noindent\rule{\textwidth}{0.5pt}

\noindent
\textbf{Question:} What item on the desk could help with a cold?

\noindent
\textbf{GT:} \textcolor{magenta}{cough drops}/\textcolor{magenta}{tissue}/\textcolor{magenta}{ricola}

\begin{minipage}{0.3\textwidth} 
\noindent
\includegraphics[width=\textwidth]{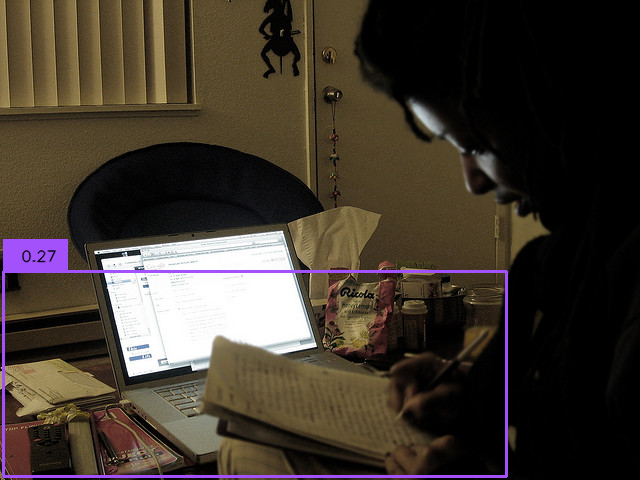} 
\end{minipage}%
\hspace{0.02\textwidth}
\begin{minipage}{0.6\textwidth} 
    \textbf{KeyWords:} cold, desk
    \vspace{0.3cm}
    
    \textbf{Question Concept:} item that helps alleviate a cold
    \vspace{0.3cm}
    
    \textbf{Caption1:} On the desk, there is a bottle of vitamin C, which can help alleviate the symptoms of a cold
    
    \textbf{Caption2:} By taking vitamin C supplements or consuming foods high in vitamin C, such as citrus fruits, berries, and leafy greens, individuals can help reduce the severity and duration of their cold symptoms.

    \textbf{Caption3:} There are several items on the desk that could potentially help alleviate a cold in the image.
    
    
\end{minipage}

\noindent
\textbf{QA1:} What is on the desk that can help alleviate cold symptoms? Vitamin C

\noindent
\textbf{QA2:} What types of foods are high in vitamin C and can help reduce cold symptoms? Citrus fruits

\noindent
\vspace{0.3cm}
\textbf{Predicted Answer:} \textcolor{red}{Vitamin C}
  \\ \noindent\rule{\textwidth}{0.5pt}

\noindent
\textbf{Question:} What activity does the cat appear most likely to do?

\noindent
\textbf{GT:} \textcolor{magenta}{watch television}/\textcolor{magenta}{eat}/\textcolor{magenta}{jump}/\textcolor{magenta}{beg}

\begin{minipage}{0.3\textwidth} 
\noindent
\includegraphics[width=\textwidth]{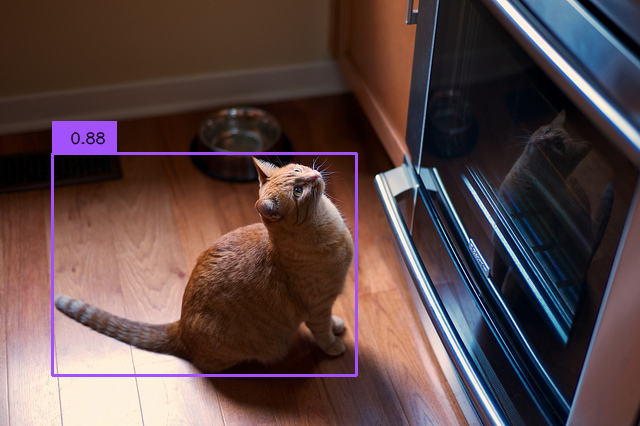} 
\end{minipage}%
\hspace{0.02\textwidth}
\begin{minipage}{0.6\textwidth} 
    \textbf{KeyWords:} cat, activity
    \vspace{0.3cm}
    
    \textbf{Question Concept:} most likely behavior of a cat
    \vspace{0.3cm}
    
    \textbf{Caption1:} The cat is sitting on the floor and looking up, possibly observing its surroundings or searching for something of interest.
    
    \textbf{Caption2:} The cat is taking a moment to look around and explore the environment.

    \textbf{Caption3:} The cat seems to be engaged in a behavior typical of cats when they are curious or attentive to their environment.
    
    
\end{minipage}

\noindent
\textbf{QA1:} What is the cat likely doing in the image? Sitting

\noindent
\textbf{QA2:} What is typical of cats when they are curious or attentive to their environment? Behavior

\noindent
\vspace{0.3cm}
\textbf{Predicted Answer:} \textcolor{red}{Looking}
  \\ \noindent\rule{\textwidth}{0.5pt}

\noindent
\textbf{Question:} Which is not part of specific dish?

\noindent
\textbf{GT:} \textcolor{magenta}{ham}/\textcolor{magenta}{ketchup}/\textcolor{magenta}{ketchup}

\begin{minipage}{0.3\textwidth} 
\noindent
\includegraphics[width=\textwidth]{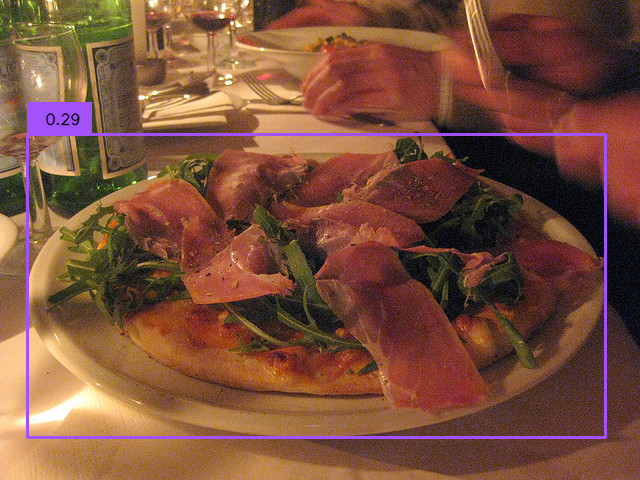} 
\end{minipage}%
\hspace{0.02\textwidth}
\begin{minipage}{0.6\textwidth} 
    \textbf{KeyWords:} ingredient, dish
    \vspace{0.3cm}
    
    \textbf{Question Concept:} ingredient that is not part of a specific dish
    \vspace{0.3cm}
    
    \textbf{Caption1:} Bacon is one of the ingredients that is not part of a specific dish in this image.
    
    \textbf{Caption2:} A slice of ham on a pizza offers an unexpected savory twist to the classic dish.

    \textbf{Caption3:} The green leafy vegetable is not included in this dish, as it is not typically used as a topping for pizza or flatbread.
    
    
\end{minipage}

\noindent
\textbf{QA1:} What is the unexpected flavor ingredient? Bacon

\noindent
\textbf{QA2:} What is the green leafy vegetable? Kale

\noindent
\vspace{0.3cm}
\textbf{Predicted Answer:} \textcolor{red}{Kale}
  \\ \noindent\rule{\textwidth}{0.5pt}

\noindent
\textbf{Question:} Who has the same color hair as this woman?

\noindent
\textbf{GT:} \textcolor{magenta}{maria sharapova}/\textcolor{magenta}{swedish persons}/\textcolor{magenta}{sitting guy}

\begin{minipage}{0.3\textwidth} 
\noindent
\includegraphics[width=\textwidth]{Images/11.png} 
\end{minipage}%
\hspace{0.02\textwidth}
\begin{minipage}{0.6\textwidth} 
    \textbf{KeyWords:} color hair, woman
    \vspace{0.3cm}
    
    \textbf{Question Concept:} someone who has the same hair color as the woman in question
    \vspace{0.3cm}
    
    \textbf{Caption1:} A woman swinging a tennis racket.
    
    \textbf{Caption2:} The woman in the image has blonde hair.

    \textbf{Caption3:} When you're blonde, you're always ready for a good time!
    
    
\end{minipage}

\noindent
\textbf{QA1:} What is the woman's hair color in the image? Blonde

\noindent
\textbf{QA2:} What is often associated with being blonde, according to the quote? A good time

\noindent
\vspace{0.3cm}
\textbf{Predicted Answer:} \textcolor{red}{Blonde}
  \\ \noindent\rule{\textwidth}{0.5pt}

\noindent
\textbf{Question:} The top shelf has many books of the same author of what genre?

\noindent
\textbf{GT:} \textcolor{magenta}{thriller}/\textcolor{magenta}{thriller fiction}/\textcolor{magenta}{non fiction}/\textcolor{magenta}{mystery}

\begin{minipage}{0.3\textwidth} 
\noindent
\includegraphics[width=0.8\textwidth]{Images/12.png} 
\end{minipage}%
\hspace{0.02\textwidth}
\begin{minipage}{0.6\textwidth} 
    \textbf{KeyWords:} many books, top shelf, author, genre
    \vspace{0.3cm}
    
    \textbf{Question Concept:} genre of books on the top shelf
    \vspace{0.3cm}
    
    \textbf{Caption1:} The genre of books on the top shelf in this image appears to be a mix of fiction and \textcolor{cyan}{non-fiction} genres.
    
    \textbf{Caption2:} A collection of diverse reads, from classic literature to modern \textcolor{cyan}{mysteries}, the top shelf of this bookshelf offers a literary journey for every taste.

    \textbf{Caption3:} The top shelf in the image appears to be filled with a variety of books, but without more specific information, it's difficult to determine the exact genre of each book.
    
    
\end{minipage}

\noindent
\textbf{QA1:} What type of books are found on the top shelf in the image? Fiction and \textcolor{orange}{non-fiction}

\noindent
\textbf{QA2:} What kind of literary experience does the top shelf of the bookshelf offer? A literary journey

\noindent
\vspace{0.3cm}
\textbf{Predicted Answer:} \textcolor{red}{Mmix}

\noindent\rule{\textwidth}{0.5pt}

\noindent
\textbf{Question:} Why would we assume this surfer is not about to engage in his sport?

\noindent
\textbf{GT:} \textcolor{magenta}{no water}/\textcolor{magenta}{he looks tired}/\textcolor{magenta}{holding it}/\textcolor{magenta}{walking away}/

\begin{minipage}{0.3\textwidth} 
\noindent
\includegraphics[width=\textwidth]{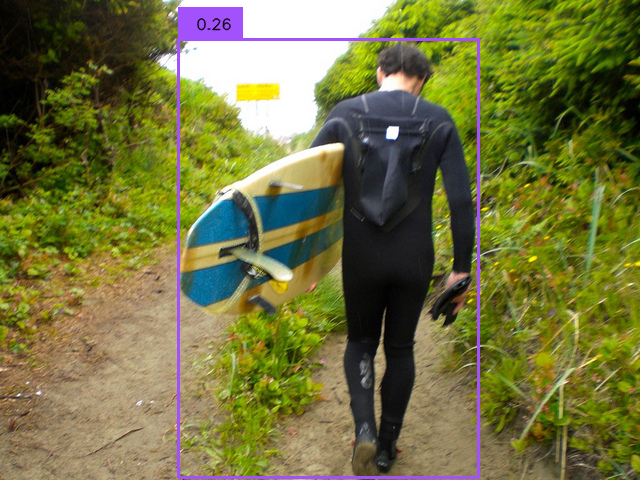} 
\end{minipage}%
\hspace{0.02\textwidth}
\begin{minipage}{0.6\textwidth} 
    \textbf{KeyWords:} surfer
    \vspace{0.3cm}
    
    \textbf{Question Concept:} Reason why the surfer will not do surfing
    \vspace{0.3cm}
    
    \textbf{Caption1:} A surfer's journey begins with a leap into the unknown.
    
    \textbf{Caption2:} A man wearing a wetsuit and carrying a surfboard

    \textbf{Caption3:} The person is wearing a black wetsuit, which is a common attire for surfers, indicating that they might be heading to or returning from a surfing session.
    
    
\end{minipage}

\noindent
\textbf{QA1:} What is the surfer wearing? Wetsuit

\noindent
\textbf{QA2:} What is the surfer carrying? Surfboard

\noindent
\vspace{0.3cm}
\textbf{Predicted Answer:} \textcolor{red}{not surfing}
  \\ \noindent\rule{\textwidth}{0.5pt}

\noindent
\textbf{Question:} Where might this bench be located?

\noindent
\textbf{GT:} \textcolor{magenta}{island}/\textcolor{magenta}{beach}/\textcolor{magenta}{oceanside}/\textcolor{magenta}{ocean}

\begin{minipage}{0.3\textwidth} 
\noindent
\includegraphics[height=110pt, width=\textwidth]{Images/18.png} 
\end{minipage}%
\hspace{0.02\textwidth}
\begin{minipage}{0.6\textwidth} 
    \textbf{KeyWords:} bench
    \vspace{0.3cm}
    
    \textbf{Question Concept:} possible location for a bench
    \vspace{0.3cm}
    
    \textbf{Caption1:} A serene park bench overlooking a lush green field, perfect for a quiet afternoon of reflection.
    
    \textbf{Caption2:} The image shows two people sitting on a wooden bench in a grassy area.

    \textbf{Caption3:} The possible location for the bench in the image is a grassy field.
    
    
\end{minipage}

\noindent
\textbf{QA1:} What is the setting for the scene? Park

\noindent
\textbf{QA2:} What is the material of the bench? Wooden

\noindent
\vspace{0.3cm}
\textbf{Predicted Answer:} \textcolor{red}{Park}
  \\ \noindent\rule{\textwidth}{0.5pt}

\noindent
\textbf{Question:} What is this wall used for?

\noindent
\textbf{GT:} \textcolor{magenta}{artwork}/\textcolor{magenta}{decoration art}/\textcolor{magenta}{holding paintings}/\textcolor{magenta}{hanging pictures}

\begin{minipage}{0.3\textwidth} 
\noindent
\includegraphics[height=110pt, width=\textwidth]{Images/19.png} 
\end{minipage}%
\hspace{0.02\textwidth}
\begin{minipage}{0.6\textwidth} 
    \textbf{KeyWords:} wall
    \vspace{0.3cm}
    
    \textbf{Question Concept:} function of the wall
    \vspace{0.3cm}
    
    \textbf{Caption1:} The wall in this image serves as a decorative backdrop, contributing to the overall decor of the living room.
    
    \textbf{Caption2:} A wall that serves as a canvas for artistic expression, creating a harmonious and inviting atmosphere in the living space.

    \textbf{Caption3:} The wall is used to make the living room look nice.
    
    
\end{minipage}

\noindent
\textbf{QA1:} What is the purpose of the wall in the image? Decorative backdrop

\noindent
\textbf{QA2:} What does the wall serve as in the living space? Canvas

\noindent
\vspace{0.3cm}
\textbf{Predicted Answer:} \textcolor{red}{Decorative backdrop}
  \\ \noindent\rule{\textwidth}{0.5pt}

\noindent
\textbf{Question:} What percent of the stands are full?

\noindent
\textbf{GT:} \textcolor{magenta}{15}/\textcolor{magenta}{ten}/\textcolor{magenta}{twenty}

\begin{minipage}{0.3\textwidth} 
\noindent
\includegraphics[width=\textwidth]{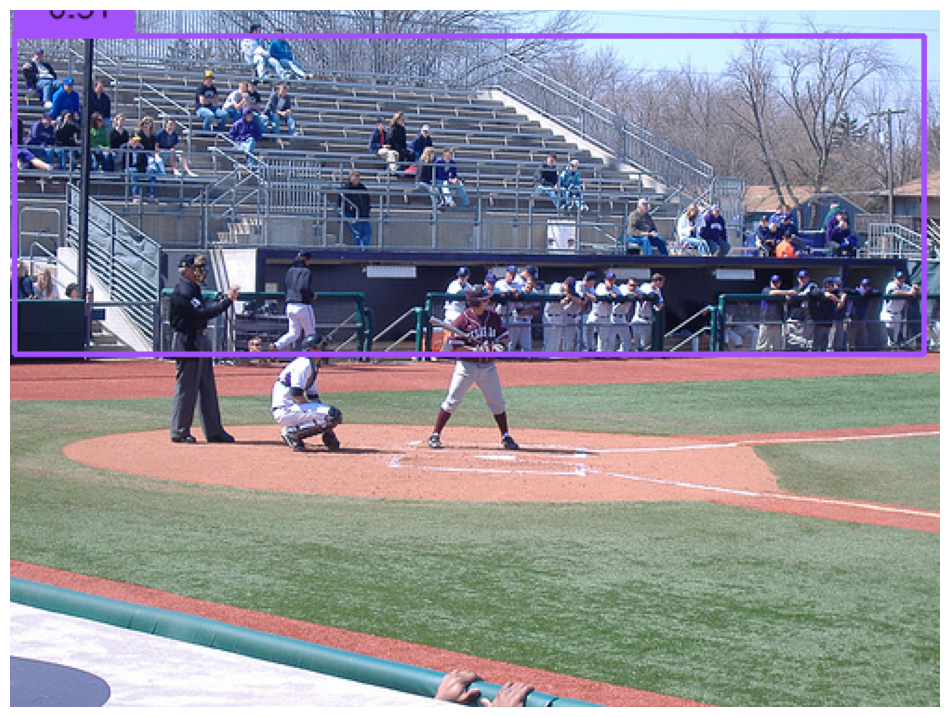} 
\end{minipage}%
\hspace{0.02\textwidth}
\begin{minipage}{0.6\textwidth} 
    \textbf{KeyWords:} percent, stands
    \vspace{0.3cm}
    
    \textbf{Question Concept:} percentage of stands that are full
    \vspace{0.3cm}
    
    \textbf{Caption1:} The stands are only half full, but the energy in the air is palpable as the game unfolds
    
    \textbf{Caption2:} In the image, the stands appear to be approximately 50\% full.

    \textbf{Caption3:} It shows a grandstand with spectators seated on bleachers. 
    
    
\end{minipage}

\noindent
\textbf{QA1:} What percentage of the stands are filled? 50\%

\noindent
\textbf{QA2:} How full do the stands appear to be in the image? 50\%.

\noindent
\vspace{0.3cm}
\textbf{Predicted Answer:} \textcolor{red}{50}
  \\ \noindent\rule{\textwidth}{0.5pt}

\noindent
\textbf{Question:} What is the name of the operating system used for this device?

\noindent
\textbf{GT:} \textcolor{magenta}{mac}/\textcolor{magenta}{windows}/\textcolor{magenta}{apple}

\begin{minipage}{0.3\textwidth} 
\noindent
\includegraphics[width=\textwidth]{Images/33.png} 
\end{minipage}%
\hspace{0.02\textwidth}
\begin{minipage}{0.6\textwidth} 
    \textbf{KeyWords:} operating system, device
    \vspace{0.3cm}
    
    \textbf{Question Concept:} operating system used by a device
    \vspace{0.3cm}
    
    \textbf{Caption1:} The operating system used by the laptop in this image is not specified.
    
    \textbf{Caption2:} The operating system used on this Apple's Mac computer is \textcolor{cyan}{macOS}.

    \textbf{Caption3:}  A dog's curious gaze meets the power of a laptop running on the latest operating system, ready to explore the digital world.
    
    
\end{minipage}

\noindent
\textbf{QA1:} What is the operating system used on this Apple's Mac computers? \textcolor{orange}{MacOS}

\noindent
\textbf{QA2:} How is the dog looking at the computer? Curiously

\noindent
\vspace{0.3cm}
\textbf{Predicted Answer:} \textcolor{red}{MacOS}
  \\ \noindent\rule{\textwidth}{0.5pt}

\noindent
\textbf{Question:} Based on the clothing what year was this picture taken?

\noindent
\textbf{GT:} \textcolor{magenta}{1930}/\textcolor{magenta}{1960}/\textcolor{magenta}{1913}/\textcolor{magenta}{1953}

\begin{minipage}{0.3\textwidth} 
\noindent
\includegraphics[width=\textwidth]{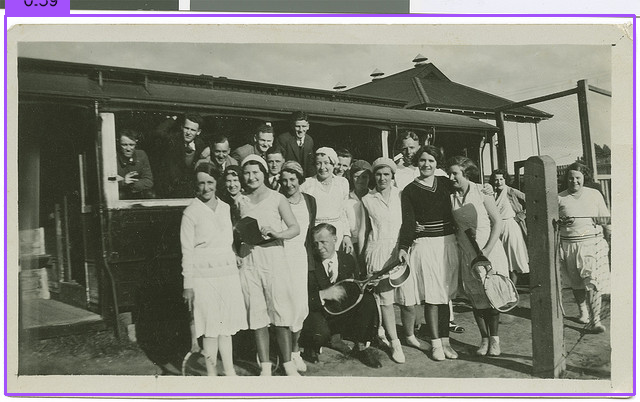} 
\end{minipage}%
\hspace{0.02\textwidth}
\begin{minipage}{0.6\textwidth} 
    \textbf{KeyWords:} year, clothing
    \vspace{0.3cm}
    
    \textbf{Question Concept:} year a picture was taken based on the clothing
    \vspace{0.3cm}
    
    \textbf{Caption1:} The year a picture was taken based on the clothing depicted in the image is around the 1920s or \textcolor{cyan}{1930s}.
    
    \textbf{Caption2:}  A moment frozen in time: 1945, as seen through the lens of a vintage camera.

    \textbf{Caption3:} The image appears to be a vintage black and white photograph, likely from the early to mid-20th century, judging by the clothing and hairstyles of the individuals in the picture.
    
    
\end{minipage}

\noindent
\textbf{QA1:} What is the likely time period of the picture? Mid-20th century

\noindent
\textbf{QA2:} What type of photograph is the image likely to be? Black and white

\noindent
\vspace{0.3cm}
\textbf{Predicted Answer:} \textcolor{red}{1945}
  \\ \noindent\rule{\textwidth}{0.5pt}

\noindent
\textbf{Question:} What decade is this?

\noindent
\textbf{GT:} \textcolor{magenta}{1930}/\textcolor{magenta}{1940}/\textcolor{magenta}{1930s}

\begin{minipage}{0.3\textwidth} 
\noindent
\includegraphics[width=\textwidth]{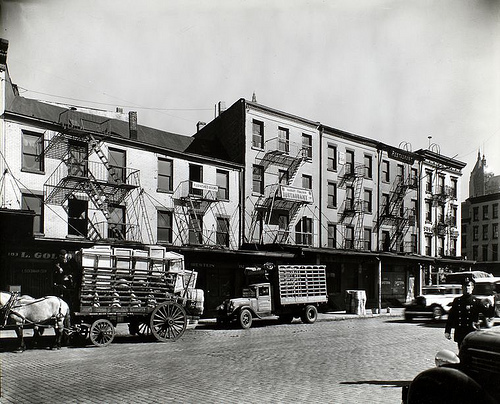} 
\end{minipage}%
\hspace{0.02\textwidth}
\begin{minipage}{0.6\textwidth} 
    \textbf{KeyWords:} decade
    \vspace{0.3cm}
    
    \textbf{Question Concept:} time period
    \vspace{0.3cm}
    
    \textbf{Caption1:} The time period or decade being referred to in this image is the 1920s
    
    \textbf{Caption2:} A moment frozen in time, where the hustle and bustle of a bygone era unfolds.

    \textbf{Caption3:} However, without more specific details, it's difficult to pinpoint the exact decade or time period.
    
    
\end{minipage}

\noindent
\textbf{QA1:} What time period is being referred to in this image? The 1920s

\noindent
\textbf{QA2:} What does the image describe as part of a bygone era? Bustle

\noindent
\vspace{0.3cm}
\textbf{Predicted Answer:} \textcolor{red}{1920s}
  \\ \noindent\rule{\textwidth}{0.5pt}

\noindent
\textbf{Question:} What age range do you think these people are?

\noindent
\textbf{GT:} \textcolor{magenta}{teenagers}/\textcolor{magenta}{20 to 50}/\textcolor{magenta}{20 to 40}

\begin{minipage}{0.3\textwidth} 
\noindent
\includegraphics[width=\textwidth]{Images/38.png} 
\end{minipage}%
\hspace{0.02\textwidth}
\begin{minipage}{0.6\textwidth} 
    \textbf{KeyWords:} age, people
    \vspace{0.3cm}
    
    \textbf{Question Concept:} age range of people
    \vspace{0.3cm}
    
    \textbf{Caption1:} The group of people in the image consists of both young and older individuals, with a range of ages from \textcolor{cyan}{teenagers to adults}.
    
    \textbf{Caption2:} A diverse group of individuals, spanning from young adults to older teens, gathered for a casual pizza party.

    \textbf{Caption3:} The age range of the group is not precisely defined, but it is clear that they are young adults.
    
    
\end{minipage}

\noindent
\textbf{QA1:} What is the age range of the group in the image? Young adults

\noindent
\textbf{QA2:} What type of gathering did the group attend? Pizza party

\noindent
\vspace{0.3cm}
\textbf{Predicted Answer:} \textcolor{red}{young adults}
  \\ \noindent\rule{\textwidth}{0.5pt}

\noindent
\textbf{Question:} In this photo provide the average weight of this type of bird when they are fully grown?

\noindent
\textbf{GT:} \textcolor{magenta}{1 lb}/\textcolor{magenta}{2 ounces}/\textcolor{magenta}{1 pound}/\textcolor{magenta}{half pound}

\begin{minipage}{0.3\textwidth} 
\noindent
\includegraphics[width=\textwidth]{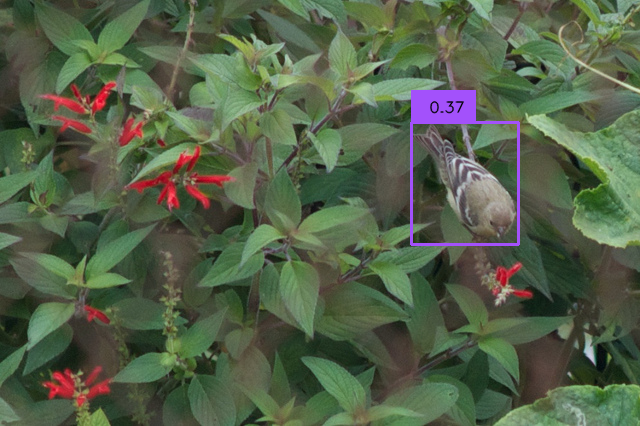} 
\end{minipage}%
\hspace{0.02\textwidth}
\begin{minipage}{0.6\textwidth} 
    \textbf{KeyWords:} average weight, bird
    \vspace{0.3cm}
    
    \textbf{Question Concept:} average weight of fully grown bird
    \vspace{0.3cm}
    
    \textbf{Caption1:} A lightweight feathered friend, gracefully perched on a vibrant red flower.
    
    \textbf{Caption2:} The image shows a small bird, likely a finch or sparrow, with a light brownish-gray body and streaked or striped darker brown/gray wings, weighing approximately 100 grams.

    \textbf{Caption3:} The weight of a bird can vary greatly depending on factors such as age, sex, and species.
    
    
\end{minipage}

\noindent
\textbf{QA1:} What is the bird's perch? Red flower

\noindent
\textbf{QA2:} What is the weight of a bird? 100 grams

\noindent
\vspace{0.3cm}
\textbf{Predicted Answer:} \textcolor{red}{100 grams}
  \\ \noindent\rule{\textwidth}{0.5pt}

\noindent
\textbf{Question:} What is the generational gap between these two people?

\noindent
\textbf{GT:} \textcolor{magenta}{50 years}/\textcolor{magenta}{60 years}/\textcolor{magenta}{40 years}/\textcolor{magenta}{2 generations}

\begin{minipage}{0.3\textwidth} 
\noindent
\includegraphics[width=\textwidth]{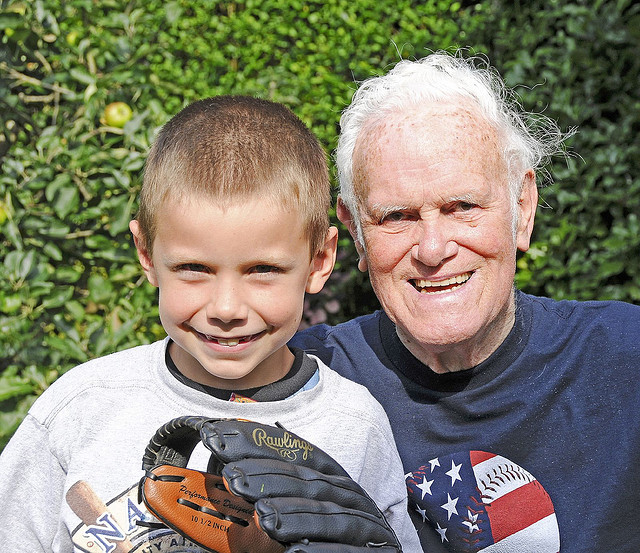} 
\end{minipage}%
\hspace{0.02\textwidth}
\begin{minipage}{0.6\textwidth} 
    \textbf{KeyWords:} people
    \vspace{0.3cm}
    
    \textbf{Question Concept:} difference in age 
    \vspace{0.3cm}
    
    \textbf{Caption1:} The age gap between the two individuals highlights the importance of passing down knowledge and skills from one generation to the next.
    
    \textbf{Caption2:} The two individuals in the image exhibit a notable age difference.

    \textbf{Caption3:} The age difference between them highlights the generational gap between the two individuals, with the older man serving as a mentor or role model for the younger boy.
    
    
\end{minipage}

\noindent
\textbf{QA1:} What is highlighted in the difference in age between the two individuals? knowledge

\noindent
\textbf{QA2:} What does the older man serve as for the younger boy in the image? Mentor

\noindent
\vspace{0.3cm}
\textbf{Predicted Answer:} \textcolor{red}{significant}
  \\ \noindent\rule{\textwidth}{0.5pt}

\noindent
\textbf{Question:} How much water can the teapot in the photo hold?

\noindent
\textbf{GT:} \textcolor{magenta}{2 quarts}/\textcolor{magenta}{1 gallon}/\textcolor{magenta}{2 liters}

\begin{minipage}{0.3\textwidth} 
\noindent
\includegraphics[width=\textwidth]{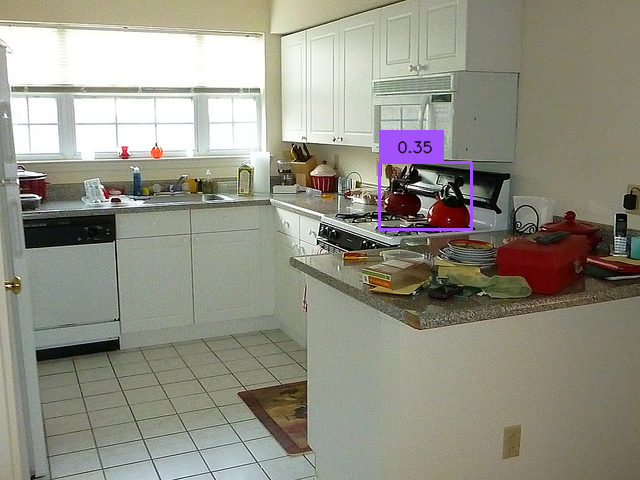} 
\end{minipage}%
\hspace{0.02\textwidth}
\begin{minipage}{0.6\textwidth} 
    \textbf{KeyWords:} teapot
    \vspace{0.3cm}
    
    \textbf{Question Concept:} capacity of the teapot
    \vspace{0.3cm}
    
    \textbf{Caption1:} The teapot has a capacity of two cups.
    
    \textbf{Caption2:} The teapot can hold up to two cups of tea.

    \textbf{Caption3:} The capacity of the teapot is not clearly visible in the image, but it seems to be a standard size for a teapot, which is typically around 1 to \textcolor{cyan}{2 liters}.
    
    
\end{minipage}

\noindent
\textbf{QA1:} What is the capacity of the teapot mentioned in the caption?
Two cups

\noindent
\textbf{QA2:} What is the typical size range of a standard teapot? 1 to 2 liters

\noindent
\vspace{0.3cm}
\textbf{Predicted Answer:} \textcolor{red}{1 to 2 liters}
  \\ \noindent\rule{\textwidth}{0.5pt}

\noindent
\textbf{Question:} How much protein is in this meal?

\noindent
\textbf{GT:} \textcolor{magenta}{lot}/\textcolor{magenta}{little}/\textcolor{magenta}{5 grams}

\begin{minipage}{0.3\textwidth} 
\noindent
\includegraphics[width=\textwidth]{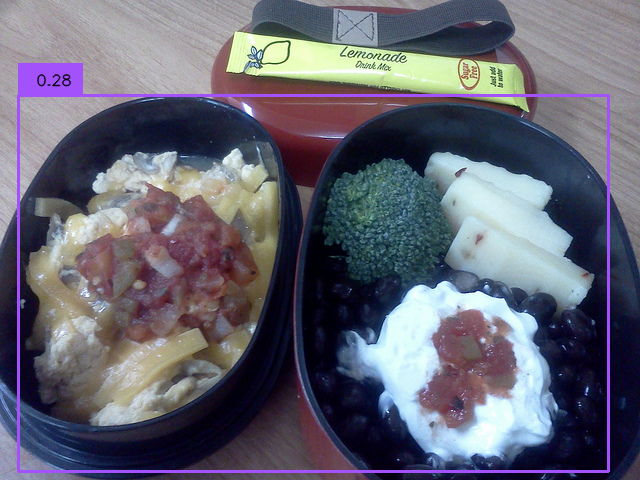} 
\end{minipage}%
\hspace{0.02\textwidth}
\begin{minipage}{0.6\textwidth} 
    \textbf{KeyWords:} protein, meal
    \vspace{0.3cm}
    
    \textbf{Question Concept:} amount of protein in a specific meal
    \vspace{0.3cm}
    
    \textbf{Caption1:} The meal in this image contains a moderate amount of protein, as evidenced by the presence of black beans, broccoli, and cheese.
    
    \textbf{Caption2:} This meal packs a protein punch with its combination of chicken, beans, and broccoli, providing a balanced and nutritious option.

    \textbf{Caption3:} The meal on the left would likely contain around 20-30 grams of protein per serving, while the meal on the right would contain around 15-20 grams of protein per serving.
    
    
\end{minipage}

\noindent
\textbf{QA1:} What is the source of protein in the meal? Beans

\noindent
\textbf{QA2:} What is the estimated amount of protein in the meal on the left?  20-30 grams

\noindent
\vspace{0.3cm}
\textbf{Predicted Answer:} \textcolor{red}{20-30 grams}
  \\ \noindent\rule{\textwidth}{0.5pt}

\noindent
\textbf{Question:} How much do these boats weigh?

\noindent
\textbf{GT:} \textcolor{magenta}{5 tons}/\textcolor{magenta}{4.57}/\textcolor{magenta}{1000 pounds}/\textcolor{magenta}{ton}

\begin{minipage}{0.3\textwidth} 
\noindent
\includegraphics[width=\textwidth]{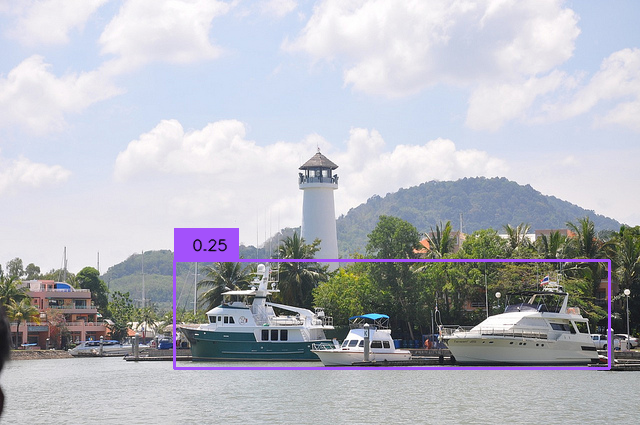} 
\end{minipage}%
\hspace{0.02\textwidth}
\begin{minipage}{0.6\textwidth} 
    \textbf{KeyWords:} boats
    \vspace{0.3cm}
    
    \textbf{Question Concept:} weight of the boats
    \vspace{0.3cm}
    
    \textbf{Caption1:} The weight of the boats in this image ranges from 10,000 to 20,000 pounds, with the largest boat weighing around 20,000 pounds and the smaller boats weighing between 10,000 and 15,000 pounds
    
    \textbf{Caption2:} A trio of vessels, each with its own story, anchored in the serene waters, ready to embark on their next adventure.

    \textbf{Caption3:} Without specific dimensions or weight information, it's not possible to provide an exact weight for these boats
    
    
\end{minipage}

\noindent
\textbf{QA1:} What is the range of weights for the boats in the image? 10,000 to 20,000 pounds

\noindent
\textbf{QA2:} Where are the vessels anchored? Serene water

\noindent
\vspace{0.3cm}
\textbf{Predicted Answer:} \textcolor{red}{10,000 to 20,000 pounds}
  \\ \noindent\rule{\textwidth}{0.5pt}

\noindent
\textbf{Question:} How much does it cost for city parking?

\noindent
\textbf{GT:} \textcolor{magenta}{50 cents}/\textcolor{magenta}{60 cents}/\textcolor{magenta}{50 cents}/\textcolor{magenta}{0.5}/\textcolor{magenta}{0.6}

\begin{minipage}{0.3\textwidth} 
\noindent
\includegraphics[width=\textwidth]{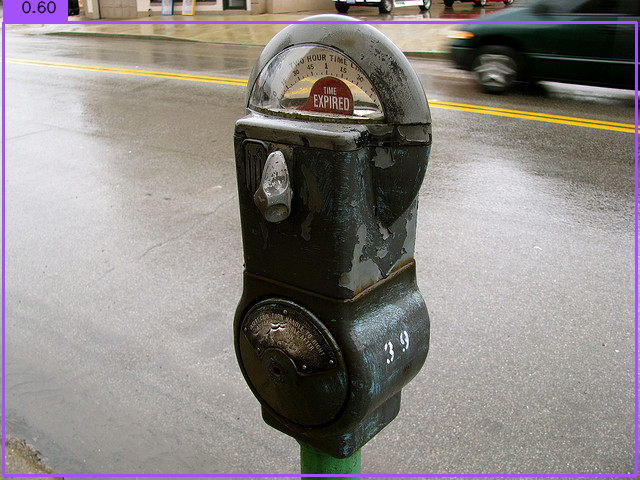} 
\end{minipage}%
\hspace{0.02\textwidth}
\begin{minipage}{0.6\textwidth} 
    \textbf{KeyWords:} city parking
    \vspace{0.3cm}
    
    \textbf{Question Concept:} city parking cost
    \vspace{0.3cm}
    
    \textbf{Caption1:} This suggests that the parking fees in this city are relatively low, as there are only two cars parked in the vicinity of the meter.
    
    \textbf{Caption2:} A reminder of the city's parking fees, standing tall on the side of the road

    \textbf{Caption3:} The fees are usually based on the duration of parking and can range from a few cents to several dollars per hour, depending on the location and the city's parking.
    
    
\end{minipage}

\noindent
\textbf{QA1:} What is the basis of parking fees in the city? Duration

\noindent
\textbf{QA2:} What is the range of parking fees in the city? A few cents to several dollars per hour

\noindent
\vspace{0.3cm}
\textbf{Predicted Answer:} \textcolor{red}{A few cents to several dollars}

\newpage

\begin{longtable}{|>{\centering\arraybackslash}m{0.49\textwidth}|>{\centering\arraybackslash}m{0.49\textwidth}|} 
    \hline
    \textbf{Ours} & \textbf{Img2LLM} \\
    \hline
    \endfirsthead

    \hline
    \textbf{Ours} & \textbf{Img2LLM} \\
    \hline
    \endhead

    \hline
    \endlastfoot

    \begin{minipage}[t]{0.48\textwidth}
        \centering
        \includegraphics[width=0.8\textwidth]{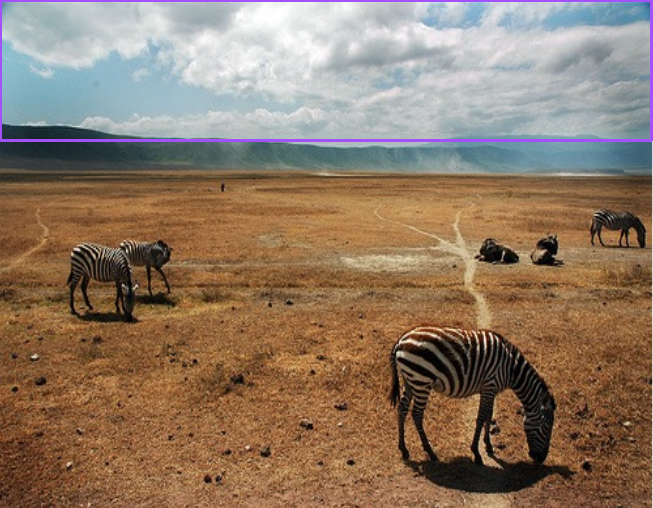} 
        \vspace{1mm} \\ 
        \raggedright 
        \textbf{Q:} What type of clouds are in the picture?\\
        \textbf{GT:} \textcolor{magenta}{cumulus}/\textcolor{magenta}{cumuli}/\textcolor{magenta}{nimbus}\\
        \textbf{KWs:} clouds \\
        \textbf{Cap1:} In the image, there are various types of clouds visible in the sky, including fluffy white clouds and darker, more ominous-looking clouds.\\
        \textbf{Cap2:} A sky filled with fluffy, \textcolor{cyan}{cumulus} clouds, suggesting a pleasant day with a chance of rain.\\
        \textbf{Cap3:} In the given image, there are several types of clouds visible in the sky.\\
        \textbf{QA1:} What is visible in the sky? Clouds\\
        \textbf{QA2:} What type of clouds are fluffy and white? \textcolor{orange}{Cumulus}\\
        \textbf{Generated Answer:} \textcolor{green}{cumulus}
        \vspace{1mm} \\
    \end{minipage}
    &
    \begin{minipage}[t]{0.48\textwidth}
        \centering
        \includegraphics[width=0.8\textwidth]{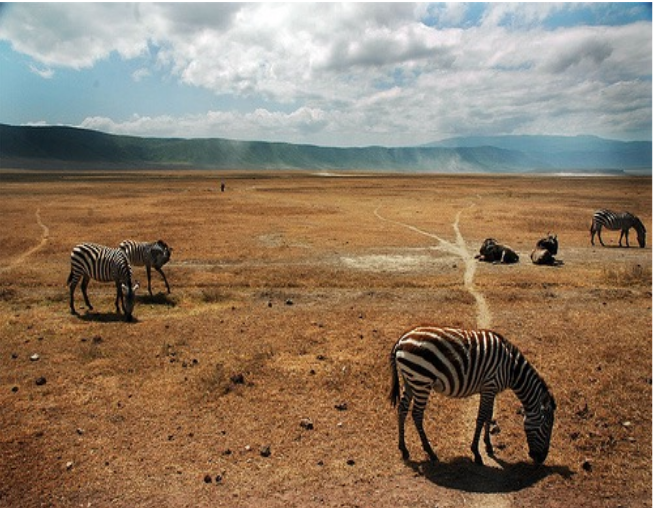} 
        \vspace{1mm} \\ 
        \raggedright 
        \textbf{Q:} What type of clouds are in the picture?\\
        \textbf{GT:} \textcolor{magenta}{cumulus}/\textcolor{magenta}{cumuli}/\textcolor{magenta}{nimbus}\\
        \textbf{Cap1:} a cloudy - filled sky on a cloudy day over a zebras.\\
        \textbf{Cap2:} the clouds are gray and full of clouds.\\
        \textbf{Cap3:} there are many different clouds in this sky.\\
        \textbf{QA1:} what is in the background of a photo of a zebra? sky\\
        \textbf{QA2:} what type of sky is above on a cloudy day? cloudy\\
        \textbf{Generated Answer:} \textcolor{red}{cloud}
        \vspace{1mm} \\
    \end{minipage}
    \\ \hline

    \begin{minipage}[t]{0.48\textwidth}
        \centering
        \includegraphics[width=0.8\textwidth]{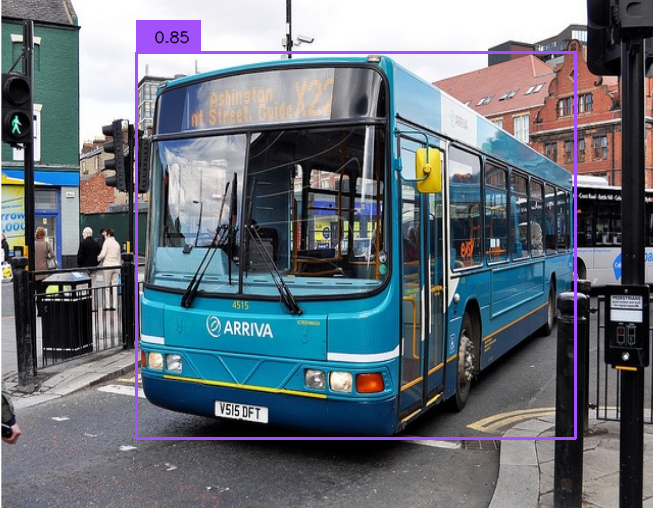} 
        \vspace{1mm} \\ 
        \raggedright 
        \textbf{Q:} How many people can this bus carry?\\
        \textbf{GT:} \textcolor{magenta}{50}/\textcolor{magenta}{40}/\textcolor{magenta}{39}\\
        \textbf{KWs:} bus, people \\
        \textbf{Cap1:} A blue bus is parked on the side of the road.\\
        \textbf{Cap2:} Arriving at Ashington, the X22 bus is ready to transport passengers to their destinations.\\
        \textbf{Cap3:} There are several people visible in the scene, some standing near the bus and others walking along the sidewalk.\\
        \textbf{QA1:} What is parked on the side of the road? bus\\
        \textbf{QA2:} Where is the X22 bus ready to transport passengers? Ashington\\
        \textbf{Generated Answer:} \textcolor{green}{50}
        \vspace{1mm} \\
    \end{minipage}
    &
    \begin{minipage}[t]{0.48\textwidth}
        \centering
        \includegraphics[width=0.8\textwidth]{Images/Compare_22.png} 
        \vspace{1mm} \\ 
        \raggedright 
        \textbf{Q:} How many people can this bus carry?\\
        \textbf{GT:} \textcolor{magenta}{50}/\textcolor{magenta}{40}/\textcolor{magenta}{39}\\
        \textbf{Cap1:} a passenger bus traveling on a street side.\\
        \textbf{Cap2:} blue commuter bus with parked on the side of the road.\\
        \textbf{Cap3:} a bus that says aradara rides down the street.\\
        \textbf{QA1:} what color bus is driving down the street? blue\\
        \textbf{QA2:} what is making it's way down the street? bus\\
        \textbf{Generated Answer:} \textcolor{red}{many}
        \vspace{1mm} \\
    \end{minipage}
    \\ \hline

    \begin{minipage}[t]{0.48\textwidth}
        \centering
        \includegraphics[width=0.8\textwidth]{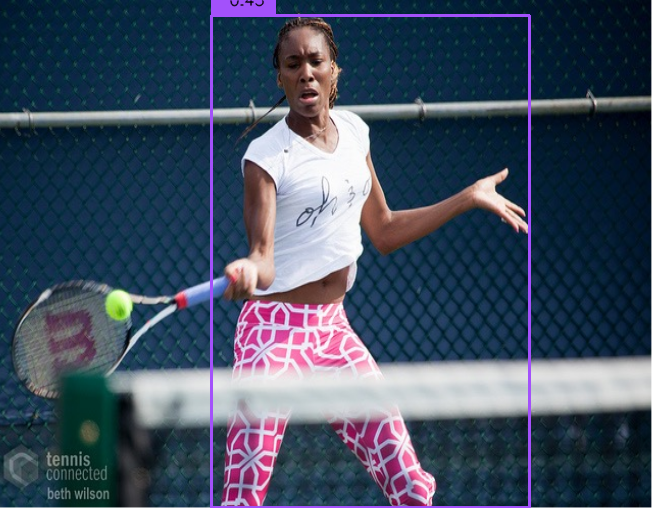} 
        \vspace{1mm} \\ 
        \raggedright 
        \textbf{Q:} What type of shot is the woman about to hit?\\
        \textbf{GT:} \textcolor{magenta}{forehand}/\textcolor{magenta}{tennis shot}/\textcolor{magenta}{swing}\\
        \textbf{KWs:} shot, woman \\
        \textbf{Cap1:} The woman in the image is \textcolor{cyan}{swinging} a tennis racket at a tennis ball, preparing to hit it during a game of tennis.\\
        \textbf{Cap2:} In the image, a woman is captured in the midst of a powerful \textcolor{cyan}{forehand} shot on a tennis court.\\
        \textbf{Cap3:} She is holding a tennis racket and has just hit the ball with it.\\
        \textbf{QA1:} What is the woman holding in her hand? Tennis racket\\
        \textbf{QA2:} What is the woman about to do on the court? \textcolor{orange}{Swing}\\
        \textbf{Generated Answer:} \textcolor{green}{Swing}
        \vspace{1mm} \\
    \end{minipage}
    &
    \begin{minipage}[t]{0.48\textwidth}
        \centering
        \includegraphics[width=0.8\textwidth]{Images/Compare_32.png} 
        \vspace{1mm} \\ 
        \raggedright 
        \textbf{Q:} What type of shot is the woman about to hit?\\
        \textbf{GT:} \textcolor{magenta}{forehand}/\textcolor{magenta}{tennis shot}/\textcolor{magenta}{swing}\\
        \textbf{Cap1:} tennis player is hitting a tennis ball with her racket.\\
        \textbf{Cap2:} a woman in pink outfit hitting a tennis ball.\\
        \textbf{Cap3:} a woman in a cropped top and pants swinging a tennis racquet.\\
        \textbf{QA1:} what is a tennis player doing with a tennis racket? swinging\\
        \textbf{QA2:} who is swinging a tennis racket at a tennis ball? woman\\
        \textbf{Generated Answer:} \textcolor{red}{volley}
        \vspace{1mm} \\
    \end{minipage}
    \\ \hline

        \begin{minipage}[t]{0.48\textwidth}
        \centering
        \includegraphics[width=0.8\textwidth]{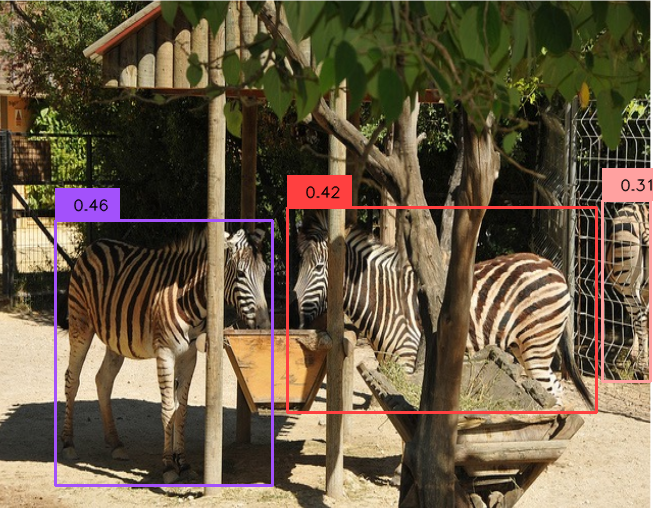} 
        \vspace{1mm} \\ 
        \raggedright 
        \textbf{Q:} How many zebras are there?\\
        \textbf{GT:} \textcolor{magenta}{3}\\
        \textbf{KWs:} zebras \\
        \textbf{Generated Answer:} \textcolor{green}{3}
        \vspace{1mm} \\
    \end{minipage}
    &
    \begin{minipage}[t]{0.48\textwidth}
        \centering
        \includegraphics[width=0.8\textwidth]{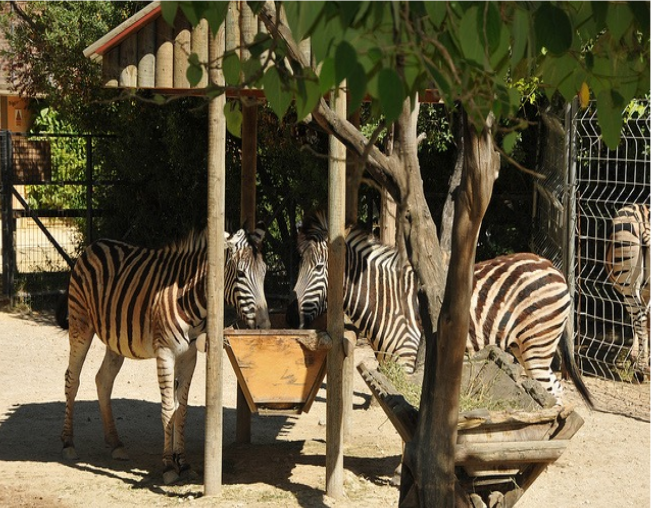} 
        \vspace{1mm} \\ 
        \raggedright 
        \textbf{Q:} How many zebras are there?\\
        \textbf{GT:} \textcolor{magenta}{3} \\
        \textbf{Cap1:} two zebras walking by a feeder full of food.\\
        \textbf{Cap2:} pair of zebras standing together at water trough in zoo.\\
        \textbf{Cap3:} the zebras are eating out of a feeder box.\\
        \textbf{QA1:} how many zebras are standing next to each other? two\\
        \textbf{QA2:} what are the zebras doing? eating\\
        \textbf{Generated Answer:} \textcolor{red}{2}
        \vspace{1mm} \\
    \end{minipage}
    \\ \hline

    \begin{minipage}[t]{0.48\textwidth}
        \centering
        \includegraphics[width=0.8\textwidth]{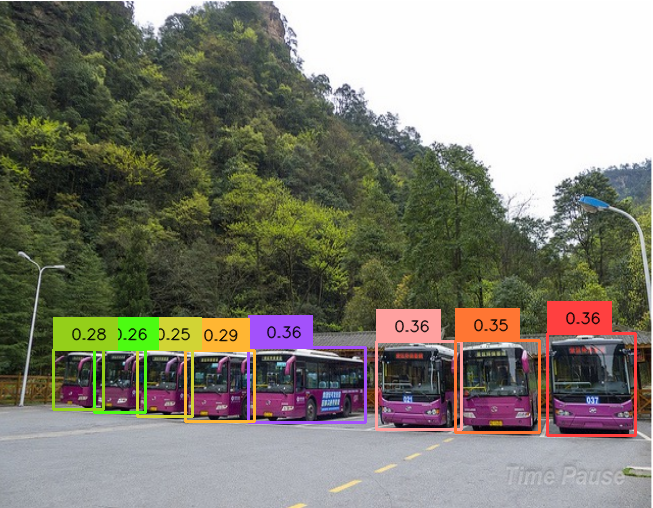} 
        \vspace{1mm} \\ 
        \raggedright 
        \textbf{Q:} How many buses are in the picture?\\
        \textbf{GT:} \textcolor{magenta}{8}\\
        \textbf{KWs:} buses \\
        \textbf{Generated Answer:} \textcolor{green}{8}
        \vspace{1mm} \\
    \end{minipage}
    &
    \begin{minipage}[t]{0.48\textwidth}
        \centering
        \includegraphics[width=0.8\textwidth]{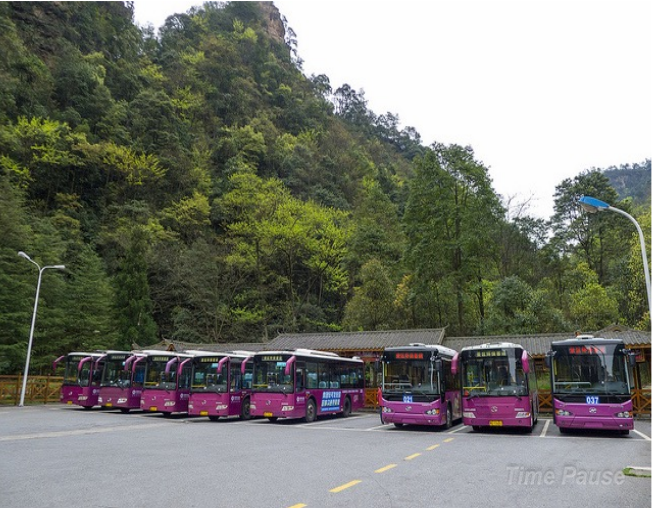} 
        \vspace{1mm} \\ 
        \raggedright 
        \textbf{Q:} How many buses are in the picture?\\
        \textbf{GT:} \textcolor{magenta}{8} \\
        \textbf{Cap1:} a lot of buses sit parked in a line in front of a hill.\\
        \textbf{Cap2:} a group of purple passenger buses all in a row.\\
        \textbf{Cap3:} a row of purple bus buses next to each other.\\
        \textbf{QA1:} how are the buses parked? a line\\
        \textbf{QA2:} what color buses are parked in front of each other? purple\\
        \textbf{Generated Answer:} \textcolor{red}{several}
        \vspace{1mm} \\
    \end{minipage}
    \\ \hline

    \begin{minipage}[t]{0.48\textwidth}
        \centering
        \includegraphics[width=0.8\textwidth]{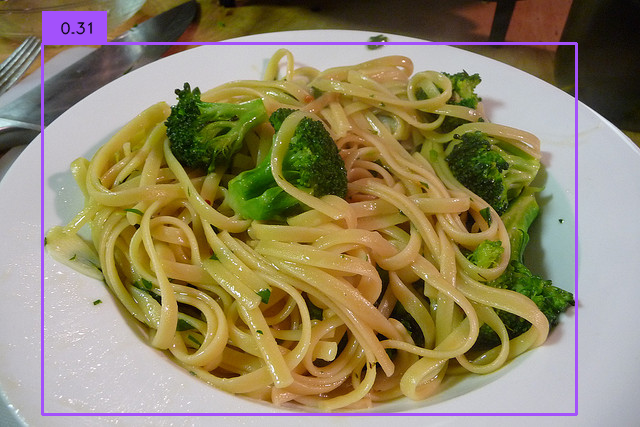} 
        \vspace{1mm} \\ 
        \raggedright 
        \textbf{Q:} This dish is suitable for which group of people?\\
        \textbf{GT:} \textcolor{magenta}{vegetarian}/\textcolor{magenta}{vegan}/\textcolor{magenta}{family}\\
        \textbf{KWs:} dish \\
        \textbf{Cap1:} This dish is suitable for a variety of people, including \textcolor{cyan}{vegetarians} and those who enjoy a simple, hearty meal.\\
        \textbf{Cap2:} The target group of people for the pasta and broccoli dish in this image is likely health-conscious individuals who are looking for a low-calorie, nutritious meal option.\\
        \textbf{Cap3:} The target group of people for this dish could include \textcolor{cyan}{families}, individuals looking for a quick and easy meal, \textcolor{cyan}{vegetarians}, and those who enjoy a balance of carbohydrates and vegetables in their diet.\\
        \textbf{QA1:} What is the target group of people for the pasta and broccoli dish? \textcolor{orange}{Vegetarians}\\
        \textbf{QA2:} What type of meal is the pasta and broccoli dish? Low-calorie\\
        \textbf{Generated Answer:} \textcolor{green}{Vegetarians}
        \vspace{1mm} \\
    \end{minipage}
    &
    \begin{minipage}[t]{0.48\textwidth}
        \centering
        \includegraphics[width=0.8\textwidth]{Images/Compare_42.png} 
        \vspace{1mm} \\ 
        \raggedright 
        \textbf{Q:} This dish is suitable for which group of people?\\
        \textbf{GT:} \textcolor{magenta}{vegetarian}/\textcolor{magenta}{vegan}/\textcolor{magenta}{family}\\
        \textbf{Cap1:} a pasta dish sitting on top of a white plate.\\
        \textbf{Cap2:} a broccoli pasta dish that has very pasta.\\
        \textbf{Cap3:} a dish of pasta with noodles and tomato sauce.\\
        \textbf{QA1:} what vegetable is on a white plate? broccoli\\
        \textbf{QA2:} what color is a plate of pasta with broccoli on it? white\\
        \textbf{Generated Answer:} \textcolor{red}{children}
        \vspace{1mm} \\
    \end{minipage}
    \\ \hline

    \begin{minipage}[t]{0.48\textwidth}
        \centering
        \includegraphics[width=0.8\textwidth]{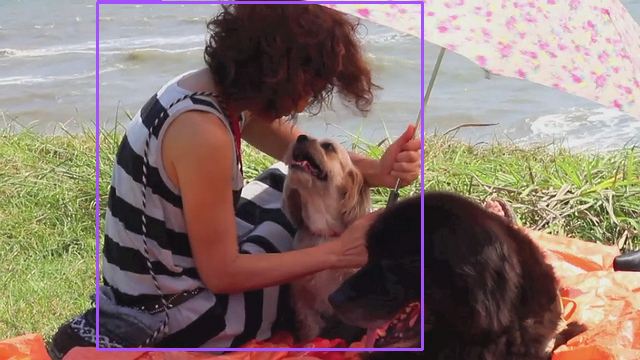} 
        \vspace{2mm} \\ 
        \raggedright 
        \textbf{Q:} Why is the woman holding the umbrella?\\
        \textbf{GT:} \textcolor{magenta}{shade}/\textcolor{magenta}{sun protection}/\textcolor{magenta}{get shadow}\\
        \textbf{KWs:} woman, umbrella \\
        \textbf{Cap1:} The umbrella provides \textcolor{cyan}{shade} and protection from the sun for both the woman and the dogs.\\
        \textbf{Cap2:} The umbrella is open and provides \textcolor{cyan}{shade} for the woman and the two dogs she is interacting with.\\
        \textbf{Cap3:} The woman is holding the umbrella over her head, which indicates that she is using it to shield herself and the dogs from the weather.\\
        \textbf{QA1:} What is the woman using to shield herself and the dogs from the weather? Umbrella\\
        \textbf{QA2:} What is the woman doing with the umbrella? Holding\\
        \textbf{Generated Answer:} \textcolor{green}{Shade}
        \vspace{1mm} \\
    \end{minipage}
    &
    \begin{minipage}[t]{0.48\textwidth}
        \centering
        \includegraphics[width=0.8\textwidth]{Images/Compare_52.png} 
        \vspace{2mm} \\ 
        \raggedright 
        \textbf{Q:} Why is the woman holding the umbrella?\\
        \textbf{GT:} \textcolor{magenta}{shade}/\textcolor{magenta}{sun protection}/\textcolor{magenta}{get shadow}\\
        \textbf{Cap1:} a young woman and the umbrella are on an orange blanket.\\
        \textbf{Cap2:} a woman's umbrella and two dogs under an umbrella.\\
        \textbf{Cap3:} a woman holding an umbrella is getting some light under her umbrella.\\
        \textbf{QA1:} who is holding an umbrella while her dog sits under it? woman\\
        \textbf{QA2:} : what is a woman holding and a dog under it? an umbrella\\
        \textbf{Generated Answer:} \textcolor{red}{to protect herself from the sun}
        \vspace{1mm} \\
    \end{minipage}
    \\ \hline

    \begin{minipage}[t]{0.48\textwidth}
        \centering
        \includegraphics[width=0.8\textwidth]{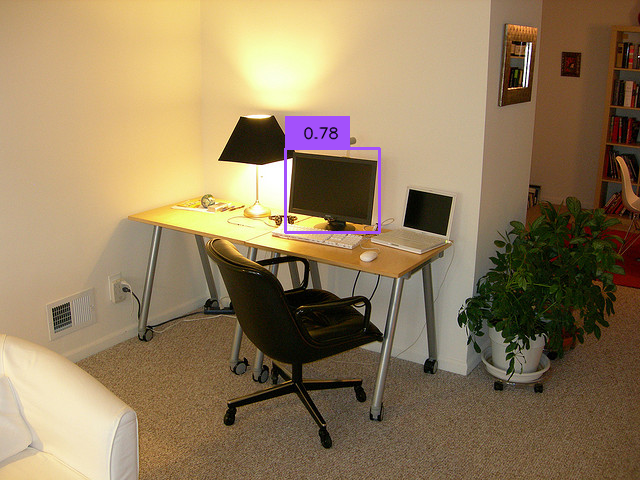} 
        \vspace{1mm} \\ 
        \raggedright 
        \textbf{Q:} What is in front of the monitor?\\
        \textbf{GT:} \textcolor{magenta}{chair}/\textcolor{magenta}{keyboard}/\textcolor{magenta}{webcam}\\
        \textbf{KWs:} monitor, front \\
        \textbf{Cap1:} The computer features a black monitor and a \textcolor{cyan}{keyboard}, which are placed next to each other on the desk.\\
        \textbf{Cap2:} A white \textcolor{cyan}{keyboard} and mouse, ready for productivity\\
        \textbf{Cap3:} A clean and organized computer desk with a black monitor and a white \textcolor{cyan}{keyboard}.\\
        \textbf{QA1:} What color is the keyboard on the desk? White\\
        \textbf{QA2:} What is placed next to the monitor? \textcolor{orange}{Keyboard}\\
        \textbf{Generated Answer:} \textcolor{green}{Keyboard}
        \vspace{1mm} \\
    \end{minipage}
    &
    \begin{minipage}[t]{0.48\textwidth}
        \centering
        \includegraphics[width=0.8\textwidth]{Images/Compare_62.png} 
        \vspace{1mm} \\ 
        \raggedright 
        \textbf{Q:} what is in front of the monitor?\\
        \textbf{GT:} \textcolor{magenta}{chair}/\textcolor{magenta}{keyboard}/\textcolor{magenta}{webcam}\\
        \textbf{Cap1:} a corner table with computer computer on the desk.\\
        \textbf{Cap2:} a computer on the small desk in a small office area.\\
        \textbf{Cap3:} view of a computer monitor in a light lit room.\\
        \textbf{QA1:} what is a computer sitting on in a corner of a room? desk\\
        \textbf{QA2:} how big is the desk in the corner? small\\
        \textbf{Generated Answer:} \textcolor{red}{desk}
        \vspace{1mm} \\
    \end{minipage}
    \\ \hline

    \begin{minipage}[t]{0.48\textwidth}
        \centering
        \includegraphics[width=0.8\textwidth]{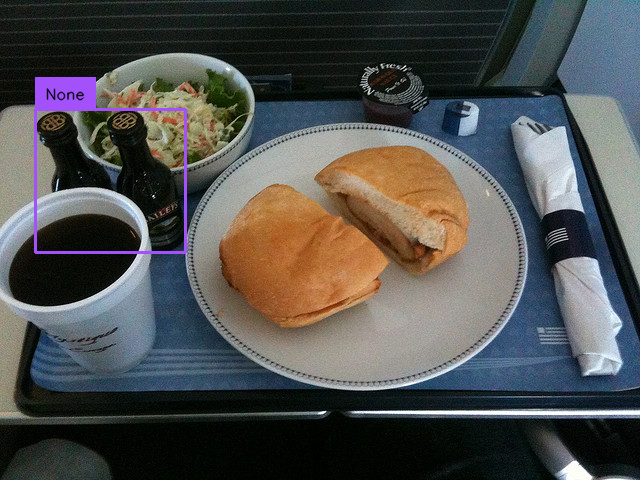} 
        \vspace{1mm} \\ 
        \raggedright 
        \textbf{Q:} What is in the bottles?\\
        \textbf{GT:} \textcolor{magenta}{alcohol}/\textcolor{magenta}{liqueur}/\textcolor{magenta}{baileys}\\
        \textbf{KWs:} bottles \\
        \textbf{Cap1:} The two bottles of beer are positioned close to the bowl of salad, indicating that they may have been used to accompany the meal.\\
        \textbf{Cap2:}  A pair of dark beverages, possibly coffee or tea, served alongside a fresh salad, ready to complement the meal with a warm or cool sip.\\
        \textbf{Cap3:} There are two bottles of dark liquid, possibly coffee or tea, with a white cup of dark liquid in the image.\\
        \textbf{QA1:} What are placed on the table? Bottles\\
        \textbf{QA2:} What could be the contents of the bottles? Beer \\
        \textbf{Generated Answer:} \textcolor{red}{Beer}
        \vspace{1mm} \\
    \end{minipage}
    &
    \begin{minipage}[t]{0.48\textwidth}
        \centering
        \includegraphics[width=0.8\textwidth]{Images/Compare_92.png} 
        \vspace{1mm} \\ 
        \raggedright 
        \textbf{Q:} what is in the bottles?\\
        \textbf{GT:} \textcolor{magenta}{alcohol}/\textcolor{magenta}{liqueur}/\textcolor{magenta}{baileys} \\
        \textbf{Cap1:} a sandwich on a plate with a glass of beer bottle.\\
        \textbf{Cap2:} a table that has a sandwich, beer, and beer on it.\\
        \textbf{Cap3:} a sandwich on a plate with a glass of beer bottle.\\
        \textbf{QA1:} What is next to a sandwich and a beer? bottle\\
        \textbf{QA2:} where is a sandwich with a beer and beer on a plate? table\\
        \textbf{Generated Answer:} \textcolor{red}{beer}
        \vspace{1mm} \\
    \end{minipage}
    \\ \hline


\end{longtable}
\end{document}